\documentclass[lettersize,journal]{IEEEtran}
\usepackage{amsmath,amsfonts}
\usepackage{array}
\usepackage{algorithmicx,algorithm}
\usepackage[caption=false,font=normalsize,labelfont=sf,textfont=sf]{subfig}
\usepackage[noend]{algpseudocode}
\usepackage{textcomp}
\usepackage{stfloats}
\usepackage{url}
\usepackage{verbatim}
\usepackage{graphicx}
\usepackage{cite}
\usepackage{hyperref}
\usepackage{booktabs}
\usepackage{blindtext}
\usepackage{tabularx}
\usepackage{multirow}
\usepackage{pifont}
\usepackage[capitalize]{cleveref}
\usepackage{bbold}
\usepackage{amssymb}
\usepackage{indentfirst}
\usepackage{autobreak}
\usepackage{tikz}
\usepackage{epstopdf}
\usepackage{diagbox}
\usepackage{epstopdf}
\usepackage{wasysym}
\usepackage{balance}
\usepackage{makecell}
\usepackage[normalem]{ulem}
\crefname{section}{Sec.}{Secs.}
\Crefname{section}{Section}{Sections}
\Crefname{table}{Table}{Tables}
\crefname{table}{Tab.}{Tabs.}

\usepackage{adjustbox} 

\hypersetup{hypertex=true,
	colorlinks=true,
	linkcolor=blue,
	anchorcolor=blue,
	citecolor=blue}

\begin{document}

\title{Evaluating Human Perception of Novel View Synthesis: Subjective Quality Assessment of Gaussian Splatting and NeRF in Dynamic Scenes}

\author{Yuhang Zhang$^1$, Joshua Maraval$^1$, Zhengyu Zhang$^*$, Nicolas Ramin, Shishun Tian, Lu Zhang
\thanks{Yuhang Zhang and Shishun Tian with the Guangdong Key Laboratory of Intelligent Information Processing,
College of Electronics and Information Engineering, Shenzhen University, Shenzhen 518060, China (e-mail: zhangyuhang2019@email.szu.edu.cn, stian@szu.edu.cn).}
\thanks{Nicolas Ramin is with IRT b$<>$com, France (e-mail: Nicolas.ramin@b-com.com).}
\thanks{Zhengyu Zhang is with the School of Electronics and Communication Engineering, Guangzhou University, Guangzhou, China (e-mail: zhengyuzhang@gzhu.edu.cn).}
\thanks{Joshua Maraval and Lu Zhang are with the Univ Rennes, INSA Rennes, CNRS, IETR - UMR 6164, F-35000 Rennes, France (e-mail: Joshua.maraval@insa-rennes.fr, lu.zhang@insa-rennes.fr).}
\thanks{$^1$ Contributed equally to this work.}
\thanks{$^*$ Corresponding author.}}

\markboth{Journal of \LaTeX\ Class Files,~Vol.~14, No.~8, August~2021}%
{Shell \MakeLowercase{\textit{et al.}}: A Sample Article Using IEEEtran.cls for IEEE Journals}


\maketitle

\begin{abstract}
Gaussian Splatting (GS) and Neural Radiance Fields (NeRF) are two groundbreaking technologies that have revolutionized the field of Novel View Synthesis (NVS), enabling immersive photorealistic rendering and user experiences by synthesizing multiple viewpoints from a set of images of sparse views. The potential applications of NVS, such as high-quality virtual and augmented reality, detailed 3D modeling, and realistic medical organ imaging, underscore the importance of quality assessment of NVS methods from the perspective of human perception. Although some previous studies have explored subjective quality assessments for NVS technology, they still face several challenges, especially in NVS methods selection, scenario coverage, and evaluation methodology. To address these challenges, we conducted two subjective experiments for the quality assessment of NVS technologies containing both GS-based and NeRF-based methods, focusing on dynamic and real-world scenes. This study covers 360°, front-facing, and single-viewpoint videos while providing a richer and greater number of real scenes. Meanwhile, it's the first time to explore the impact of NVS methods in dynamic scenes with moving objects. The two types of subjective experiments help to fully comprehend the influences of different viewing paths from a human perception perspective and pave the way for future development of full-reference and no-reference quality metrics. In addition, we established a comprehensive benchmark of various state-of-the-art objective metrics on the proposed database, highlighting that existing methods still struggle to accurately capture subjective quality. The results give us some insights into the limitations of existing NVS methods and may promote the development of new NVS methods.
\end{abstract}


\section{Introduction}
\label{sec:1}
\IEEEPARstart{3}{D} immersive media such as virtual reality, augmented reality, and 360-degree videos, rapidly occupied human daily life in recent years. These media and technologies provide realistic immersive user environments and experiences, allowing users to have lifelike experiences without leaving home, such as virtual shopping, medical diagnosis, games, and virtual guided tours\footnote{Online example of Radiance Field-based virtual visit of a gallery: https://current-exhibition.com/laboratorio31/}. Even though some techniques such as Depth Image Based Rendering (DIBR) \cite{tian2018benchmark} and Multi-View Stereo (MVS) \cite{chen2019point} can render realistic novel views in ideal conditions, they have limitations for complex light effects in scene reconstruction. To achieve a more realistic immersive experience, two new methods, Neural Radiance Fields (NeRF) \cite{mildenhall2021nerf} and Gaussian Splatting \cite{kerbl20233d} (GS), have revolutionized the 3D model representation for photorealistic Novel View Synthesis (NVS). NeRF employs a neural network to model 3D representation, predicting the color and opacity of any point in 3D space, rather than relying on discrete geometric primitives such as triangle meshes or voxel grids in typical traditional 3D rendering methods. GS constructs 3D Gaussian ellipsoids to represent the influence of each data point within a volume, then overlays and integrates the colors and opacities of related 3D Gaussian ellipsoids when rendering a novel view to create smooth and continuous visual transitions. Note that both NeRF and GS have their unique advantages: NeRF leverages deep learning techniques to easily handle large and complex scenes, while GS achieves faster rendering since it does not require the usage of neural networks. 

\begin{figure}[t]
    \begin{center}
        \includegraphics[width=1\linewidth]{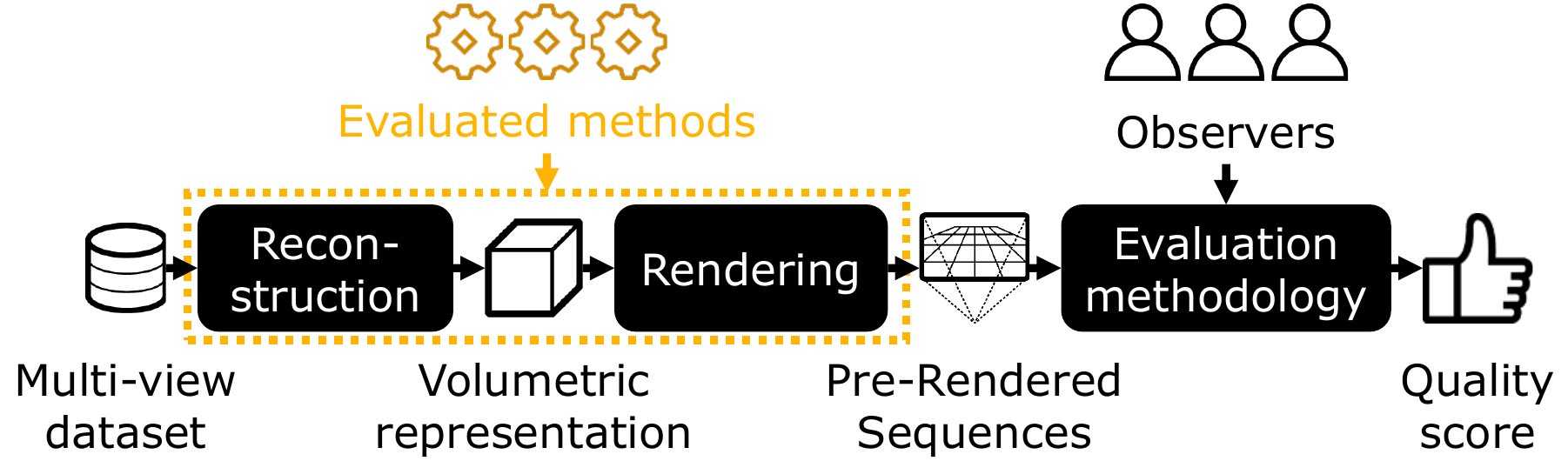}
    \end{center}
    \caption{Subjective evaluation of Radiance Field methods illustration. Multiple multi view video sequences are selected as sources. Pre-rendered Video sequences (PVS) are prepared by reconstructing the sequences with an ensemble of Radiance Field methods to be evaluated. Following a selected evaluation methodology, a quality score is obtained from the observers' evaluation of the sequences.}
    \label{fig:subjective_evaluation_methodology}
\end{figure}
\par While these NVS methods have demonstrated better performance than traditional methods, they may have difficulties in certain complex scenarios, especially when encountering rapid camera movement, large-scale scenes, and a sparse set of data. Thus, it is essential to assess the efficacy of NVS methods, as this enables a deeper insight into the differences and limitations of these methods and to re-design and refine them around existing shortcomings. Typically, the quality assessment can be done by some models, such as PSNR \cite{hore2010image} and SSIM \cite{wang2004image}, which is called objective assessment; but the ultimate and the most reliable test should be done by using human observers when the final user is human, which is called subjective assessment. We illustrate the subjective evaluation process of Radiance Field methods in Fig. \ref{fig:subjective_evaluation_methodology}: a dataset of selection of multi-view sources is used for generating Pre-rendered Video sequences from an ensemble of reconstruction methods to be evaluated by the observers. Some studies\cite{martin2023nerf, martin2024nerfqa, liang2024perceptual, onuoha2023evaluation, xing2024explicit_nerf_qa, tabassum2024quality, yang2024benchmark} have investigated the subjective quality assessment of NVS methods, i.e., using human eyes to rate the Processed Video Sequences (PVS) generated by various NVS methods one by one, but existing studies have not fully explored the relationship between human visual perception and the PVS generated by NVS methods. The limitations are summarized as follows:

\par 1) Existing datasets \cite{martin2023nerf, martin2024nerfqa, onuoha2023evaluation, xing2024explicit_nerf_qa} tend to only focus on the quality evaluation of NeRF-based methods, and ignore the quality evaluation of GS-based methods, which are the most recent highly influential approach. 
\par 2) Multiple previous studies \cite{martin2023nerf, martin2024nerfqa, onuoha2023evaluation, xing2024explicit_nerf_qa, yang2024benchmark} use synthetic Multi-view dataset (generated by 3D modelling software such as Blender\cite{blender2018}). From a synthetic 3D model, both input views for volumetric reconstruction and reference views of the rendered sequences can be generated, which permits full-reference evaluation from the observers. However, while synthetic datasets allow for controlled full-reference evaluation, they often lack the complexity and unpredictability of real-world scenes, making evaluation on real content more representative of practical performance in real applications.
\par 3) Existing datasets majorly employ either front-facing or 360° scenes PVS as the carrier for quality assessment, rather than utilizing them simultaneously. Additionally, existing datasets usually use PVS with multi-view paths for quality assessment, neglecting the quality assessment on static viewpoint PVS. Consequently, these methods difficultly conduct a comprehensive performance evaluation of NVS methods.
\par 4) There is still a research gap related to NVS quality assessment in dynamic scenes. The previous NVS quality assessment benchmarks perform subjective experiments in static scenes without moving objects, and all the objects involved in the images are static, such as buildings, trees, and so on. Conversely, NVS for dynamic scenes models time-varying structure and appearance, further extending the utility of NVS methods, which can be reflected in many applications such as virtual 3D teleportation and live sports events. 

\par To fill the above gaps, in this paper, we perform subjective experiments using the Subjective Assessment Methodology for Video Quality (SAMVIQ)\cite{blin2006new}, a methodology already explored for NVS evaluation \cite{tian2018benchmark}. We construct a subjective quality assessment dataset simultaneously involving NeRF-based and GS-based methods in dynamic scenes. Different from other existing datasets, the proposed dataset considers richer real scenes and covers PVS quality assessment from different perspectives, multi-view and single-view, in order to comprehensively understand the performance and weaknesses of existing NVS methods. The contributions of this paper can be summarized as follows:

\begin{itemize}
\item A quality assessment dataset for NVS methods in dynamic scenes is constructed. To the best of our knowledge, the proposed dataset is the first NVS quality assessment (NVSQA) dataset involving GS-based methods and dynamic scenes.
\item The proposed dataset includes 13 real-world scenes, covering 360°, front-view moving, and single-viewpoint types of PVS for the subjective test, which is richer and has a greater number of real scenes than other datasets.
\item The effectiveness of the mainstream NVS methods is verified through two different subjective experiments, no-reference multi-view path evaluation and reference-based single-view path evaluation, revealing the relationship of the same scene under different viewing paths.
\item Subjective quality assessment of NVS methods in more challenging dynamic scenes is conducted. 
\item A comprehensive evaluation of various state-of-the-art objective metrics on the proposed dataset is performed, highlighting that existing methods still face challenges in accurately reflecting subjective quality assessments. 
\end{itemize}
\par The rest of this paper is organized as follows. First, Section \ref{sec:2} gives a description about the related work. After that, Section \ref{sec:3} introduced detailed the two subjective experiments. Then, Section \ref{sec:4} presents experimental results and analyse. Section \ref{sec:5} presents the benchmark of diverse superior objective metrics on our proposed dataset. Finally, Section \ref{sec:6} concludes this paper.

\begin{table*}[t]
	\centering
	\footnotesize
	\renewcommand{\arraystretch}{1.6}
	\caption{Comparison of Various NVS Methods. '-' represents unknown or 0. Env. represents the subjective environment. Cs. denotes crowdsource.}
	\resizebox{\textwidth}{!}{
		\begin{tabularx}{\textwidth}{c|c|c|c|c|c|c|c|c|c|>{\centering\arraybackslash}X}
			\hline
			\multirow{3}{*}{\textbf{Name}} & \multirow{3}{*}{\textbf{Year}}&\multirow{3}{*}{\textbf{No. SRCs}} & \multirow{3}{*}{\textbf{PVS}} & \multirow{3}{*}{\textbf{R-SRCs}} & \multirow{3}{*}{\textbf{No. Obs.}} & \multirow{3}{*}{\textbf{Subj. Met.}} & \multirow{3}{*}{\textbf{Purpose}} & \multicolumn{2}{c|}{\textbf{No. NVS Met.}} & \multirow{3}{*}{\textbf{Env.}} \\ \cline{9-10}
			&  &  &  &  &  &  & & \textbf{NeRF} & \textbf{GS} & \\ 
			&  &  &  &  &  &  & & \textbf{-based} & \textbf{-based} & \\ \hline
			\multirow{2}{*}{Nerf-QA \cite{martin2023nerf}} & \multirow{2}{*}{2023}& 8 & \multirow{2}{*}{360°} & \multirow{2}{*}{Static} & \multirow{2}{*}{21} & \multirow{2}{*}{DSCQS} & \multirow{2}{*}{NVS quality} & \multirow{2}{*}{7} & \multirow{2}{*}{-} & \multirow{2}{*}{Lab} \\ \cline{3-3}
			& & R.: 4, S.: 4 & & & & & & & & \\ \hline
			\multirow{2}{*}{NeRF-VSQA \cite{martin2024nerfqa}} & \multirow{2}{*}{2024}&16 & \multirow{2}{*}{360°+ Front} & \multirow{2}{*}{Static} & \multirow{2}{*}{22} & \multirow{2}{*}{DSCQS} & \multirow{2}{*}{NVS quality} & \multirow{2}{*}{7} & \multirow{2}{*}{-} & \multirow{2}{*}{Lab} \\ \cline{3-3}
			& & R.: 8, S.: 8 & & & & & & & & \\ \hline
			\multirow{2}{*}{FFV \cite{liang2024perceptual}} & \multirow{2}{*}{2024}&15 & \multirow{2}{*}{Front} & \multirow{2}{*}{Static} & \multirow{2}{*}{39} & \multirow{2}{*}{PC} & \multirow{2}{*}{NVS quality} & \multirow{2}{*}{8} & \multirow{2}{*}{-} & \multirow{2}{*}{Lab} \\ \cline{3-3}
			& & R.: 15 & & & & & & & & \\ \hline
			\multirow{2}{*}{EQM \cite{onuoha2023evaluation}} &\multirow{2}{*}{2023}& 8 & \multirow{2}{*}{360°} & \multirow{2}{*}{Static} & \multirow{2}{*}{120} & \multirow{2}{*}{ACR} & \multirow{2}{*}{Environment} & \multirow{2}{*}{7} & \multirow{2}{*}{-} & \multirow{2}{*}{Cs.} \\ \cline{3-3}
			& & R.: 4, S.: 4 & & & & & & & & \\ \hline
			\multirow{2}{*}{ENeRF-QA \cite{xing2024explicit_nerf_qa}} &\multirow{2}{*}{2024}& 22 & \multirow{2}{*}{360°} & \multirow{2}{*}{Static} & \multirow{2}{*}{21} & \multirow{2}{*}{DSIS} & \multirow{2}{*}{NVS quality} & \multirow{2}{*}{4} & \multirow{2}{*}{-} & \multirow{2}{*}{Lab} \\ \cline{3-3}
			& & S.: 22 & & & & & & & & \\ \hline
			\multirow{2}{*}{QNC \cite{tabassum2024quality}} & \multirow{2}{*}{2024}&4 & \multirow{2}{*}{360°} & \multirow{2}{*}{Static} & \multirow{2}{*}{18} & \multirow{2}{*}{PC} & \multirow{2}{*}{Path effect} & \multirow{2}{*}{3} & \multirow{2}{*}{-} & \multirow{2}{*}{Lab} \\ \cline{3-3}
			& & R.: 4 & & & & & & & & \\ \hline
			\multirow{2}{*}{GSC-QA \cite{yang2024benchmark}} &\multirow{2}{*}{2024}& 15 & \multirow{2}{*}{360°} & \multirow{2}{*}{Dynamic} & \multirow{2}{*}{-} & \multirow{2}{*}{DSIS} & \multirow{2}{*}{GS compression} & \multirow{2}{*}{-} & \multirow{2}{*}{1} & \multirow{2}{*}{Lab} \\ \cline{3-3}
			& & R.: 6, S.: 9 & & & & & & & & \\ \hline
			\multirow{2}{*}{NVS-QA (Ours)} & & 13 & {360° + Front} & \multirow{2}{*}{Dynamic} & \multirow{2}{*}{34} & \multirow{2}{*}{SAMVIQ} & {NVS quality} & \multirow{2}{*}{2} & \multirow{2}{*}{3} & \multirow{2}{*}{Lab} \\ \cline{3-3}
			& & R.: 13 & + Single viewpoint & & & & + Path effect & & & \\ \hline
	\end{tabularx}}
	\label{tab:datasetcomp}
\end{table*}
\section{Related Work}
\label{sec:2}
\subsection{Novel View Synthesis Technology}
Novel View Synthesis (NVS) aims to synthesize images from arbitrary novel viewpoint, relying on a set of images that contains overlapped content. Essentially, NVS can be regarded as a subtask of 3D reconstruction and is included in it. Since 3D reconstruction targets constructing the 3D shape for the object, it is possible to project the 3D shape onto a 2D image from any angle. Common methods include Structure From Motion (SFM)\cite{ullman1979interpretation}, MVS\cite{chen2019point}, and DIBR\cite{tian2018benchmark}. SFM aims to calculate camera's parameters of the input images and generate the sparse point cloud from a set of images with motion (i.e., the coarse 3D shape). MVS utilizes the parameters calculated by SFM and performs a dense reconstruction. DIBR focuses on synthesizing the novel view using RGB and depth images instead of using only RGB images. 
\par In recent years, NeRF \cite{mildenhall2021nerf} has emerged as a new NVS technology, which has renewed the boom in NVS research due to its effectiveness and impressive performance. Given a viewpoint, NeRF trains a deep-learning model to predict the color and opacity of sample points of each ray. In particular, it utilizes differentiable volume rendering of light rays to map the 3D shape into 2D images and compares it with the original 2D image to optimize the model. Instant-NGP \cite{muller2022instant} realized a multi-resolution hash encoding and lightweight attention, ensuring a faster rendering without losing accuracy. Mip-NeRF \cite{barron2021mip} replaces ray in NeRF with anti-aliased conical frustums and fits conical frustums with 3D Gaussians, which reduces objectionable aliasing artifacts. Fridovich et al. \cite{fridovich2023k} proposed a 4D volume factorization method that makes their model K-Planes interpretable. Tancik\cite{tancik2023nerfstudio} et al. developed a Python framework that supports various NeRF-based methods, such as Mip-NeRF \cite{barron2021mip}, K-Planes \cite{fridovich2023k}, Instant-NGP \cite{muller2022instant}, and so on, significantly lowering the threshold for using the NeRF-series methods. More importantly, by integrating the advantages of these methods, Tancik \cite{tancik2023nerfstudio} et al. proposed a new approach called Nerfacto. Relying on the strong representational ability of deep learning, NeRF-based methods yield remarkable NVS results, but suffer from the high-costly training and rendering when faced with high-resolution data and large-scale scenarios. 
\par To address this challenge, GS \cite{kerbl20233d} employs explicit Gaussian functions for 3D shape construction and utilizes rasterization for efficient data visualization, combining the strengths of 3D Gaussian spatial representation and rendering speed. Similar to NeRF, only static scenes without moving objects can be handled by GS, therefore dynamic GS \cite{luiten2024dynamic} was proposed aiming at dynamic scenes. STGFS \cite{li2024spacetime} is also applicable to dynamic scenes, and it proposes optimizations for dynamic view synthesis and reducing rendering complexity, which is critical in real-time view synthesis applications where latency degrades the user experience. Several methods aim to decrease memory usage and enhance rendering speed by employing vector quantization techniques, such as C3GS\cite{niedermayr2024compressed} and Compact3D \cite{lee2024compact}. 
\par The recent success of GS-based and NeRF-based NVS methods has significantly heightened interest among researchers in assessing NVS quality from the perspective of the human visual system.

\subsection{NVSQA Databases}
\begin{figure*}[t] 
	\centering
	\begin{minipage}[b]{0.30\textwidth}
		\centering
		\includegraphics[width=\textwidth]{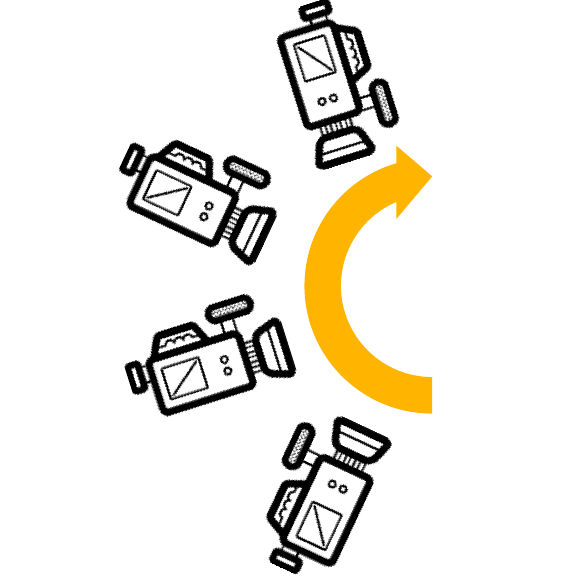}
		\centerline{(a) 360°}
		\label{fig:camera_concentric}
	\end{minipage}
	\hfill
	\begin{minipage}[b]{0.30\textwidth}
		\centering
		\includegraphics[width=\textwidth]{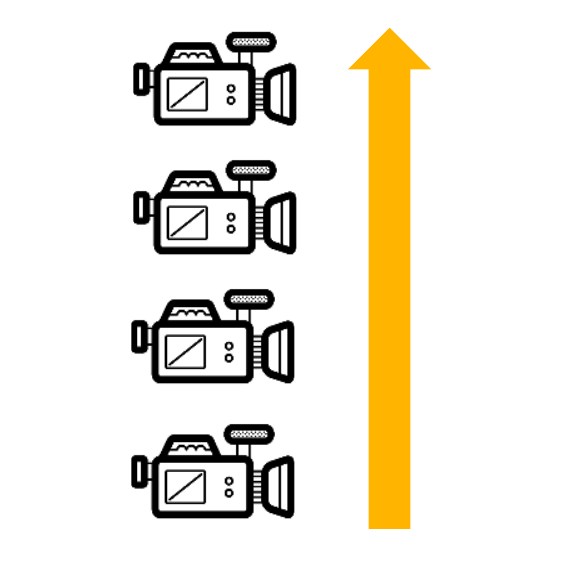}
		\centerline{(b) Front-facing}
		\label{fig:camera_front}
	\end{minipage}
	\hfill
	\begin{minipage}[b]{0.30\textwidth}
		\centering
		\includegraphics[width=\textwidth]{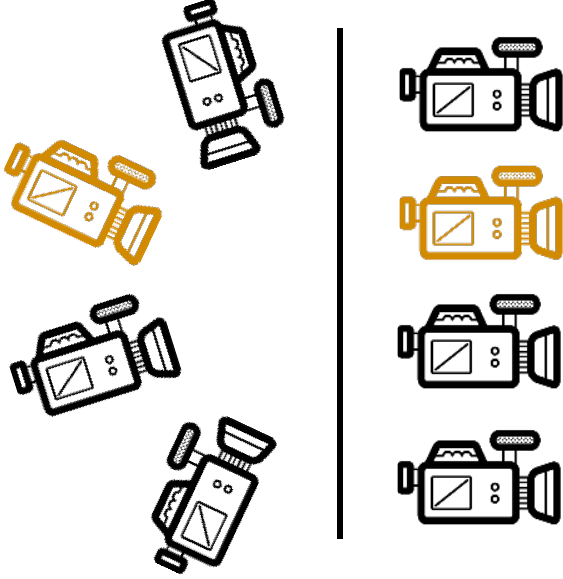}
		\centerline{(c) Single View}
		\label{fig:camera_single}
	\end{minipage}
	\caption{Each source sequence consists of a rig of camera either 360° (depicted in (a)), front-facing (depicted in (b)), or static single viewpoint (depicted in (c)). Yellow highlights the types of virtual camera for Pre-Rendered Video Sequences:  360° virtual camera transition (a), front-facing virtual camera transition (b), or static single viewpoint virtual camera (c).}
	\label{fig:virtual_camera_paths}
\end{figure*}

Several NVSQA databases \cite{martin2023nerf, martin2024nerfqa, liang2024perceptual, onuoha2023evaluation, xing2024explicit_nerf_qa, tabassum2024quality, yang2024benchmark} focusing on different perspectives have been developed, which are summarized in TABLE \ref{tab:datasetcomp}. The compared terms include Number of sources (No SRCs), types of Pre-rendered Video Sequences (PVS), presence of dynamic scenes, number of compared NVS methods (No. NVS Met.), number of observers (No. Obs), subjective evaluation methodology (Subj. Met.), purpose and environment (Env). To enhance comprehension, the terms used in the table are explained below first before delving into the discussion of these datasets.
\begin{itemize}
\item \textbf{No. SRCs} represents the number of source sequences. R. and S. relatively denote real-world and synthetic data.
\item \textbf{PVS} stands for Processed Video Sequences, which include 360° videos, front-facing videos, and single-viewpoint videos, as illustrated Fig. \ref{fig:virtual_camera_paths}. In subjective experiments, these videos will be viewed directly and given scores by observers.
\item \textbf{R-SRCs} is the type of reconstructed scene, containing static and dynamic scenes.  The distinction lies in whether the scene contains moving objects.
\item \textbf{No. Obs.} is the number of the observers. 
\item \textbf{Subj. Met.} denotes the methods using the subjective experiment. DSCQS is Double Stimulus Continuous Quality Scale \cite{ITU2023_500}. PC and ACR respectively represent Pairwise comparison and Absolute Category Rating \cite{ITU2023}. DSIS \cite{ITU2023_500} and SAMVIQ \cite{ITU2023_1788} stand for Double Stimulus Continuous Quality Scale and Subjective Assessment Methodology for Video Quality, respectively.
\item \textbf{Purpose} represents the purpose of study. Most of the databases tried to explore the quality of NVS methods from the human vision system. Some works explore the effect of the experimental environment, path change, and 3D shape/model compression.
\item \textbf{No. NVS Met.} is the number of NVS methods, which can be divided into Nerf-based and GS-based methods.
\item \textbf{Environment} in the subjective experiments is categorized into lab and crowdsource settings. In a lab setting, observers participate in the experiment within a controlled environment that typically provides consistent lighting and minimal distractions. In contrast, the crowdsource setting allows observers to take part in the experiment from various uncontrolled environments, offering a diverse range of viewing conditions and a larger number of observers.
\end{itemize}

\begin{figure*}[t] 
	\centering
	\begin{minipage}[b]{0.24\textwidth}
		\centering
		\includegraphics[width=\textwidth]{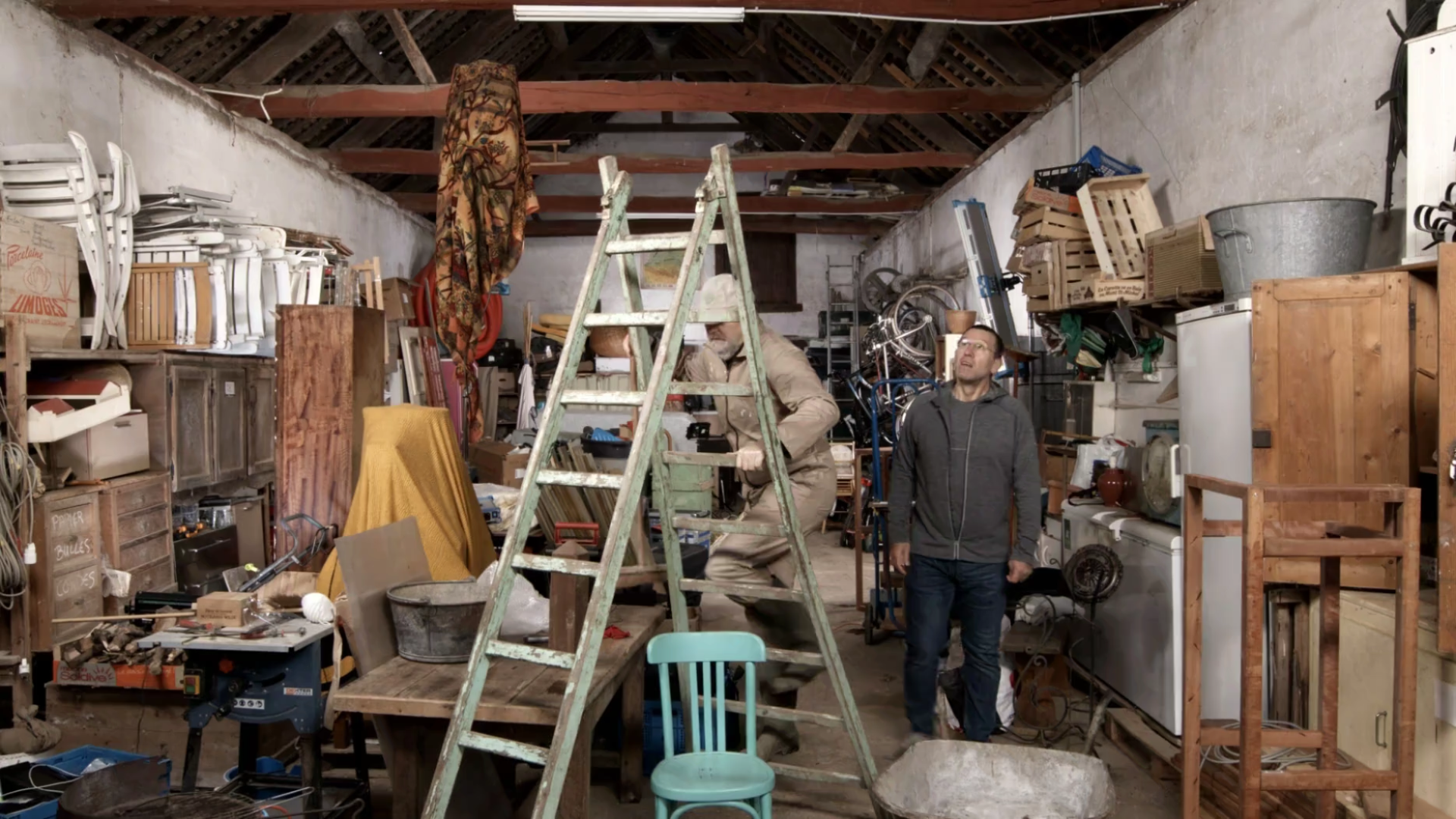}
		\centerline{(a). Barn}
	\end{minipage}
	\begin{minipage}[b]{0.24\textwidth}
		\centering
		\includegraphics[width=\textwidth]{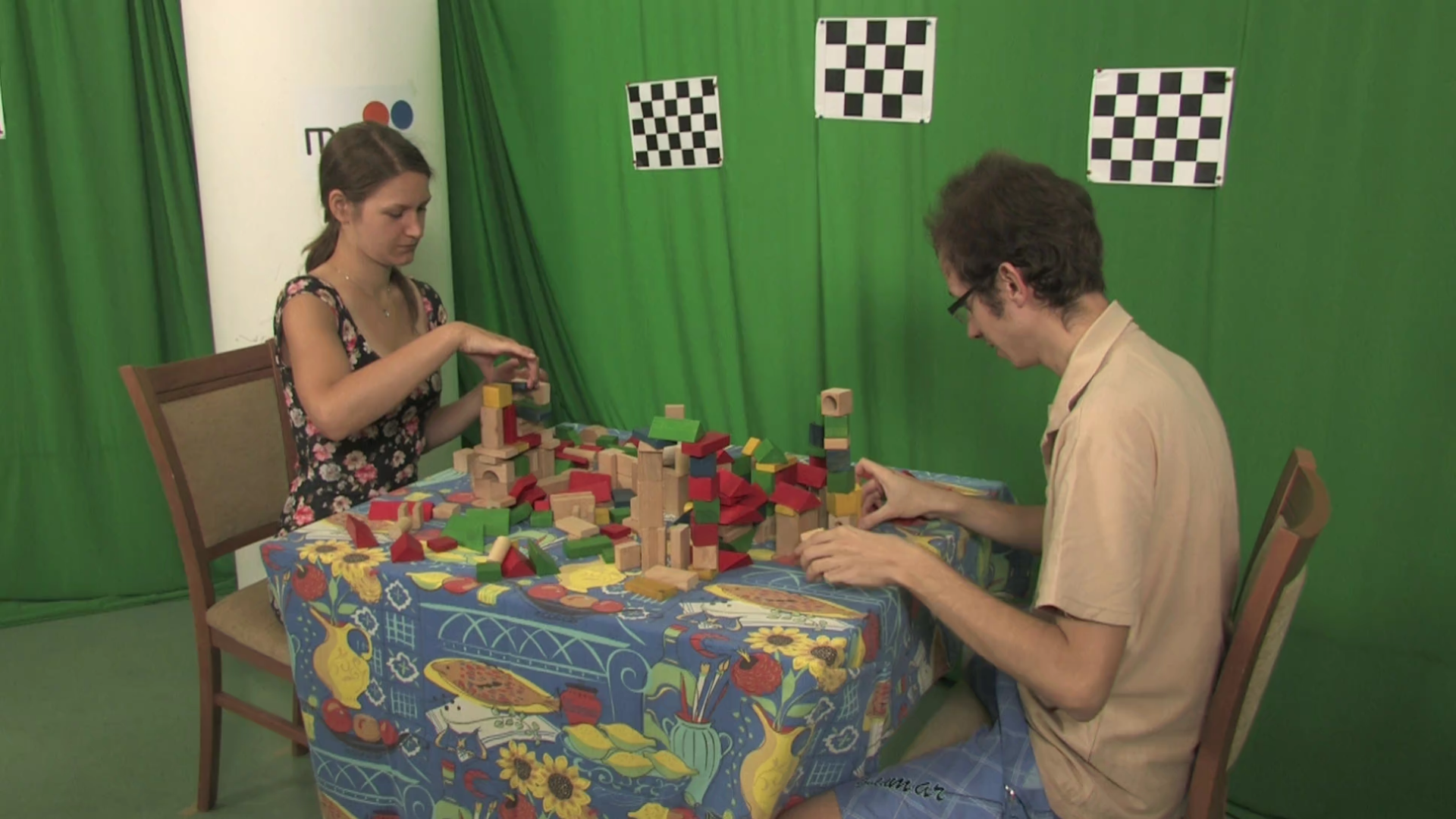}
		\centerline{(b). Block}
	\end{minipage}
	\begin{minipage}[b]{0.24\textwidth}
		\centering
		\includegraphics[width=\textwidth]{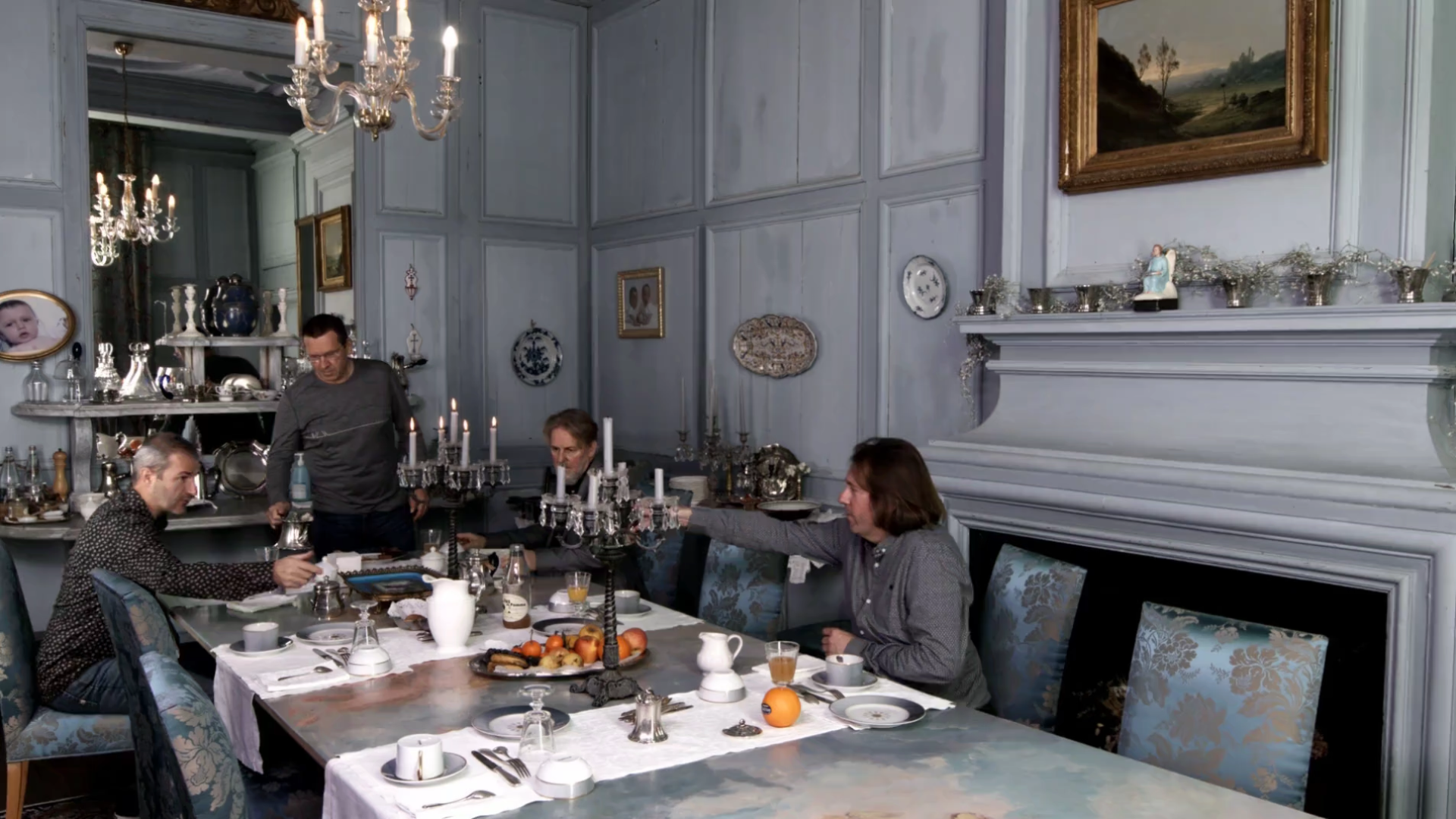}
		\centerline{(c). Breakfast}
	\end{minipage}
	\begin{minipage}[b]{0.24\textwidth}
		\centering
		\includegraphics[width=\textwidth]{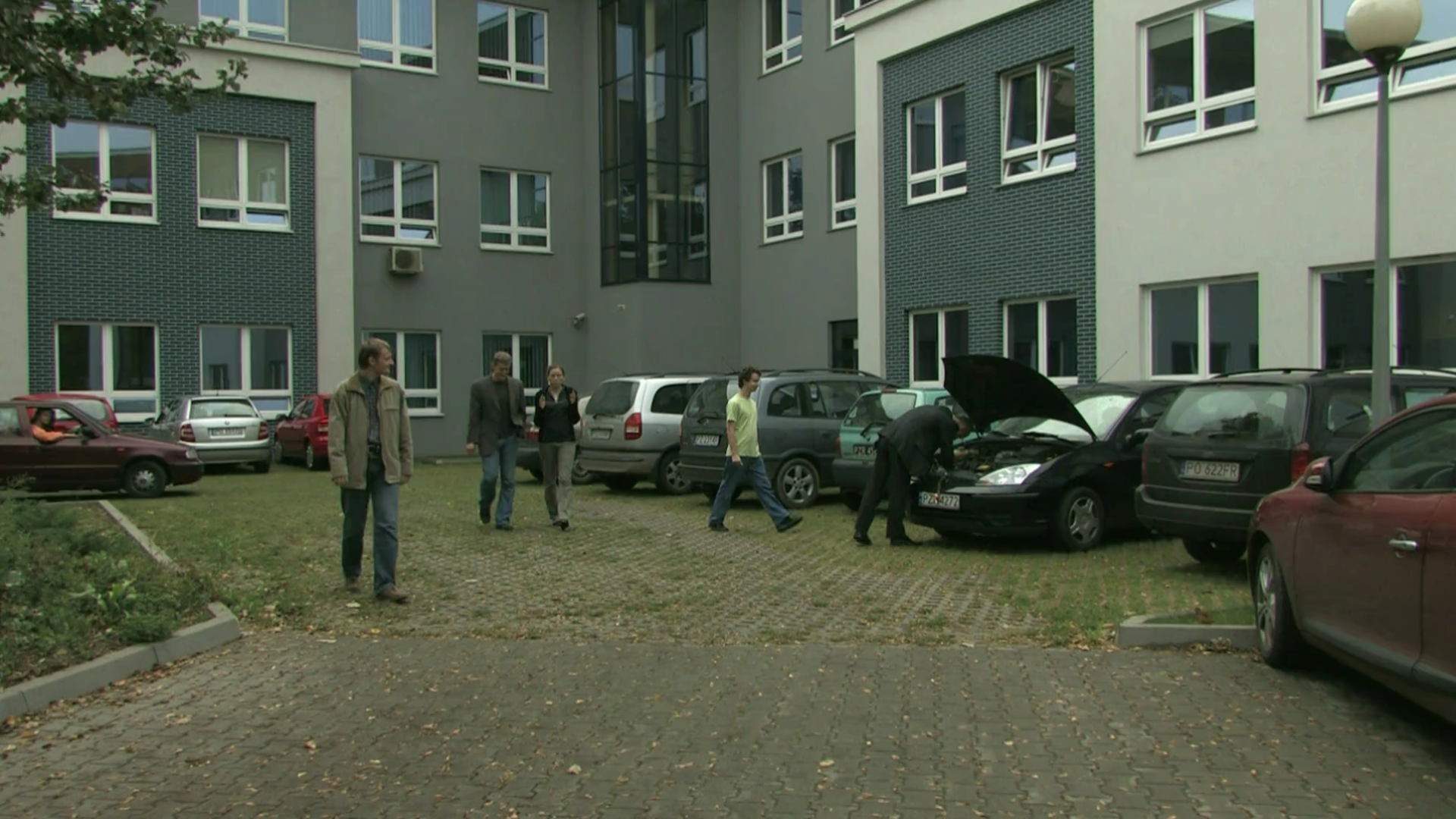}
		\centerline{(d). Carpark}
	\end{minipage}
	\begin{minipage}[b]{0.24\textwidth}
		\centering
		\includegraphics[width=\textwidth]{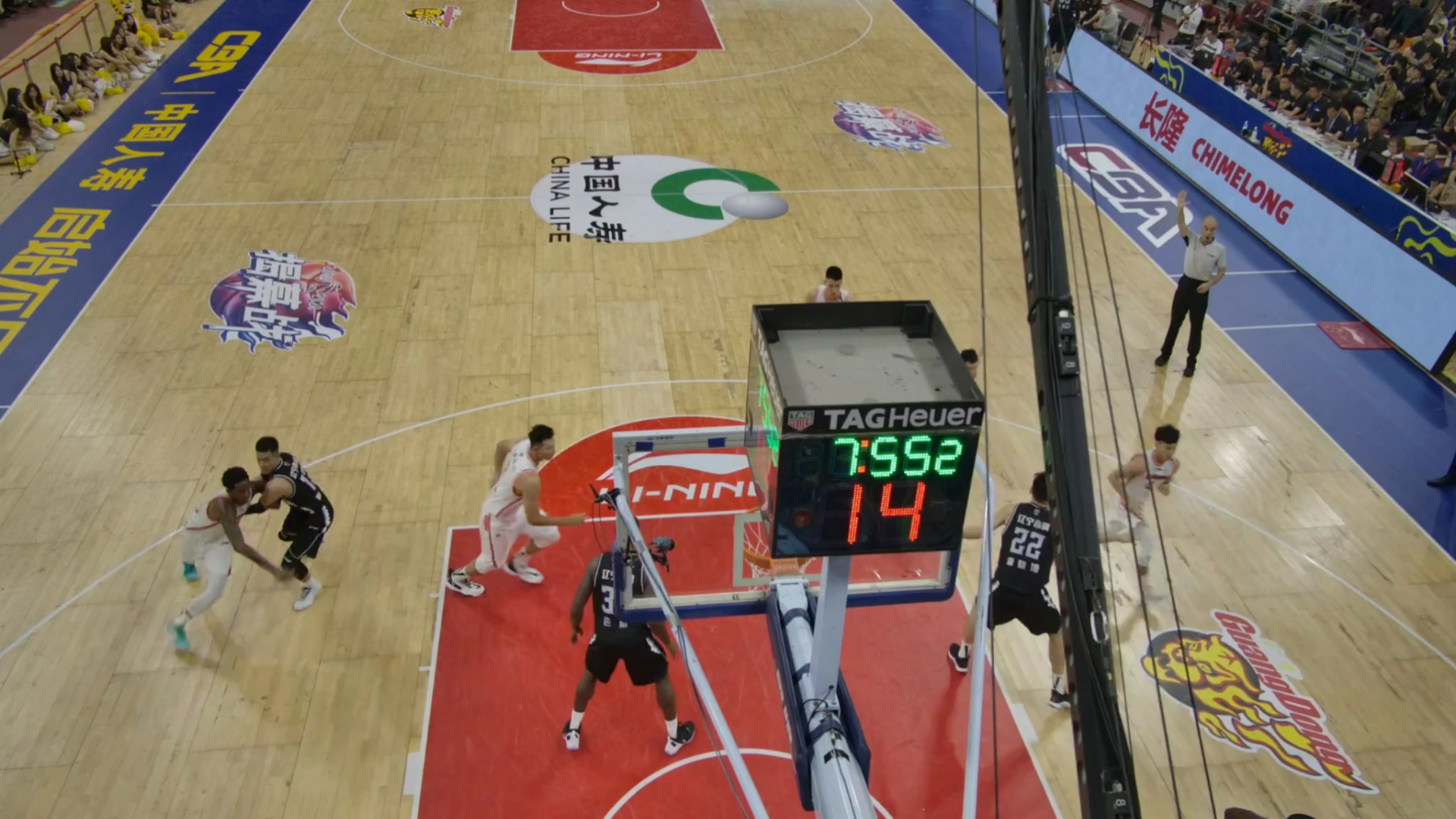}
		\centerline{(e). CBABasketball}
	\end{minipage}
	\begin{minipage}[b]{0.24\textwidth}
		\centering
		\includegraphics[width=\textwidth]{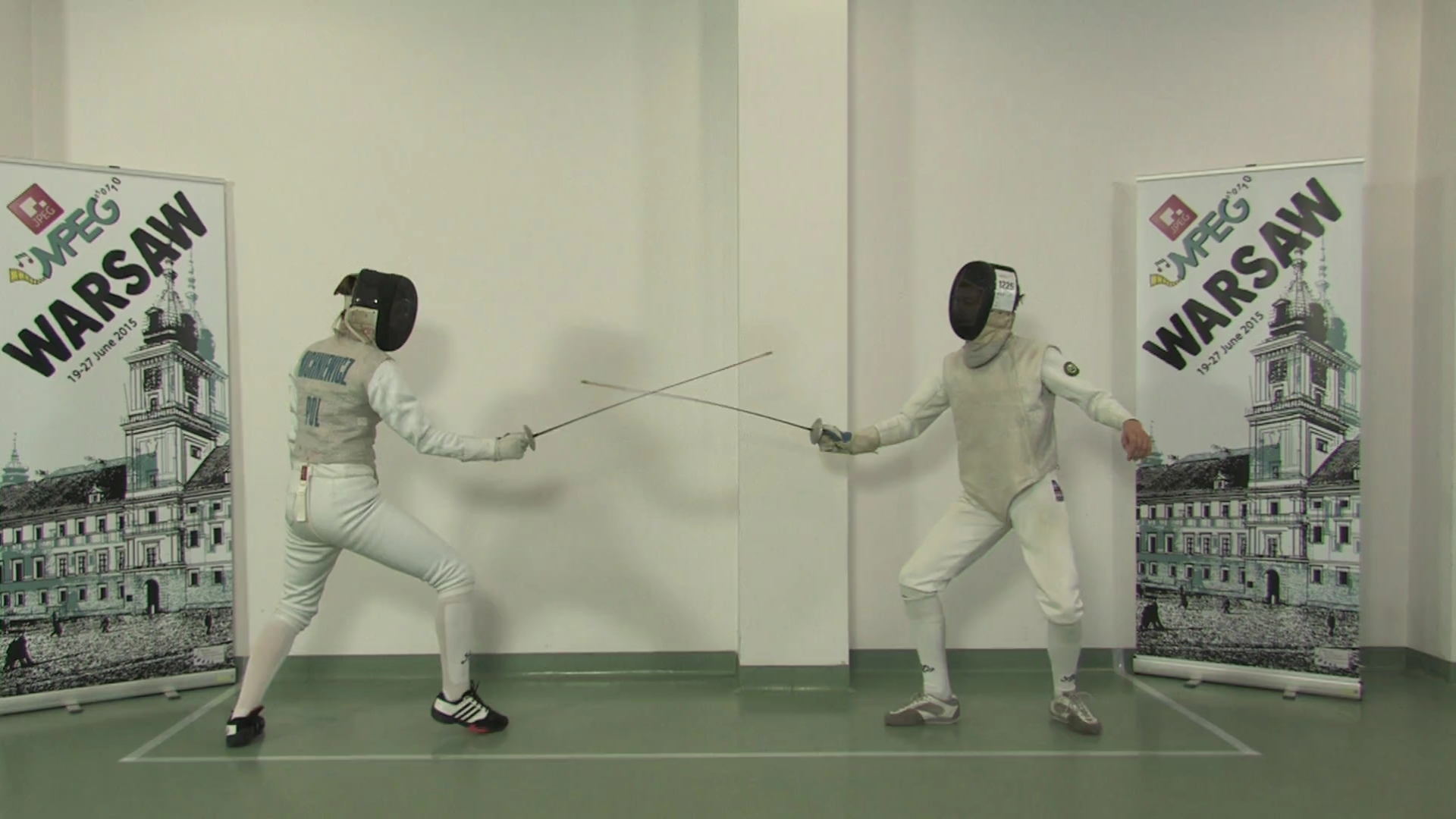}
		\centerline{(f). Fencing}
	\end{minipage}
	\begin{minipage}[b]{0.24\textwidth}
		\centering
		\includegraphics[width=\textwidth]{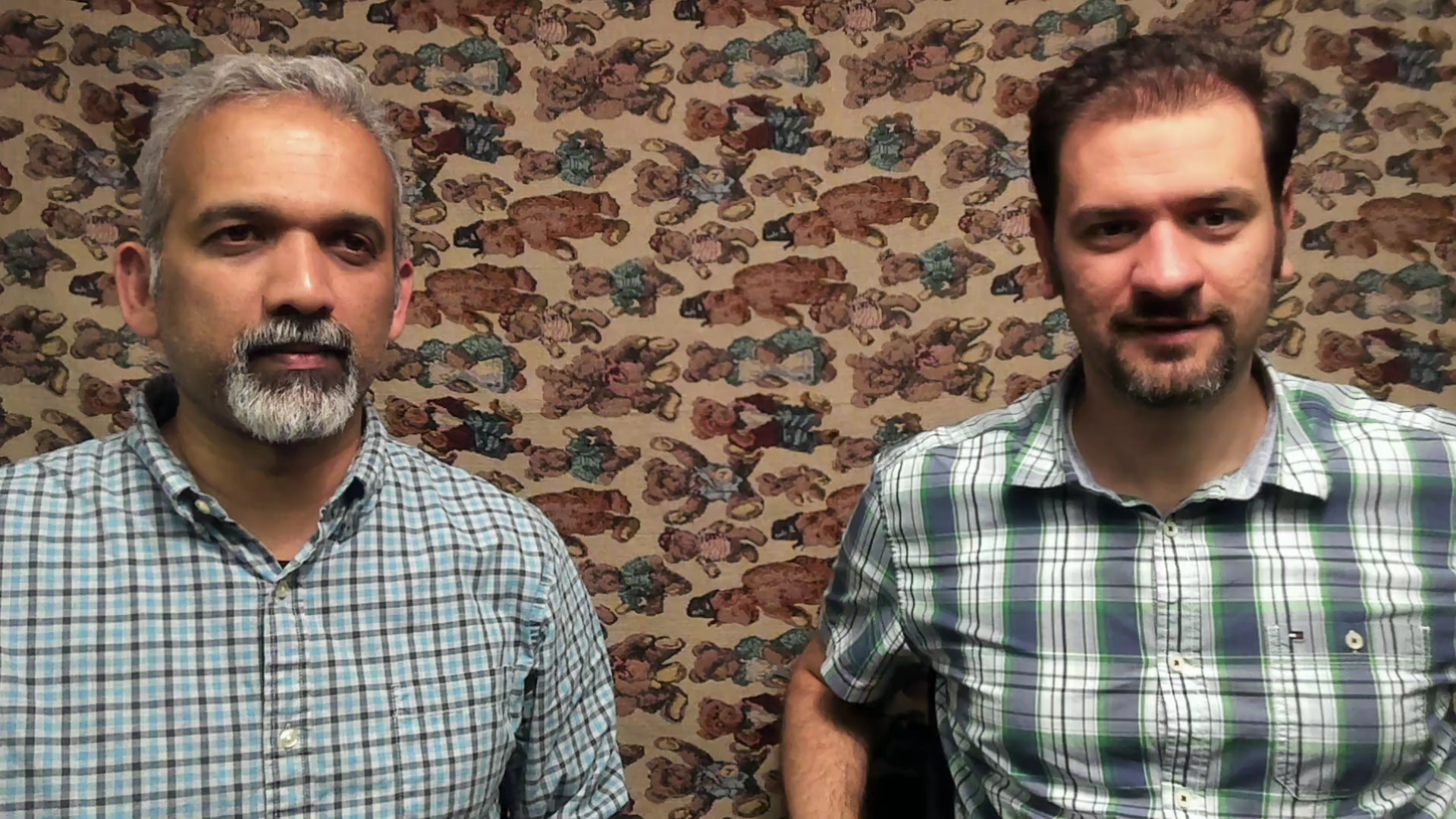}
		\centerline{(g). Frog}
	\end{minipage}
	\begin{minipage}[b]{0.24\textwidth}
		\centering
		\includegraphics[width=\textwidth]{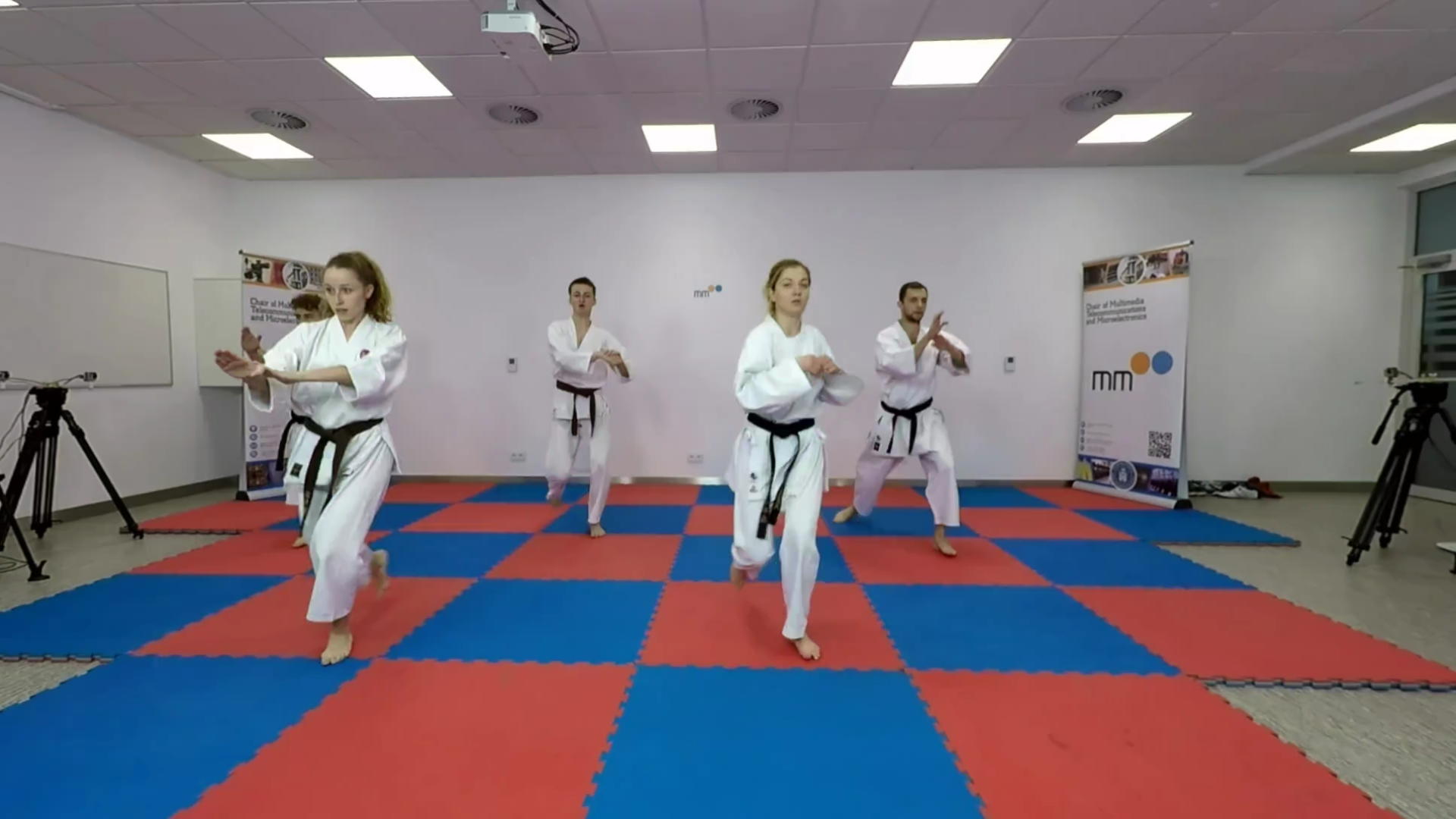}
		\centerline{(h). MartialArts}
	\end{minipage}
	\begin{minipage}[b]{0.24\textwidth}
		\centering
		\includegraphics[width=\textwidth]{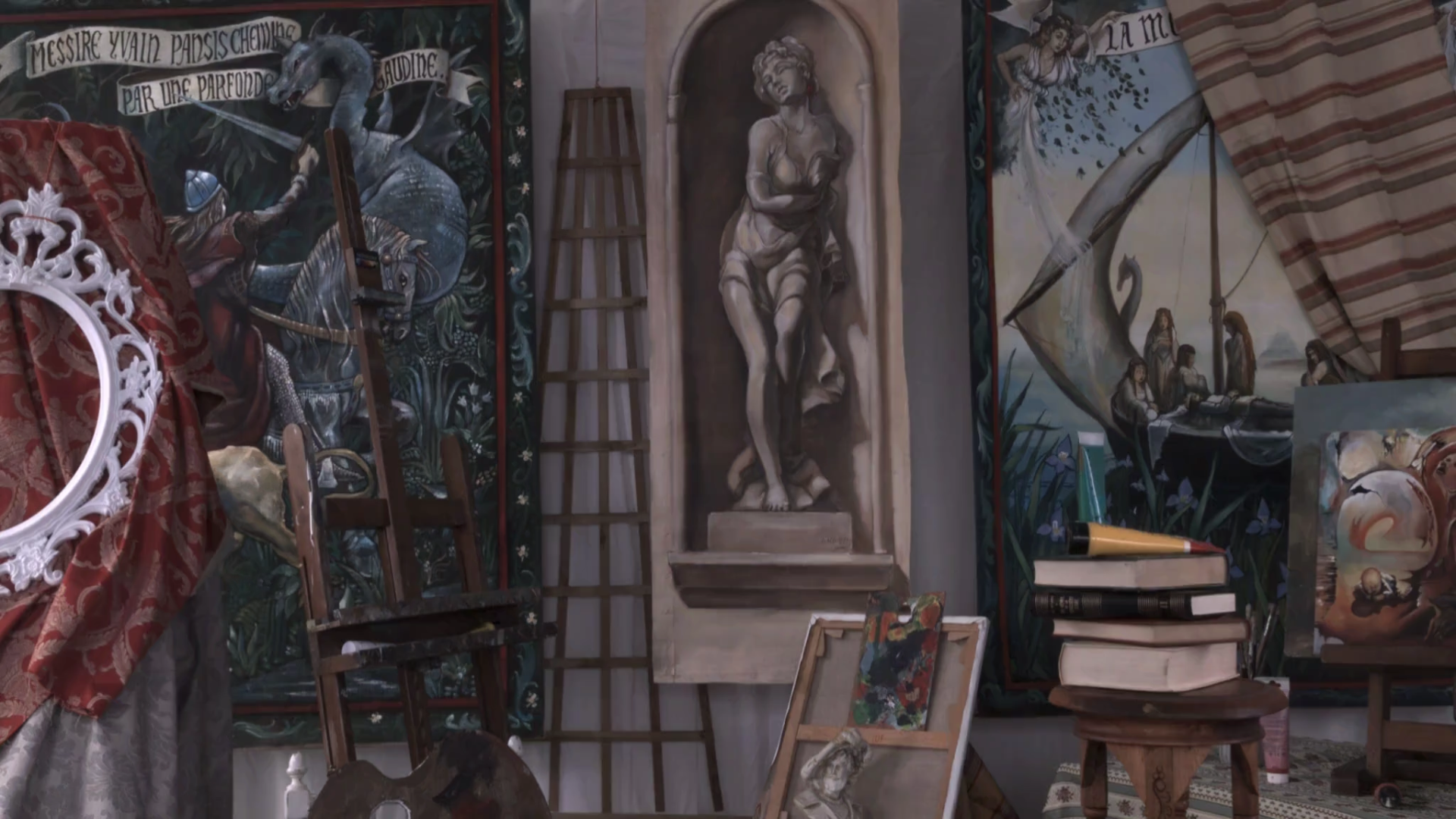}
		\centerline{(i). Painter}
	\end{minipage}
	\begin{minipage}[b]{0.24\textwidth}
		\centering
		\includegraphics[width=\textwidth]{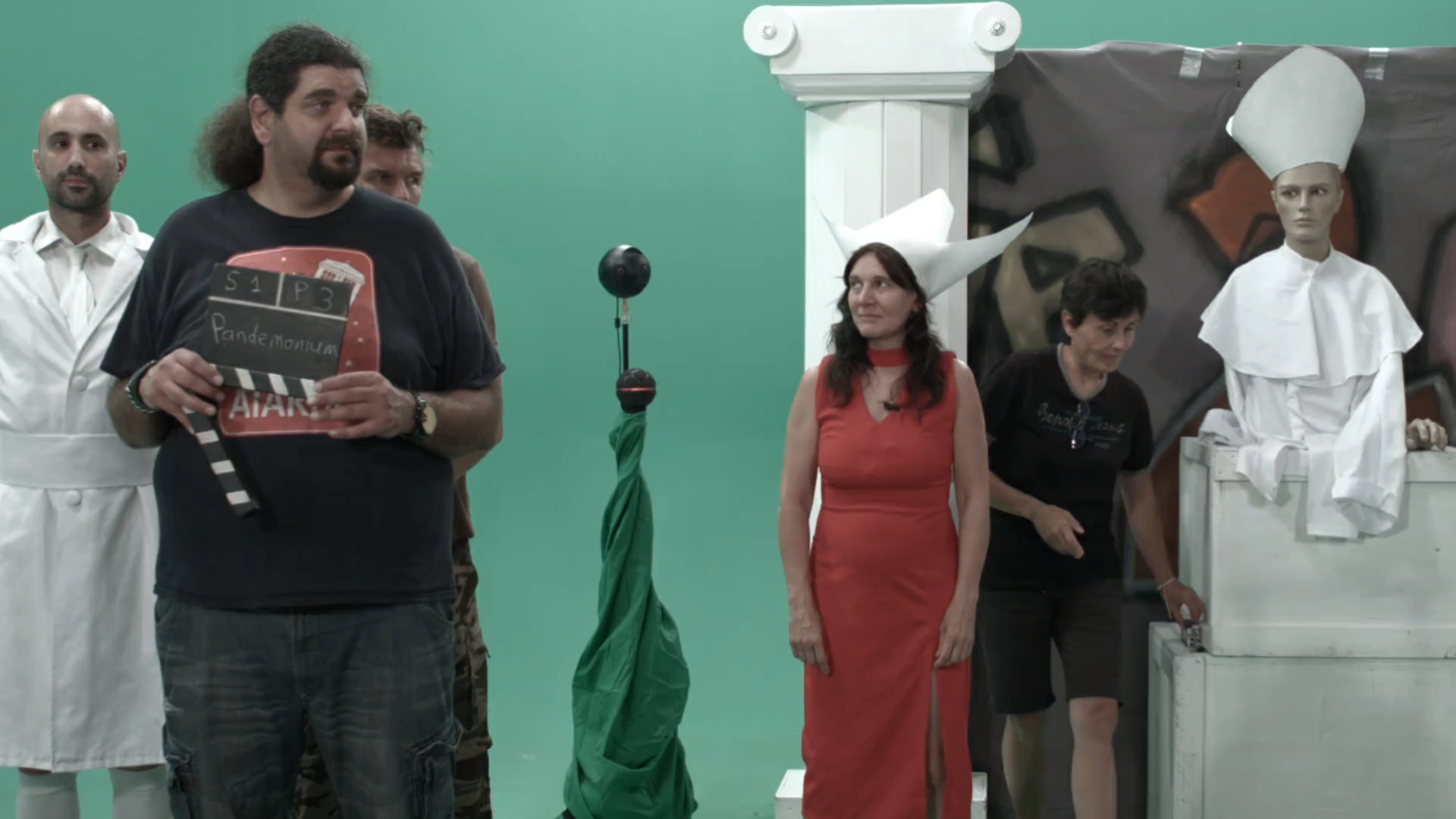}
		\centerline{(j). Pandemonium}
	\end{minipage}
	\begin{minipage}[b]{0.24\textwidth}
		\centering
		\includegraphics[width=\textwidth]{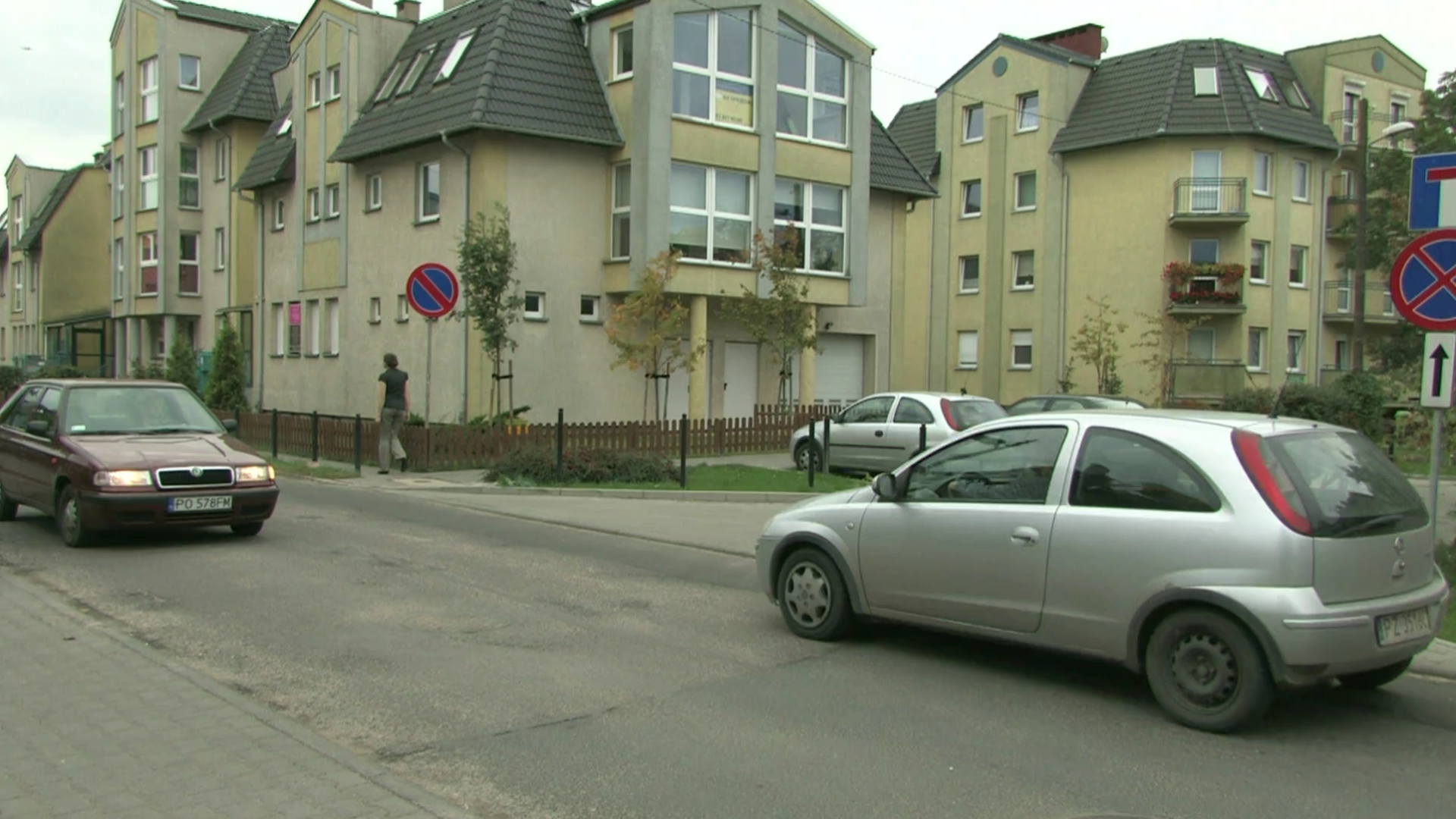}
		\centerline{(k). PoznanStreet}
	\end{minipage}
	\begin{minipage}[b]{0.24\textwidth}
		\centering
		\includegraphics[width=\textwidth]{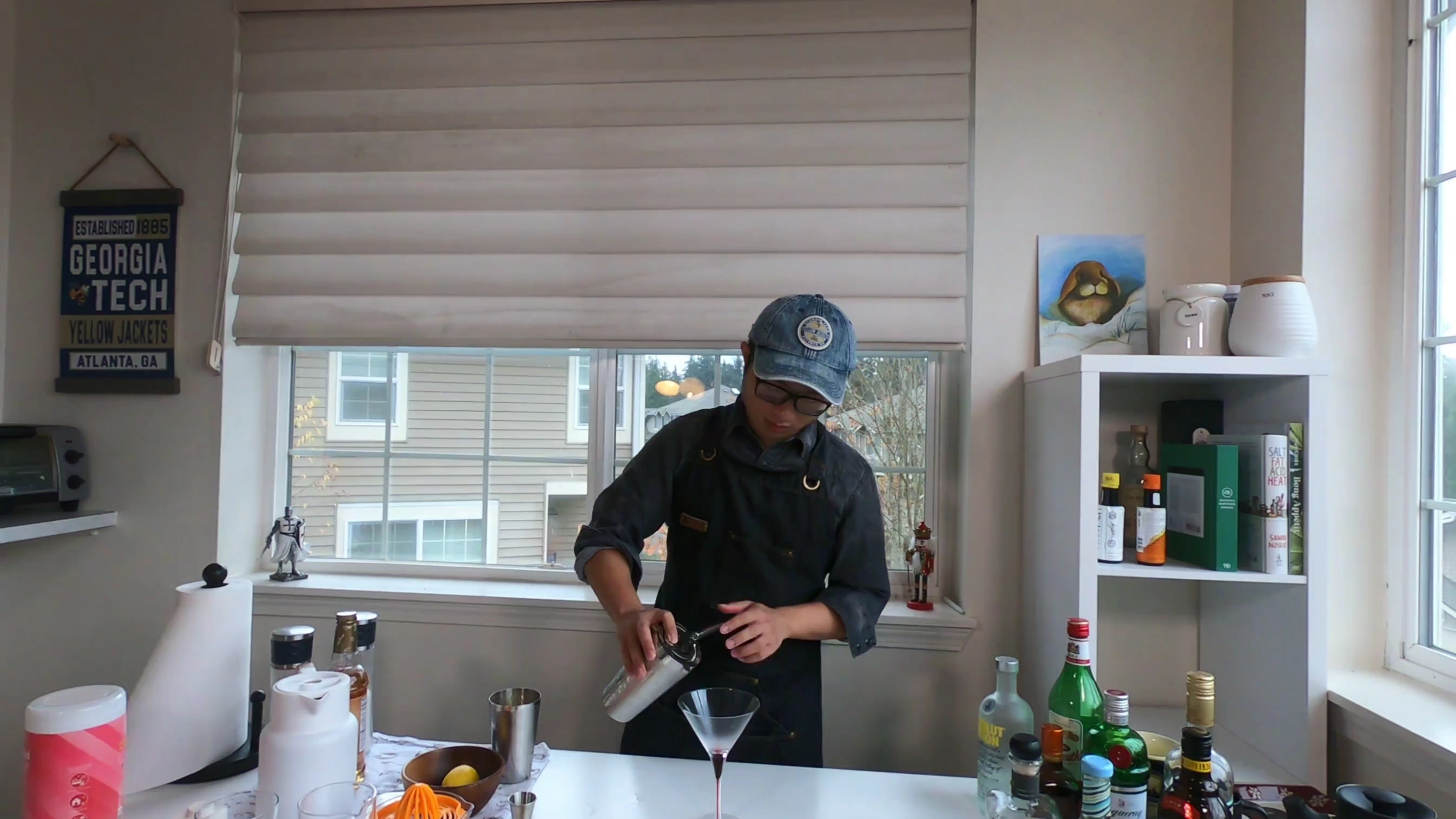}
		\centerline{(l). CoffeeMaritini}
	\end{minipage}
	\begin{minipage}[b]{0.24\textwidth}
		\centering
		\includegraphics[width=\textwidth]{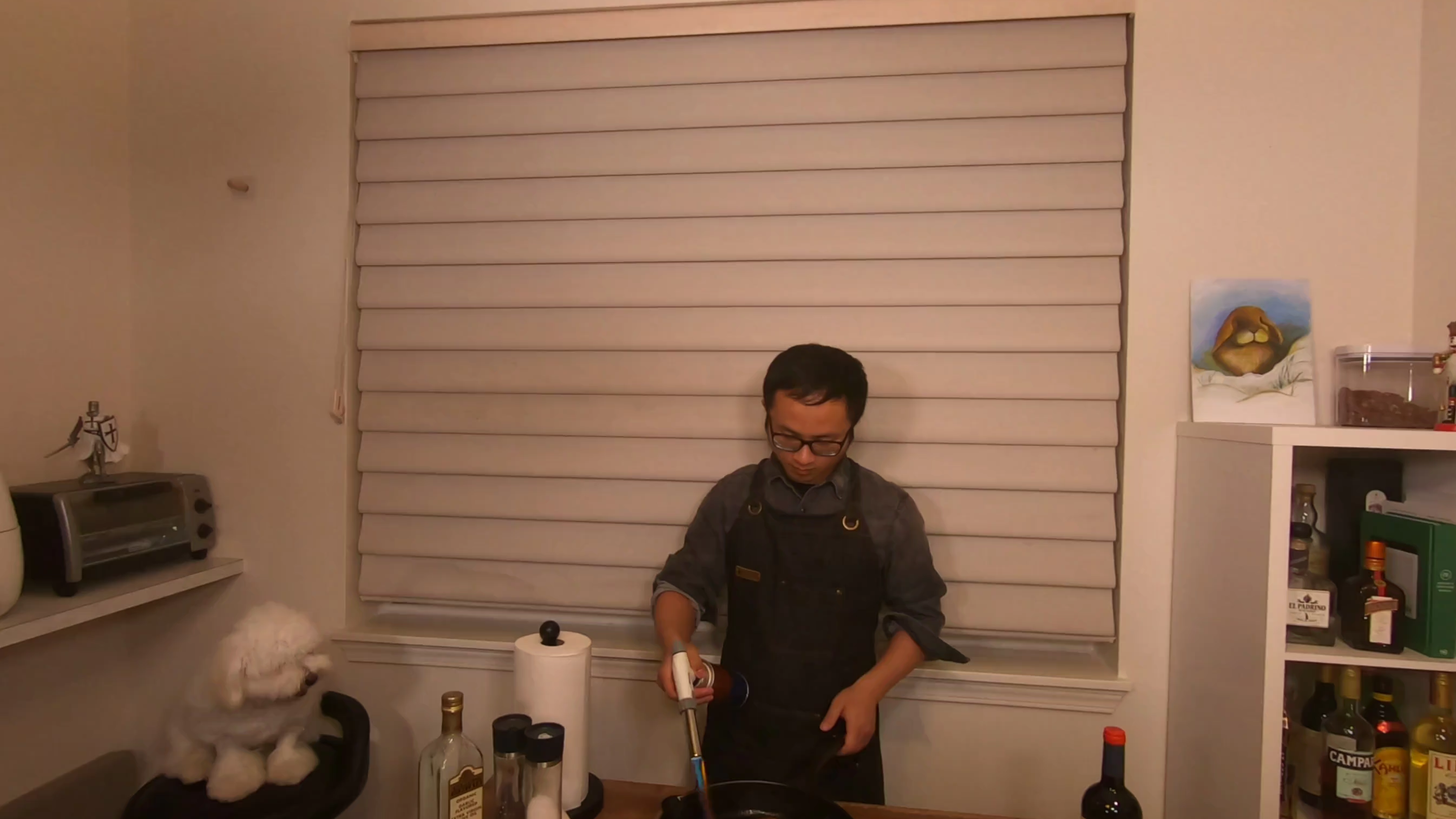}
		\centerline{(m). Flamesteak}
	\end{minipage}
	\caption{13 real-world SRCs used in our subjective experiment.}
	\label{fig:scenes}
\end{figure*}
\par Martin et al. developed both NeRF-QA \cite{martin2023nerf} and NeRF-VSQA \cite{martin2024nerfqa} subjective evaluation benchmark of NeRF-based methods. The second publication is a larger scale study. Their contributions are conducted on a small number of real sequences, completed with synthetic sequences.
\par Liang et al. \cite{liang2024perceptual} proposed a NVSQA database exclusively using front-facing PVS, containing 15 real-world static scenes with various materials, such as wood, marble, and glass. 
\par Onuoha et al. \cite{onuoha2023evaluation} employed a crowdsourcing approach to construct their database, enabling data collection under diverse conditions from a large pool of observers. The success of this method hinges on effectively screening out dishonest observers. By following the scenes and NVS methods of the NeRF-QA benchmark, it verified that conducting the experiment both in the lab and crowdsourcing yields comparable results.
\par Xing et al. \cite{xing2024explicit_nerf_qa} concentrate on quality assessment of the explicit NeRF-based methods. NVS methods can be divided into implicit and explicit types, based on whether the 3D shape of the object is explicitly stored. Explicit NVS methods have gained popularity since they reduce the computational load during the training and rendering process. 
\par Tabassum et al. \cite{tabassum2024quality} explored the effect of different moving paths. Results show that there is an obvious discrepancy in the scores while navigating the same scene along various moving paths. 
\par Yang et al. \cite{yang2024benchmark} developed a NVSQA benchmark database to demonstrate the sensitivity of different compression conditions to visual quality. This benchmark contains both static and dynamic scenes and adopts the GS method \cite{kerbl20233d} to synthesize PVS rather than NeRF-based methods. Contrary to previous contributions, some sequences are dynamic, but reconstruction is still applied on single frames. 
\par It's easy to observe a lack of studies comparing NeRF-based methods and GS-based methods. Meanwhile, the number of subjective evaluations of NeRF-based methods on real content is limited.

\begin{figure}[t]
	\centering
	\includegraphics[width=0.52\textwidth]{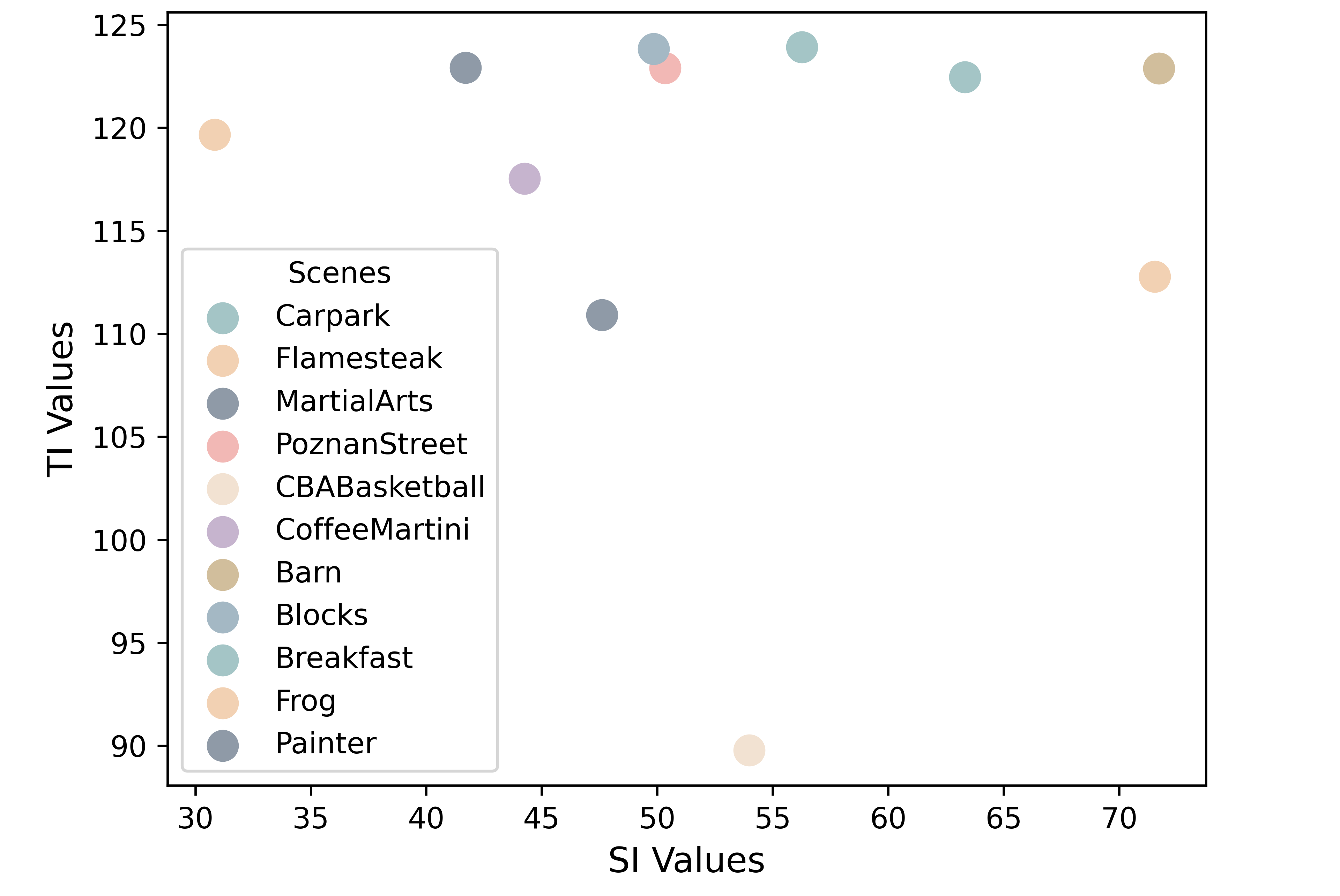}
	\caption{SI and TI of SRCs.}
	\label{fig:SITI}
\end{figure}
\begin{figure}[t]
	\centering
	\includegraphics[width=0.52\textwidth]{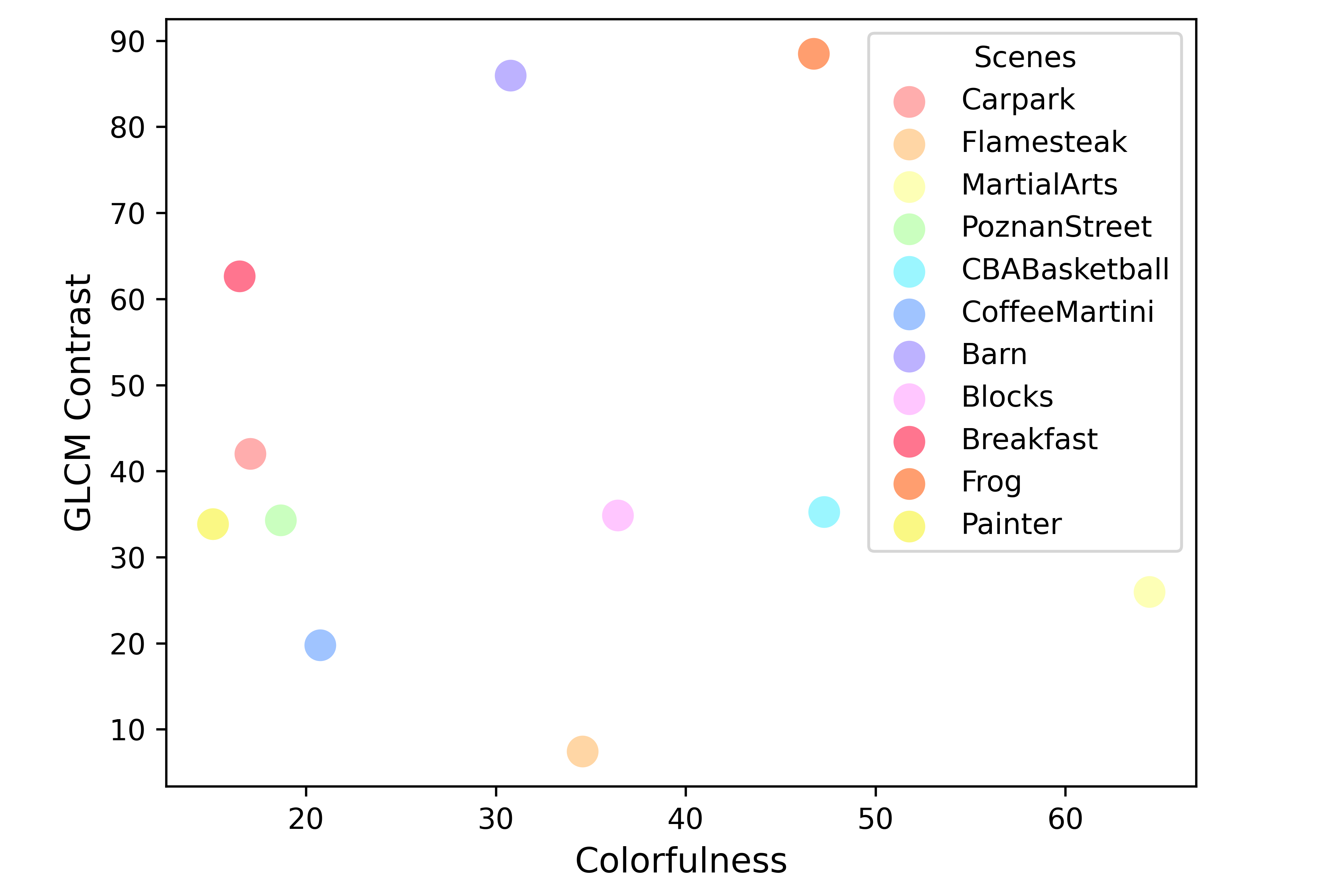}
	\caption{Colorfulness and GLCM Contrast of SRCs.}
	\label{fig:CC}
\end{figure}

\section{Proposed Benchmark}
\label{sec:3}
We contextualized our work in the previous section. The following sections will detail our subjective evaluation process, starting with the selection of SRCs in \ref{sec:3a}. We introduce the radiance field reconstruction methods we evaluate in \ref{sec:3b} as well as the virtual camera paths used for PVS in \ref{sec:3c}. Then, \ref{sec:3d} outlines the subjective evaluation procedure detailing the experimental setup, environment and participant screening. Finally, we highlight our unique contributions in \ref{sec:3e}.
\subsection{Source Sequences}
\label{sec:3a}
Source Sequences (SRCs) represent the scenes used in the subjective experiment, playing a significant role in the whole NVSQA benchmark database, as it directly affects the reliability and efficacy of quality assessment. To establish a high-quality benchmark, one of the keys is to cover diverse scenes to ensure the versatility of the method evaluation. Scene diversity consists of content diversity and style diversity. Content diversity mainly emphasizes the richness of local detail on both spatial and temporal. Style diversity indicates the visual style of the SRC. The richer the style, the stronger the visual impact. 
\par As illustrated in Fig. \ref{fig:scenes}, The selected SRCs contains 13 real-world scenes, which were captured from multiple cameras from different viewpoints synchronously. Specifically, the proposed dataset contain Barn\cite{tapie2021barn}, Block\cite{domanski2014blocks}, Breakfast\cite{tapie2021breakfast}, Carpark\cite{mielochmpeg}, CBABasketball\cite{sheng2021new}, Fencing\cite{domanski2016multiview}, Frog\cite{salahieh2018kermit}, MartialArts\cite{mieloch2023new}, Painter\cite{doyen2017light}, Pandemonium\cite{hobloss2021multi}, PoznanStreet\cite{mielochmpeg}, CoffeeMaritini\cite{Li_2022_CVPR}, and Flamesteak\cite{Li_2022_CVPR}. As shown in Fig. \ref{fig:scenes},  our SRCs cover a broad array of scene types, including, but not limited to, both indoor and outdoor settings, various shooting distances, and a wide range of lighting conditions from bright to dim, as well as color palettes from vibrant to muted.
\par Consequently, to present the content diversity within our SRCs, spatial information (SI) and temporal information (TI) of SRCs are calculated following ITU-T Rec. P.910 standard \cite{ITU2023} and their mean values are visualized as illustrated in Fig. \ref{fig:SITI}. SI is calculated based on the standard deviation of the Sobel-filtered frame, reflecting the spatial complexity of the image. TI is computed from the motion differences between consecutive frames, measuring the level of temporal change or motion in a video. Guidelines from the standard encourage interpretations of the subjective evaluations to be generalized to scenes featuring SI and TI in our SRCs range of values. Moreover, style diversity captured through measures of colorfulness \cite{amati2014study} and Gray Level Co-Occurrence Matrix (GLCM) contrast \cite{gadkari2004image} as shown in Fig. \ref{fig:CC}. Discrepancy in content and style across the SRCs demonstrates the diversity of this study.

\begin{table*}[t]
	\centering
	\renewcommand{\arraystretch}{1.6}
	\caption{Properties of HRCs used in subjective experiment.}

	\begin{tabular}{c|c|c|c|c|c}
		\hline
		\multirow{2}{*}{\diagbox[]{Property}{Method}} & \multicolumn{2}{c|}{NeRF-based} & \multicolumn{3}{c}{GS-based} \\
		\cline{2-6}
		& NeRFacto \cite{tancik2023nerfstudio} & K-Planes \cite{fridovich2023k}& GS \cite{kerbl20233d}& GS${\dag}$
		 \cite{kerbl20233d}&  STGFS 	\cite{li2024spacetime} \\
		\hline
				Static &\checkmark & \checkmark & \checkmark & \checkmark &\checkmark \\

		\hline
		Dynamic & & \checkmark & & & \checkmark \\
		\hline

		Explicit &  & \checkmark & \checkmark& \checkmark &\checkmark \\
		\hline
		Implicit & \checkmark& &   &  \\
		\hline
		Compression & &  &&  & \\
		\hline
		Full Training Iteration&\checkmark &\checkmark &\checkmark&  & \checkmark\\
		\hline
	\end{tabular}
		\label{tab:NVSmethod_comparison}
\end{table*}
\subsection{Hypothetical Reference Circuits}
\label{sec:3b}
To investigate the impact of the NVS methods of both NeRF-based and GS-based methods, 5 Hypothetical Reference Circuits (HRCs) are defined in the proposed database including a total of 5 NVS algorithms. Specifically, two representative NeRF-based methods (NeRFacto \cite{tancik2023nerfstudio} and K-planes\cite{fridovich2023k}) and three GS-based methods (GS \cite{kerbl20233d}, GS${\dag}$ \cite{kerbl20233d}, and STGFS \cite{li2024spacetime}) are employed. These methods have their own unique technical characteristics and their description are introduced in Section \ref{sec:2}. GS${\dag}$ share the same algorithm with the GS with less training iteration. Beyond comprehending the implementation of these methods, it is crucial to delve deeper into their inherent properties, which uncovers the reason for determining these methods as HRCs. As depicted in TABLE \ref{tab:NVSmethod_comparison}, properties of NVS methods contain Static, Dynamic, Explicit, Implicit, Compression, and Full Training Iteration, which are defined as follows.
\begin{itemize}
\item \textbf{Static/Dynamic:} indicates whether the model is capable of handling static or dynamic scenes.
\item \textbf{Explicit/Implicit:} describes whether the model adopts an explicit or implicit 3D representation.
\item \textbf{Compression:} denotes whether the model utilizes compressed operations for 3D representation.
\item \textbf{Full Training Iteration:} specifies whether the model use fewer training iterations compared to the original training setting.
\end{itemize}
\par TABLE \ref{tab:NVSmethod_comparison} demonstrates significant diversity in the HRCs, prompting exploration a various of interesting effects including but not limited to:
\par 1) Visual difference between the static and dynamic NVS methods, such as GS and STGFS. 
\par 2) Visual difference between the original model and its variant with fewer training iterations, such as GS and GS$\dagger$.
\par 3) Visual difference between NeRF-based and GS-based methods of the same type, such as K-Planes and STGFS.

\subsection{PVS generation}
\label{sec:3c}
Following the strategy of previous NVSQA benchmark databases \cite{martin2023nerf, martin2024nerfqa, liang2024perceptual, onuoha2023evaluation, xing2024explicit_nerf_qa, tabassum2024quality}, the problem of assessing 3D representation quality can be effectively transferred into a video quality assessment problem. As summarized in Fig.\ref{fig:subjective_evaluation_methodology}, PVSs generated by HRCs are viewed directly and scored by observers. As detailed in Section \ref{sec:2}, our PVSs contain three types of virtual camera paths, 360°, front-facing view, and single-viewpoint videos. The 360° virtual camera paths follow a movement around the scene content at a large angle and it generally but not strictly will be 180° or 360° around the object. Front-facing virtual camera paths denote that the camera captures by moving in front of the object at a small angle. Single-viewpoint virtual camera paths allow the camera to capture images from a single viewpoint. From a camera capture perspective, 360° and front-facing videos are considered multi-view visual paths while single-view video is captured by a single viewpoint path. 
\par As mentioned in Section \ref{sec:1}, we designed two subjective experiments to explore the impacts of multi-view and single-view virtual camera paths. Thus, for multi-view subjective evaluation, we follow the moving paths in previous studies, and multi-view videos are either 360° or front-facing videos. For single-view subjective evaluation, we use the middle viewpoint of all the viewpoints as the viewpoint of the single-view video and generate static PVS for all HRCs. TABLE \ref{tab:PVS} depicts the types of PVSs, which consist of 7 360° and 6 front-facing multi-view PVSs.

\begin{table}[t]
	\centering
	\caption{Types of the PVSs.}
	\renewcommand{\arraystretch}{1.2}
	\begin{tabular}{c|c|c|c}
		\hline
		\multirow{2}{*}{\textbf{Sequence}}  & \multicolumn{2}{c|}{\textbf{Multi-view}} & \multirow{2}{*}{\textbf{Single-view}} \\ \cline{2-3}
		& \textbf{360°} & \textbf{Front-facing} & \\ \hline
		Barn \cite{tapie2021barn} & & \checkmark & \checkmark \\ \hline
		Block \cite{domanski2014blocks} & \checkmark &  & \checkmark \\ \hline
		Breakfast \cite{tapie2021breakfast} &  & \checkmark & \checkmark \\ \hline
		Carpark \cite{mielochmpeg} &  & \checkmark & \checkmark \\ \hline
		CBABasketball \cite{sheng2021new} & \checkmark &  & \checkmark \\ \hline
		Fencing \cite{domanski2016multiview} & \checkmark &  & \checkmark \\ \hline
		Frog \cite{salahieh2018kermit} &  & \checkmark & \checkmark \\ \hline
		MartialArts \cite{mieloch2023new} & \checkmark & & \checkmark \\ \hline
		Painter \cite{doyen2017light} &  & \checkmark & \checkmark \\ \hline
		Pandemonium \cite{hobloss2021multi} & \checkmark &  & \checkmark \\ \hline
		PoznanStreet \cite{mielochmpeg} &  & \checkmark & \checkmark \\ \hline
		CoffeeMaritini \cite{Li_2022_CVPR} & \checkmark & & \checkmark \\ \hline
		Flamesteak \cite{Li_2022_CVPR} & \checkmark &  & \checkmark \\ \hline
	\end{tabular}
	\label{tab:PVS}
\end{table}
\subsection{Subjective Quality Assessment Methodology}
\label{sec:3d}
Given source sequences and PVSs, the concern turns to the methodology of subjective experiments, including experimental procedure, experimental environment, and participants. 
\begin{figure}[t]
	\centering
	\includegraphics[width=0.48\textwidth]{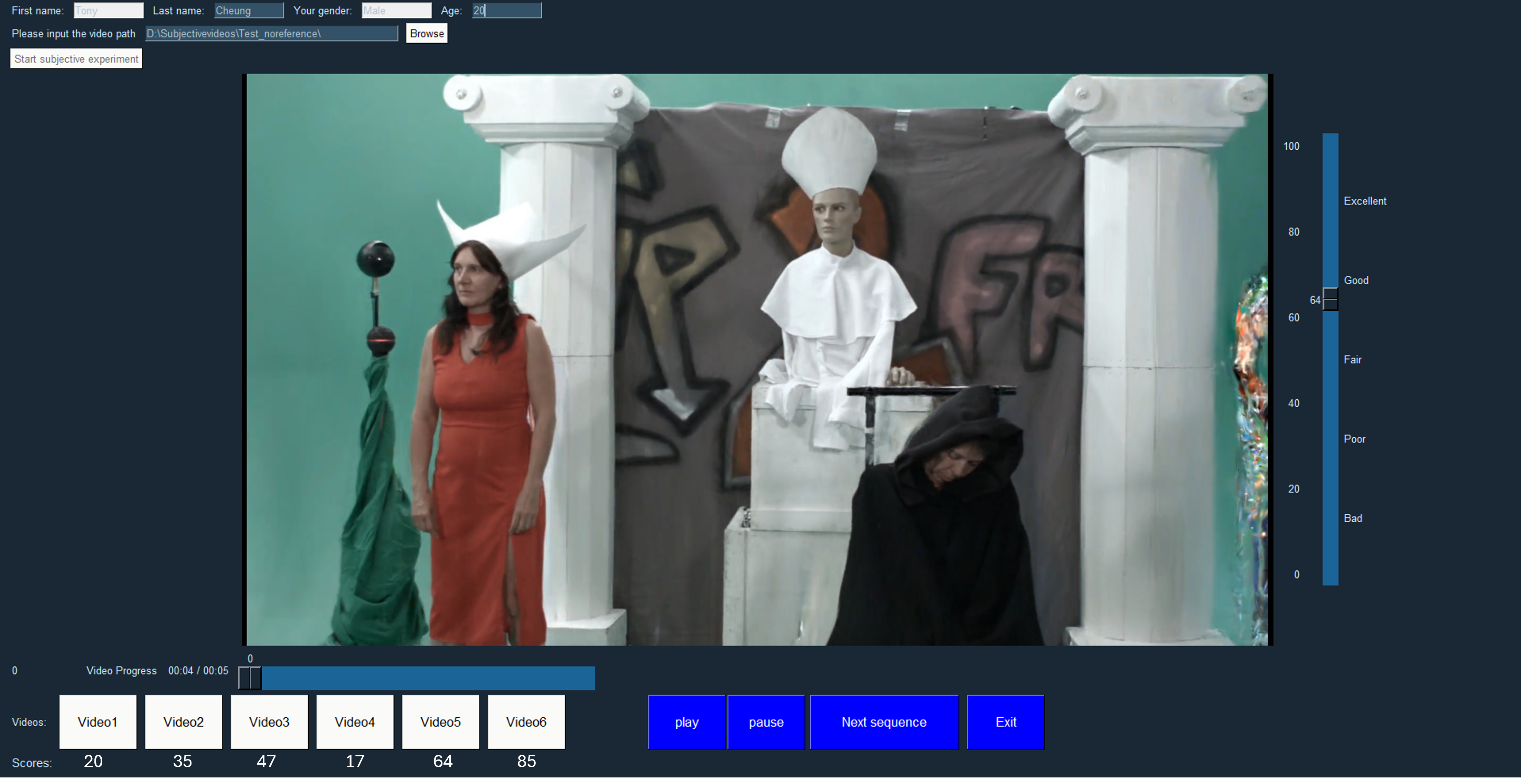}
	\caption{The GUI used in the subjective experiment.}
	\label{fig:GUI}
\end{figure}
\subsubsection{{Experiment Procedure}}
As outlined in TABLE \ref{tab:datasetcomp}, several testing protocols are employed in previous databases, such as PC, DSIS, and ACR that are recommended in ITU-T Rec. P.910 \cite{ITU2023} standard. To offer the observer a more flexible and natural viewing experience, our subjective experiment employed the SAMVIQ protocol, also recommended in ITU-T Rec. BT.1788 \cite{ITU2023_1788} standard. In this protocol, the observer first watches the PVSs generated by different HRCs in turn, without the option to skip and pause. Subsequently, the observer can freely choose to review videos repeatedly as they prefer, with the option to skip, pause, and frame selection. Finally, the observer is required to give integer scores from 0 to 100 for all videos of one sequence and move on to the next sequence. This process has significant flexibility and allows the observer to rate PVSs according to their viewing preferences. To ensure a comprehensive data capture and show perfectly the PVSs on the display screen, we developed a SAMVIQ tool, with many functions such as information recording, rating, video selection, and progress adjusting. Fig \ref{fig:GUI} shows the GUI for the no-reference multi-view test. For the referenced single-view test, the GUI has an additional 'reference' button to watch the reference video. 
\par Before the formal subjective test, there are preliminary tests for the participant. First, the observers must successfully pass both the Snellen chart test and the Ishihara chart test following SAMVIQ \cite{blin2006new}, to confirm normal visual acuity and color vision, respectively. Following these tests, we provide one sequence from another scenario different from the proposed SRCs, and a training session to familiarize observers with the SAMVIQ tool, ensuring a smooth evaluation process. Then, the formal test is conducted.

\begin{figure}[t]
	\centering
	\includegraphics[width=0.48\textwidth]{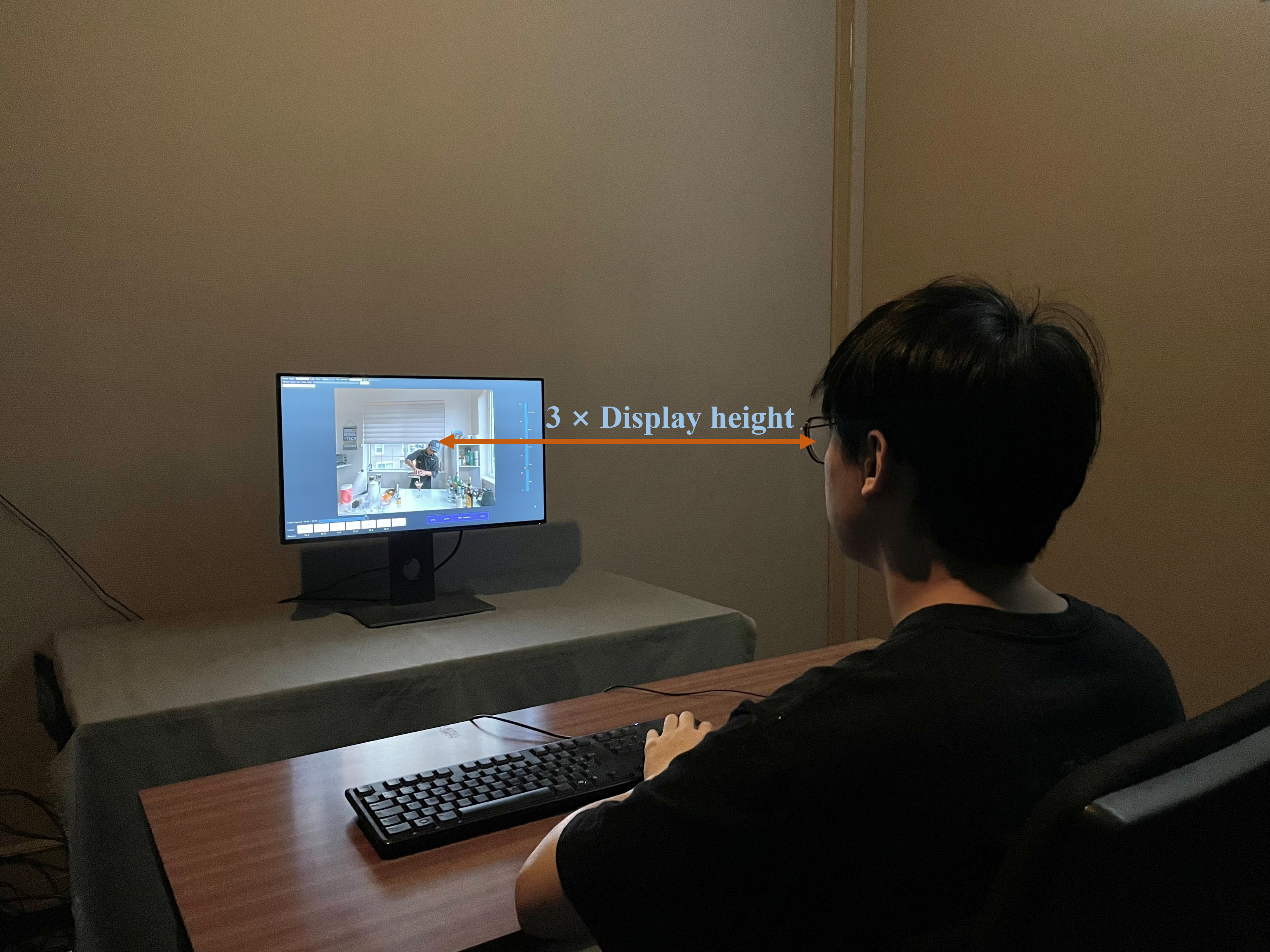}
	\caption{Experimental environment.}
	\label{fig:EE}
\end{figure}
\subsubsection{Experimental Environment}
In order to minimize the impact of uncontrollable factors on subjective experiments, we strictly follow the environmental setting of ITU-T Rec. P.910 \cite{ITU2023}. The subjective test is carried out in a dark room with appropriate lighting. All videos were displayed on a 27'' DELL U2717D monitor with a viewing distance of 1m, which is 3 times the screen height, see Fig. \ref{fig:EE}.
\subsubsection{Participants}
34 participants took part in the two subjective tests, with an age range of 20 to 50, including 18 males and 16 females. Before the formal test, all participants view the experimental instructions and sign a consent form. Meanwhile, they received the reward from this subjective test. Subsequently, following by conducting the whole subjective experiment, a post-screening procedure was conducted to discard dishonest participants. Correlation-based post-screening methodology recommended in ITU Ref. BT.500-15 standard \cite{ITU2023_500} was employed. Finally, 29 participants were retained.
\subsubsection{Experimental Setting}
\begin{figure}[t]
	\centering
	\includegraphics[width=0.48\textwidth]{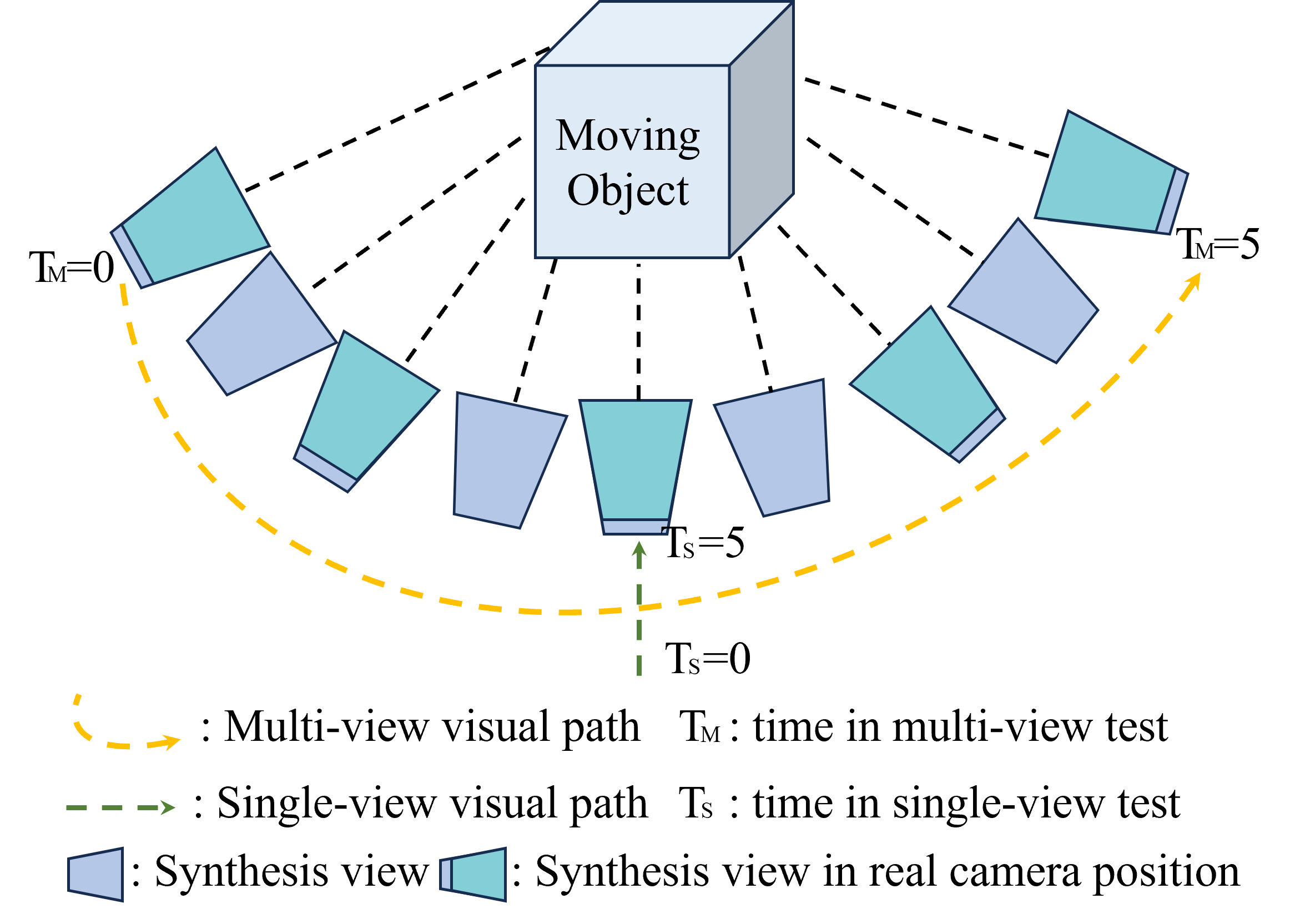}
	\caption{The illustration of the camera path.}
	\label{fig:CP}
\end{figure}

We devised two subjective tests to explore the impact of multi-view and single-view visual pathways, including a no-reference multi-view path evaluation and a referenced single-view path evaluation. For the no-reference multi-view test, since the images of scenes were captured by multiple cameras from discrete viewpoints without a continuous moving path, the multi-view test cannot provide a reference. For the referenced single-view test, the video from the central camera view is synthesized and compared with that captured by the original camera. Therefore, the NVS models are trained by using data from all views besides the central view. To fully comprehend these two tests, the camera paths over time of two subjective tests are illustrated in Fig. \ref{fig:CP}. All NVS models were trained using the default parameter as the published paper, where the GS$\dagger$ model was trained in 7000 iterations. Finally, we got 130 stimuli ((5 for the single-view test + 5 for the multi-view test) $\times$ 13 SRCs). Each stimulus lasts 5 seconds, where the frame rate of the video equals the original frame rate (25 or 30 frames per second (FPS))or half frame rate (only for the SRCs with 60 FPS) of the camera.
\par For the formal test, the no-reference multi-view test is conducted first followed by the referenced single-view test, rather than converse to avoid the effect of reference videos for the no-reference multi-view test. To alleviate visual fatigue, the first test is divided into two parts, with a mandatory five-minute rest after each part. After that, participants will finish the second test without rest, as they are already familiar with the scenes from the first test, and it's easier to rate because the second part includes reference videos.

\begin{figure}[t]
	\centering
	\includegraphics[width=0.48\textwidth]{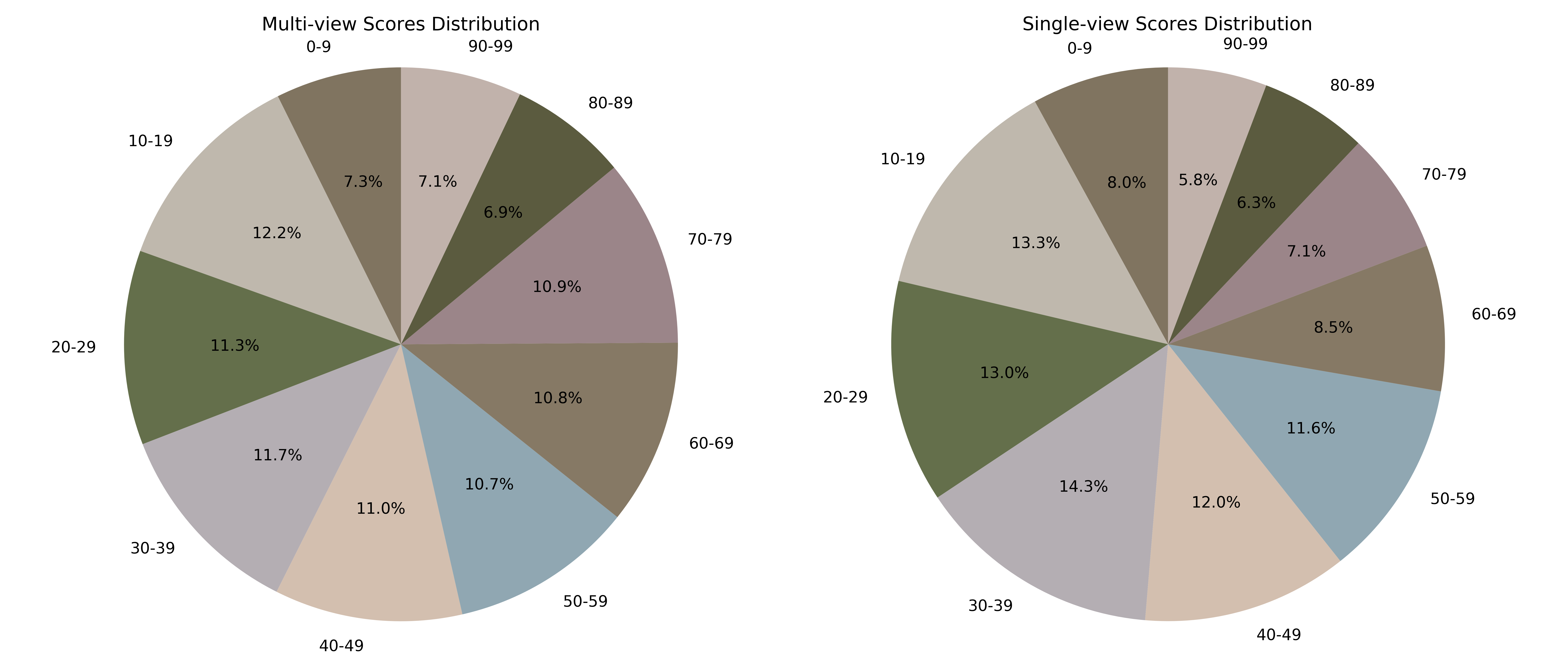}
	\caption{Distribution of all scores in the proposed database.}
	\label{fig:Frequency}
\end{figure}

\begin{figure*}[h!] 
	\centering
	\begin{minipage}[b]{0.19\textwidth}
		\centering
		\includegraphics[width=\textwidth]{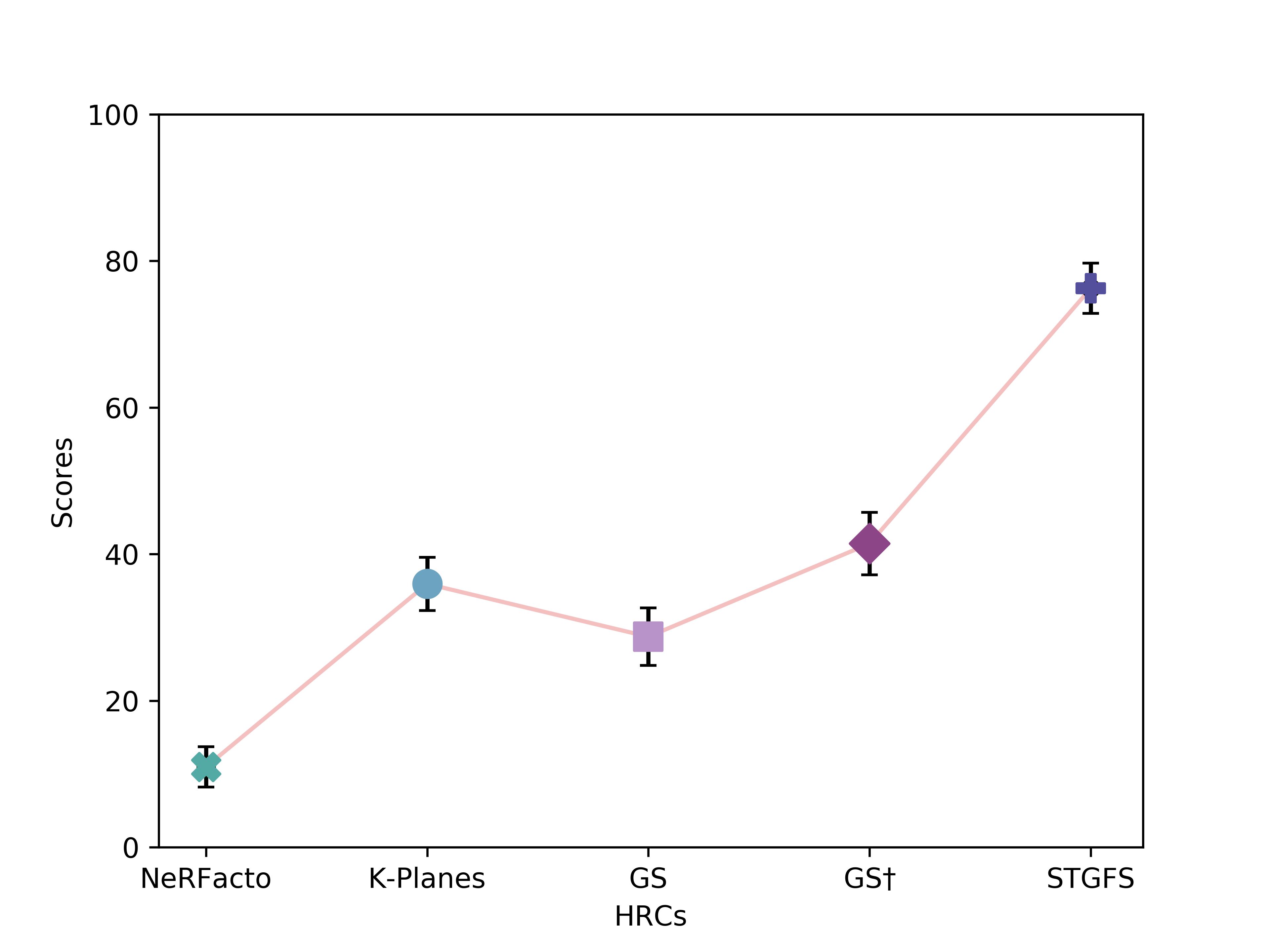}
		\centerline{(a). Barn}
	\end{minipage}
	\begin{minipage}[b]{0.19\textwidth}
		\centering
		\includegraphics[width=\textwidth]{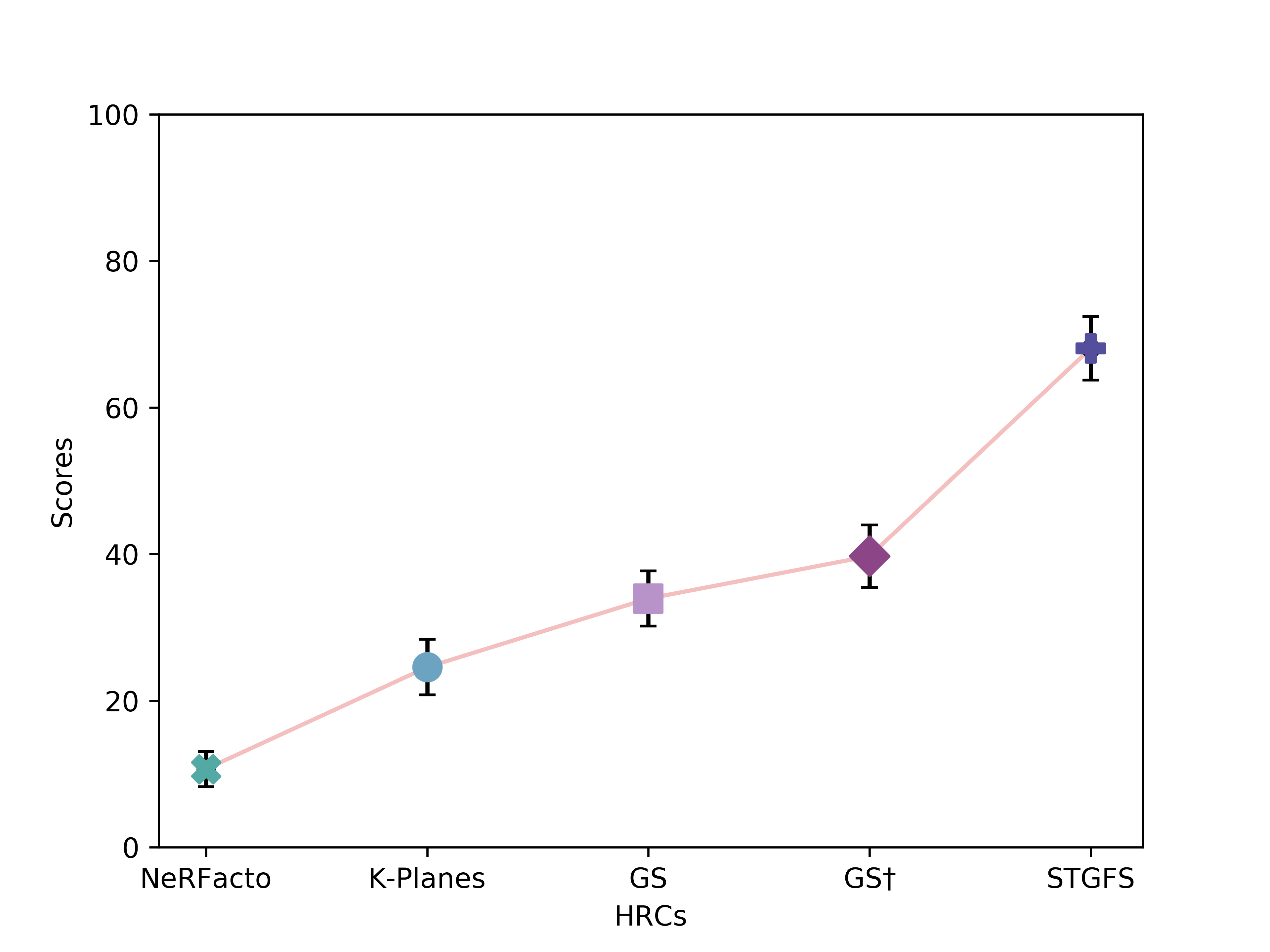}
		\centerline{(b). Blocks}
	\end{minipage}
	\begin{minipage}[b]{0.19\textwidth}
		\centering
		\includegraphics[width=\textwidth]{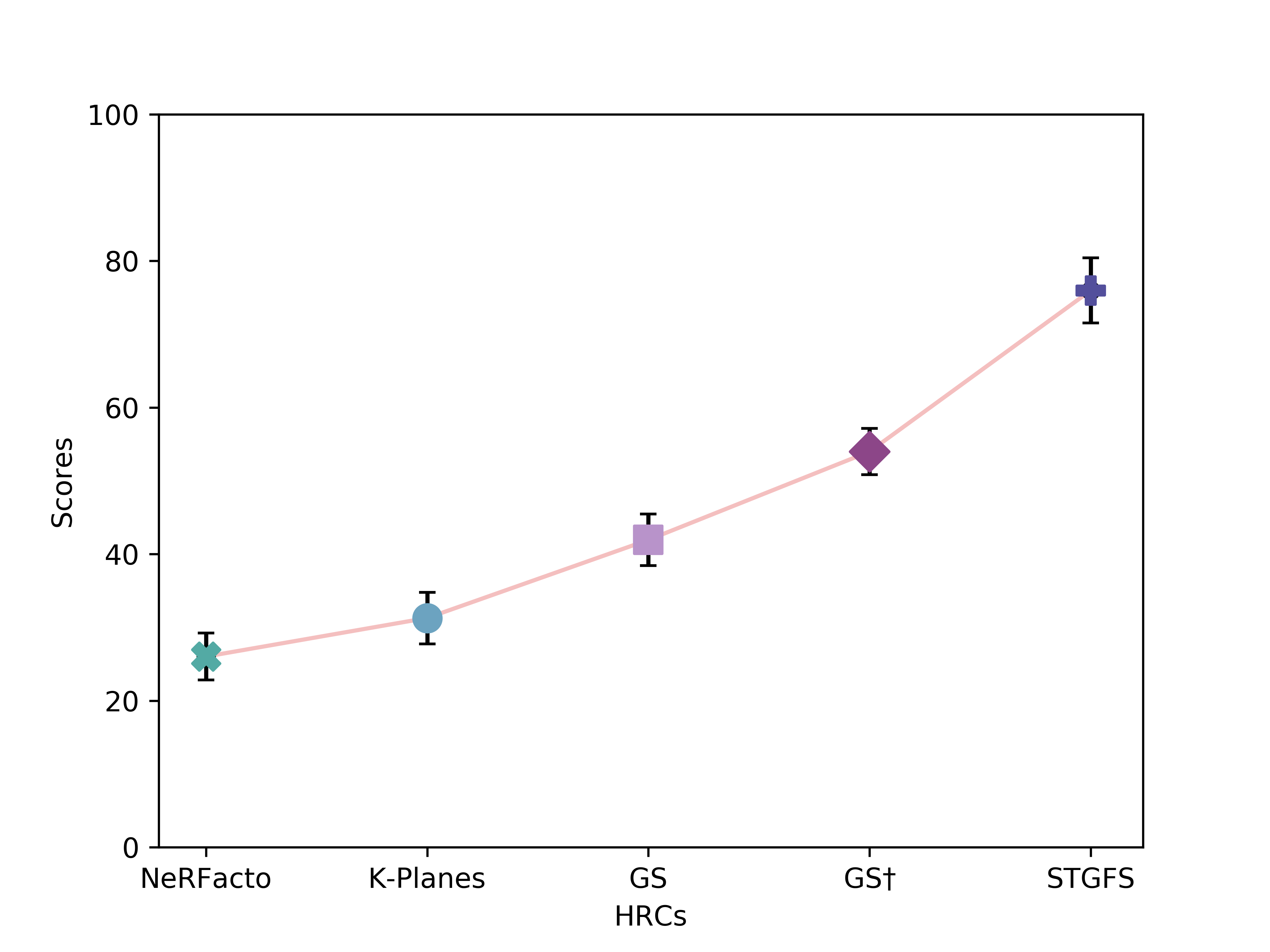}
		\centerline{(c). Breakfast}
	\end{minipage}
	\begin{minipage}[b]{0.19\textwidth}
		\centering
		\includegraphics[width=\textwidth]{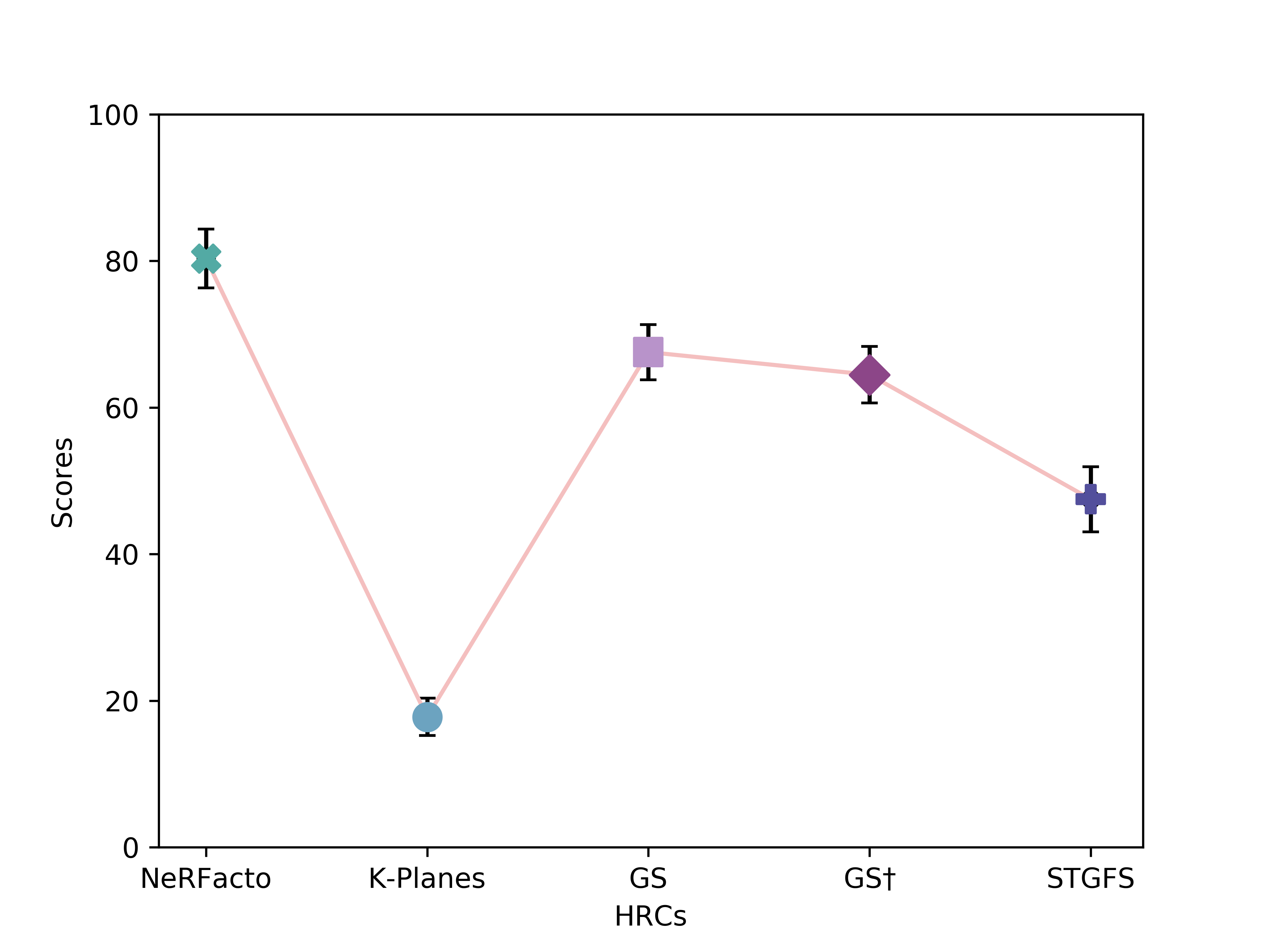}
		\centerline{(d). Carpark}
	\end{minipage}
	\begin{minipage}[b]{0.19\textwidth}
		\centering
		\includegraphics[width=\textwidth]{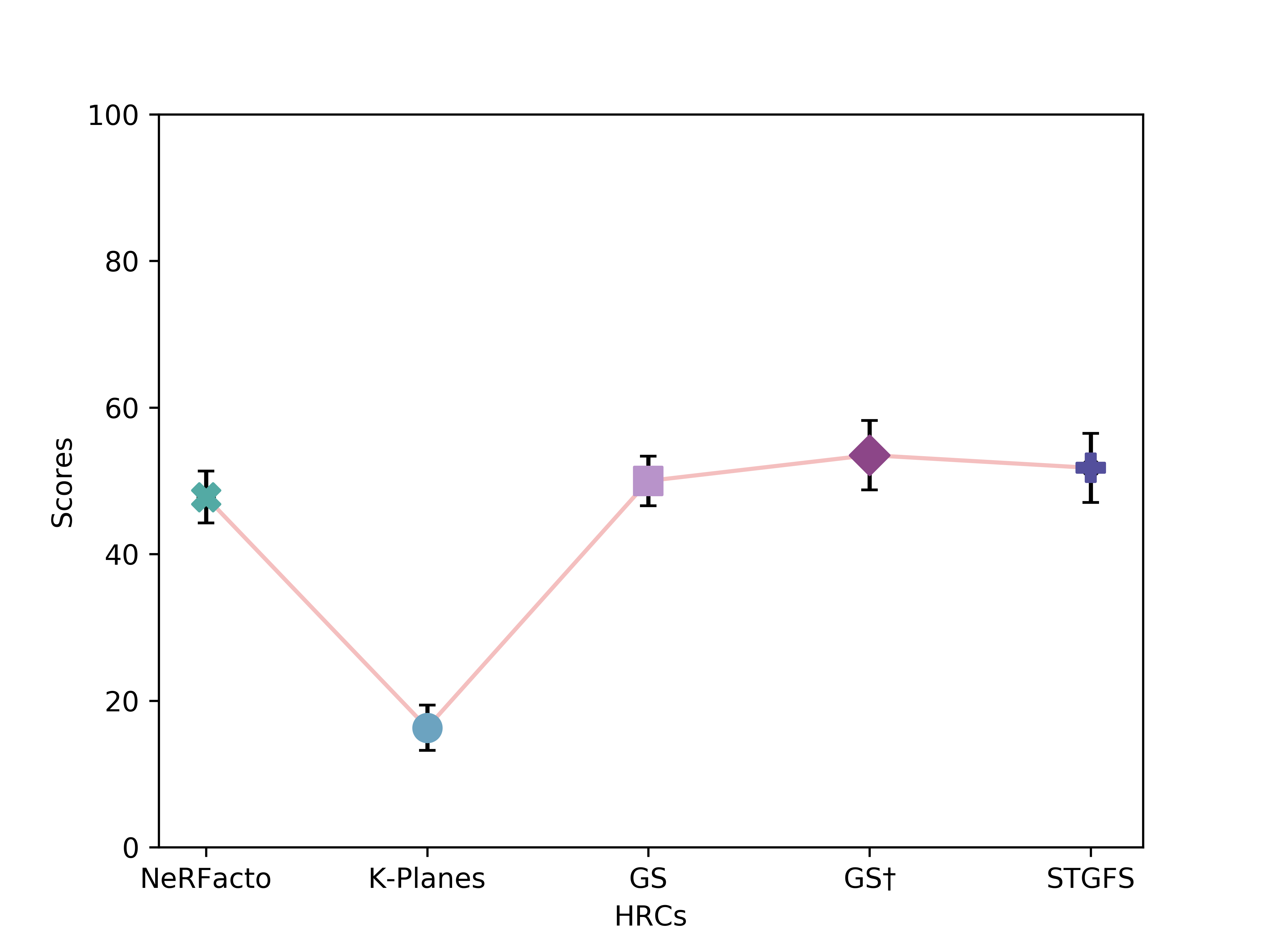}
		\centerline{(e). CBABasketball}
	\end{minipage}
	\begin{minipage}[b]{0.19\textwidth}
		\centering
		\includegraphics[width=\textwidth]{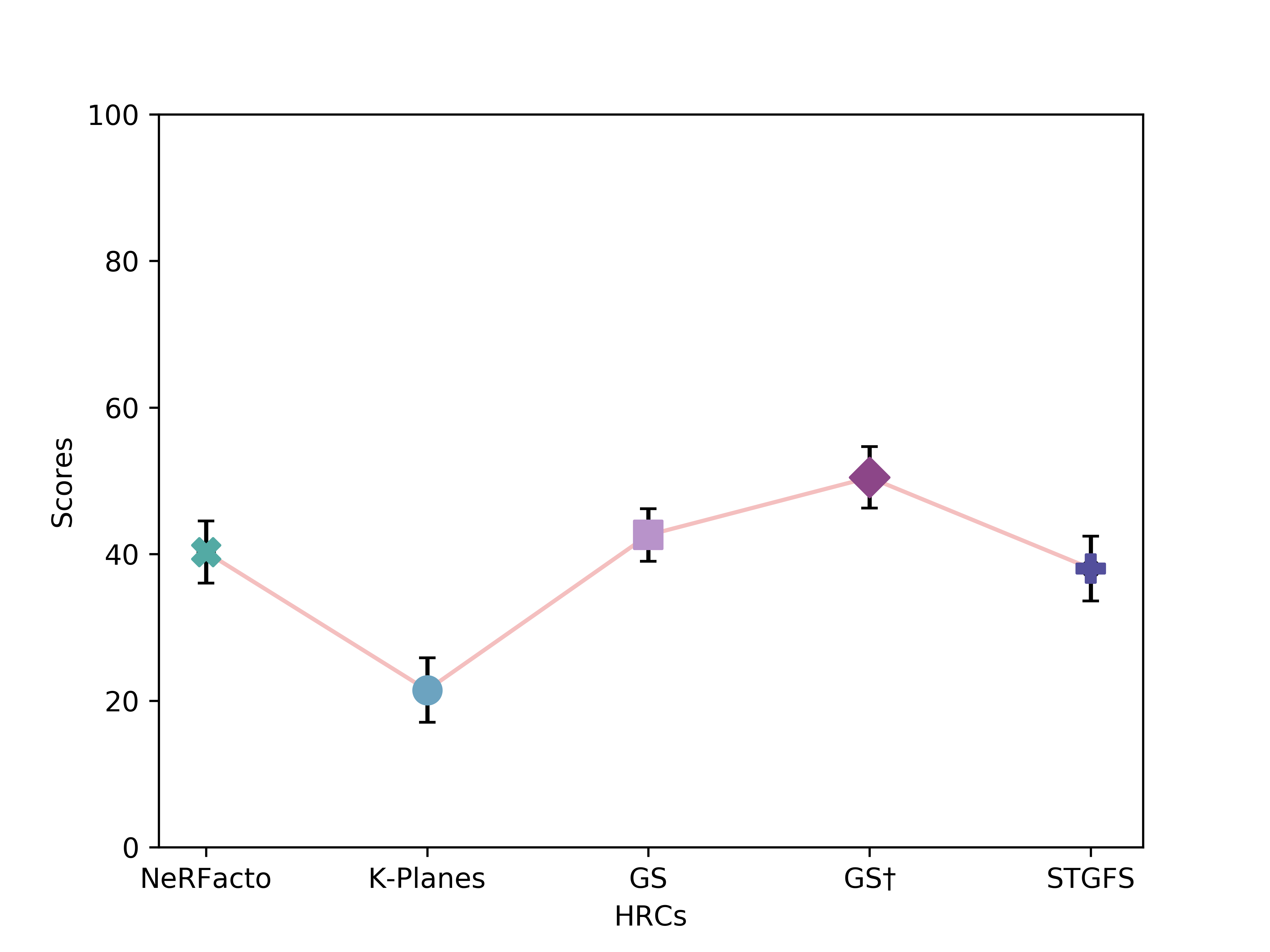}
		\centerline{(f). Fencing}
	\end{minipage}
	\begin{minipage}[b]{0.19\textwidth}
		\centering
		\includegraphics[width=\textwidth]{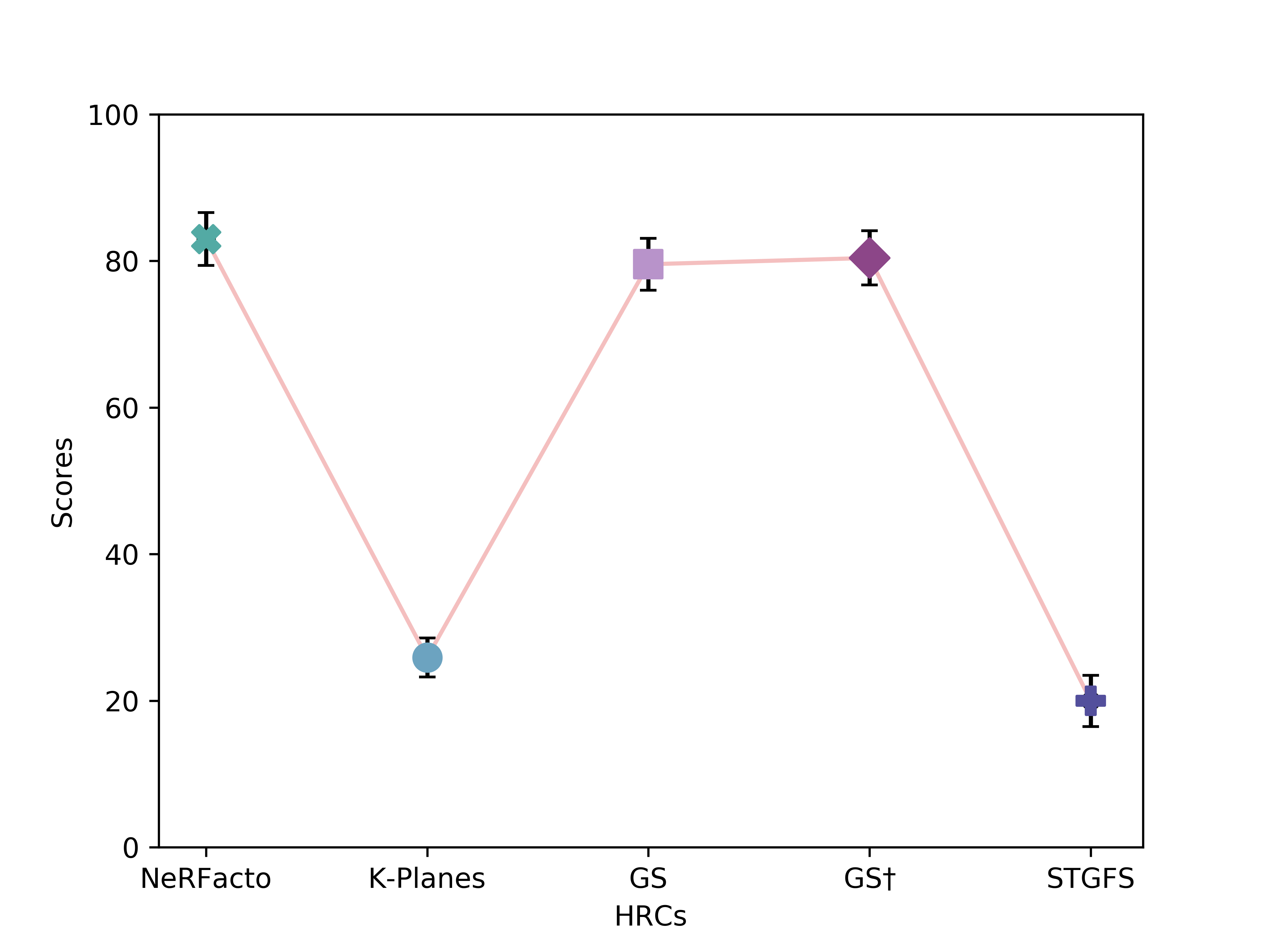}
		\centerline{(g). Frog}
	\end{minipage}
	\begin{minipage}[b]{0.19\textwidth}
		\centering
		\includegraphics[width=\textwidth]{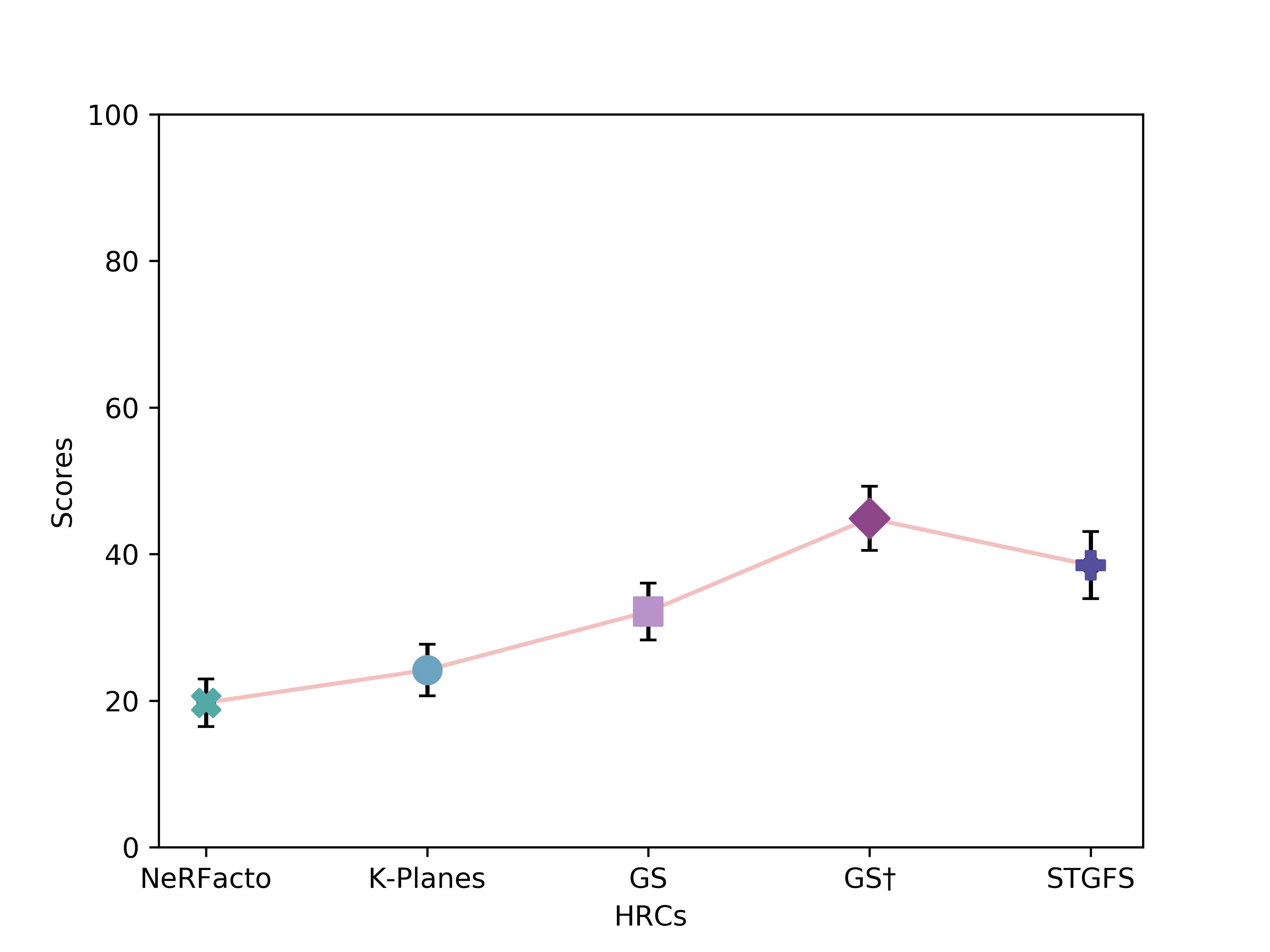}
		\centerline{(h). MartialArts}
	\end{minipage}
	\begin{minipage}[b]{0.19\textwidth}
		\centering
		\includegraphics[width=\textwidth]{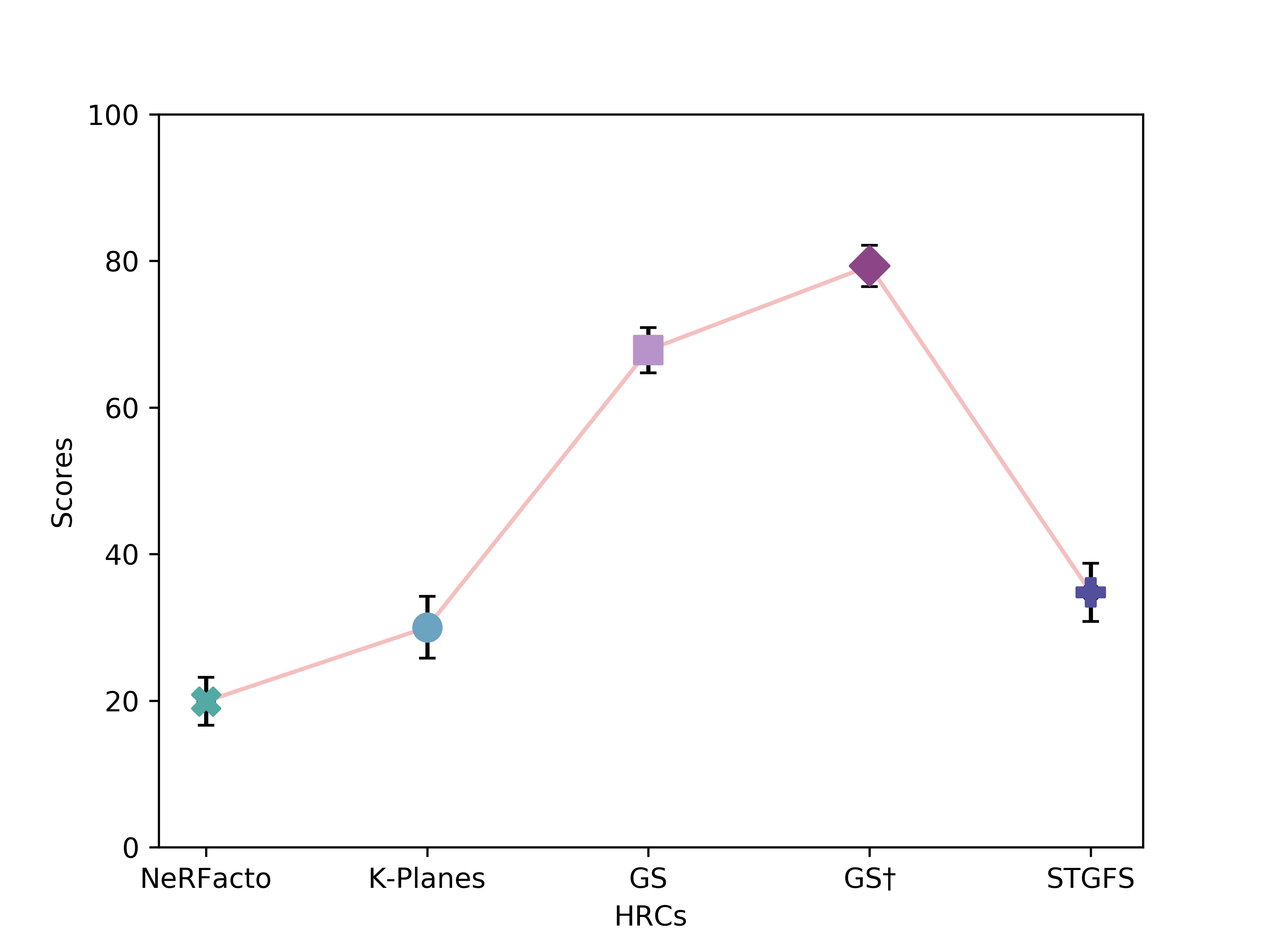}
		\centerline{(i). Painter}
	\end{minipage}
	\begin{minipage}[b]{0.19\textwidth}
		\centering
		\includegraphics[width=\textwidth]{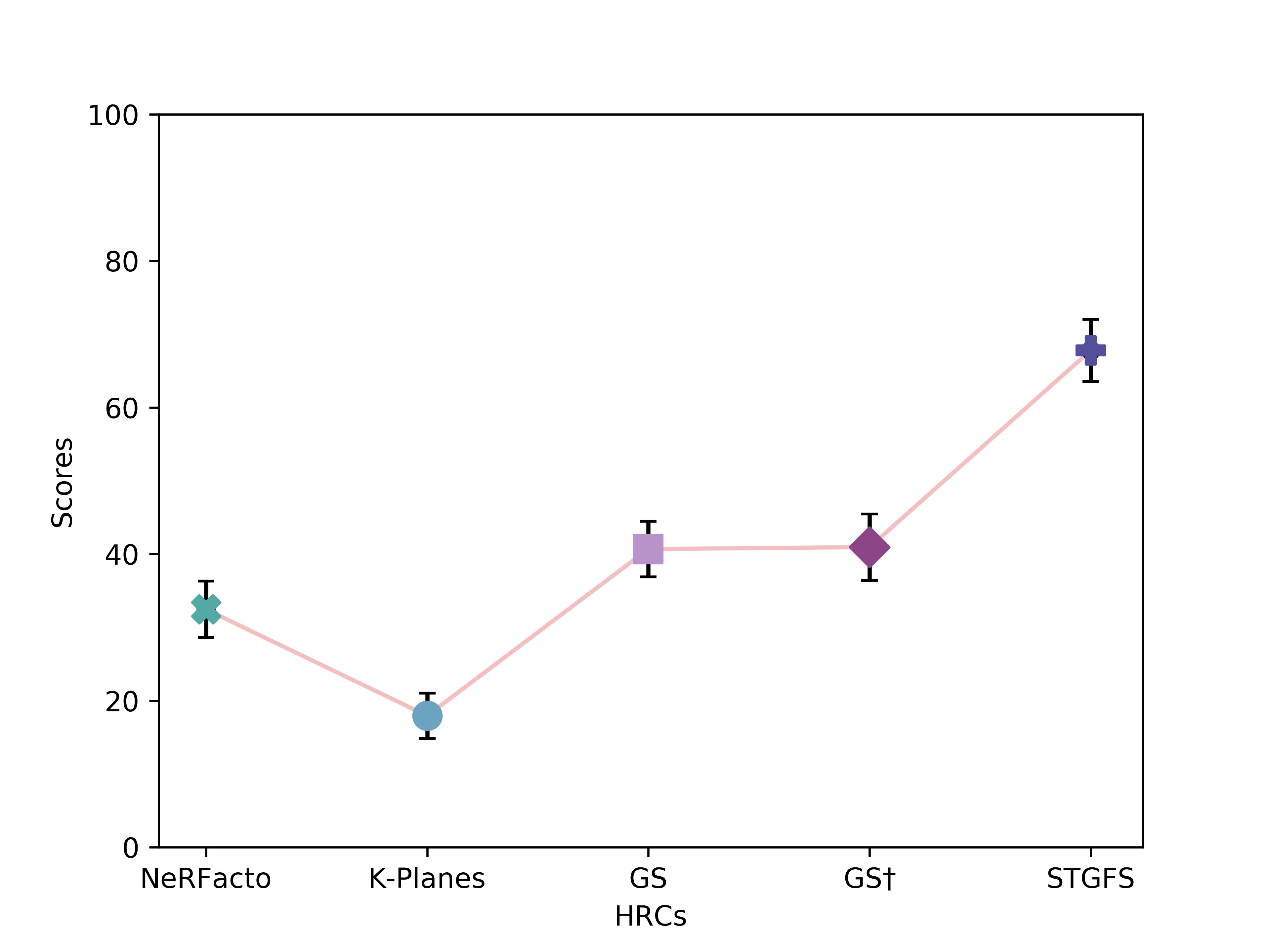}
		\centerline{(j). Pandemonium}
	\end{minipage}
	\begin{minipage}[b]{0.19\textwidth}
		\centering
		\includegraphics[width=\textwidth]{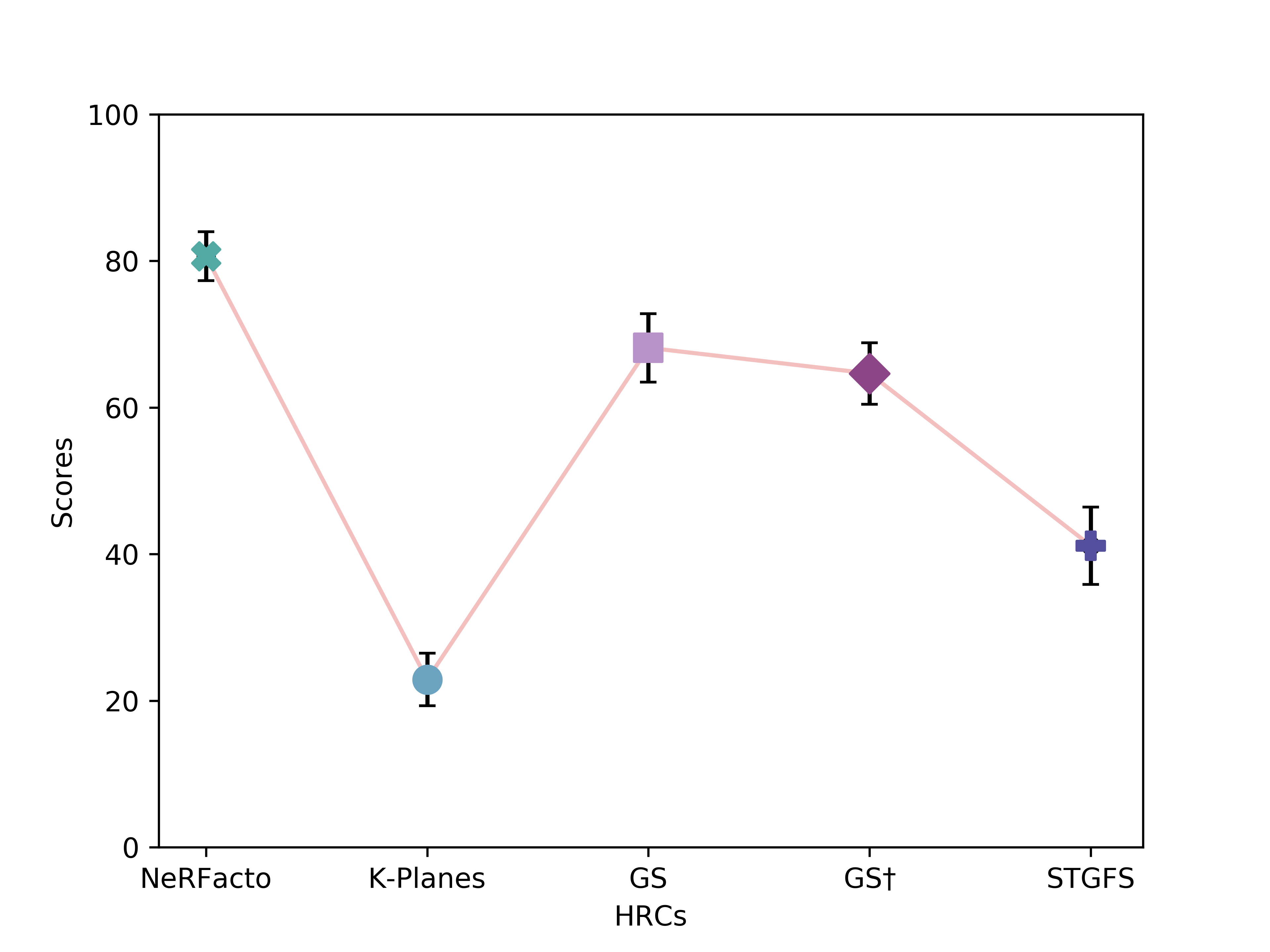}
		\centerline{(k). PoznanStreet}
	\end{minipage}
	\begin{minipage}[b]{0.19\textwidth}
		\centering
		\includegraphics[width=\textwidth]{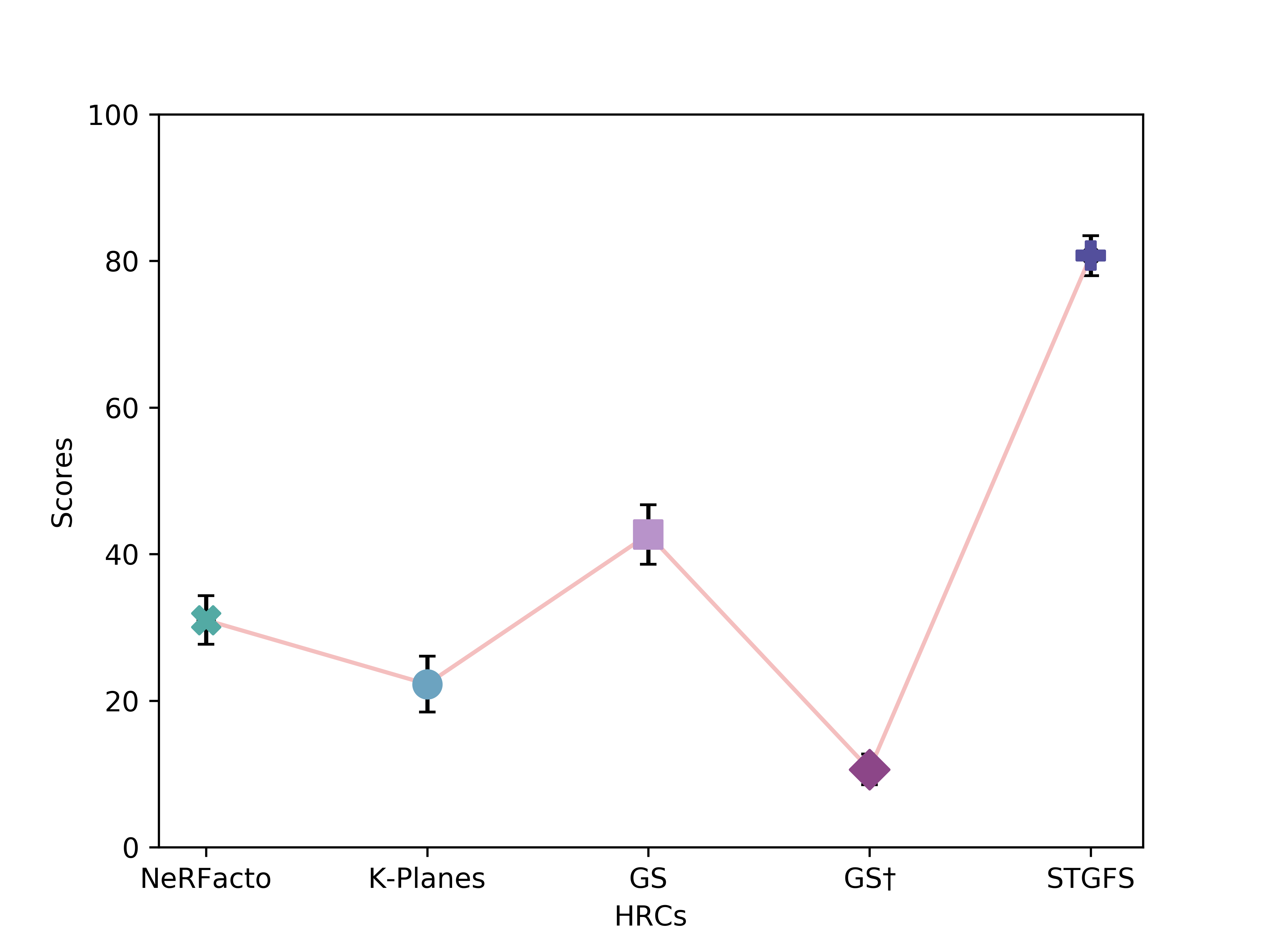}
		\centerline{(l). CoffeeMartini}
	\end{minipage}
	\begin{minipage}[b]{0.19\textwidth}
		\centering
		\includegraphics[width=\textwidth]{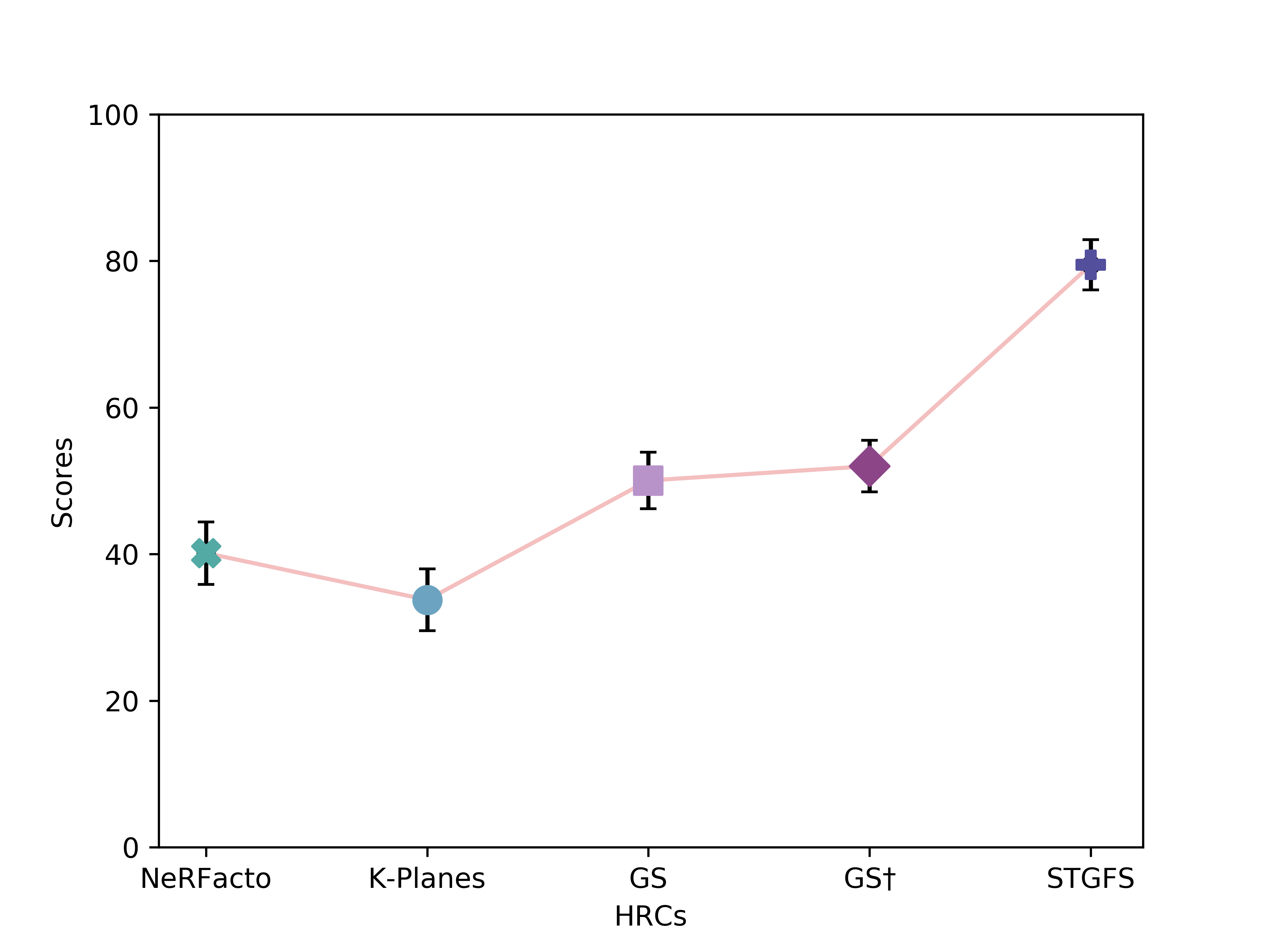}
		\centerline{(m). Flamesteak}
	\end{minipage}
	\caption{Overall MOS.}
	\label{fig:MOS-Overall}
\end{figure*}

\begin{figure*}[h!] 
	\centering
	\begin{minipage}[b]{0.19\textwidth}
		\centering
		\includegraphics[width=\textwidth]{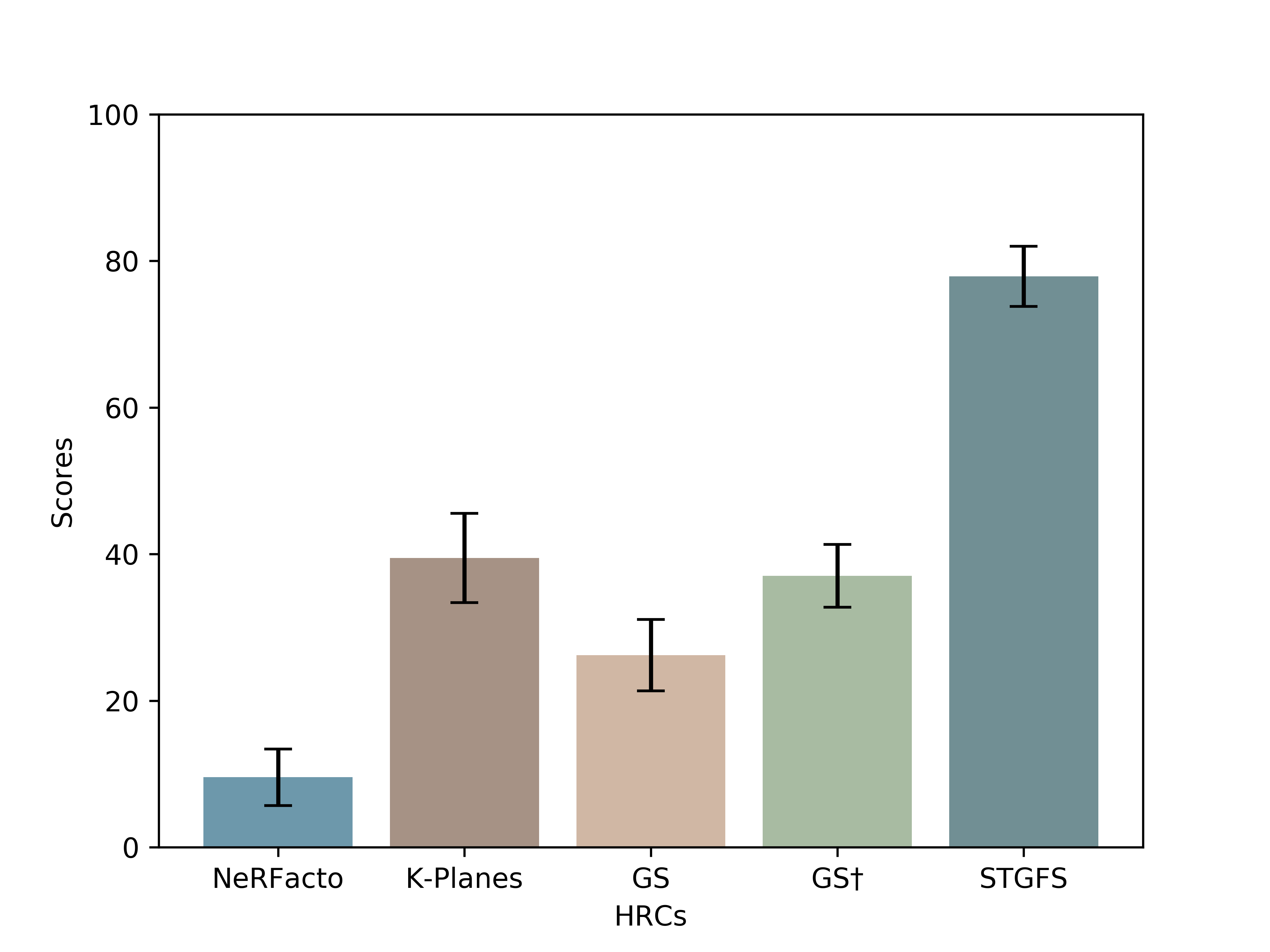}
		\centerline{(a). Barn}
	\end{minipage}
	\begin{minipage}[b]{0.19\textwidth}
		\centering
		\includegraphics[width=\textwidth]{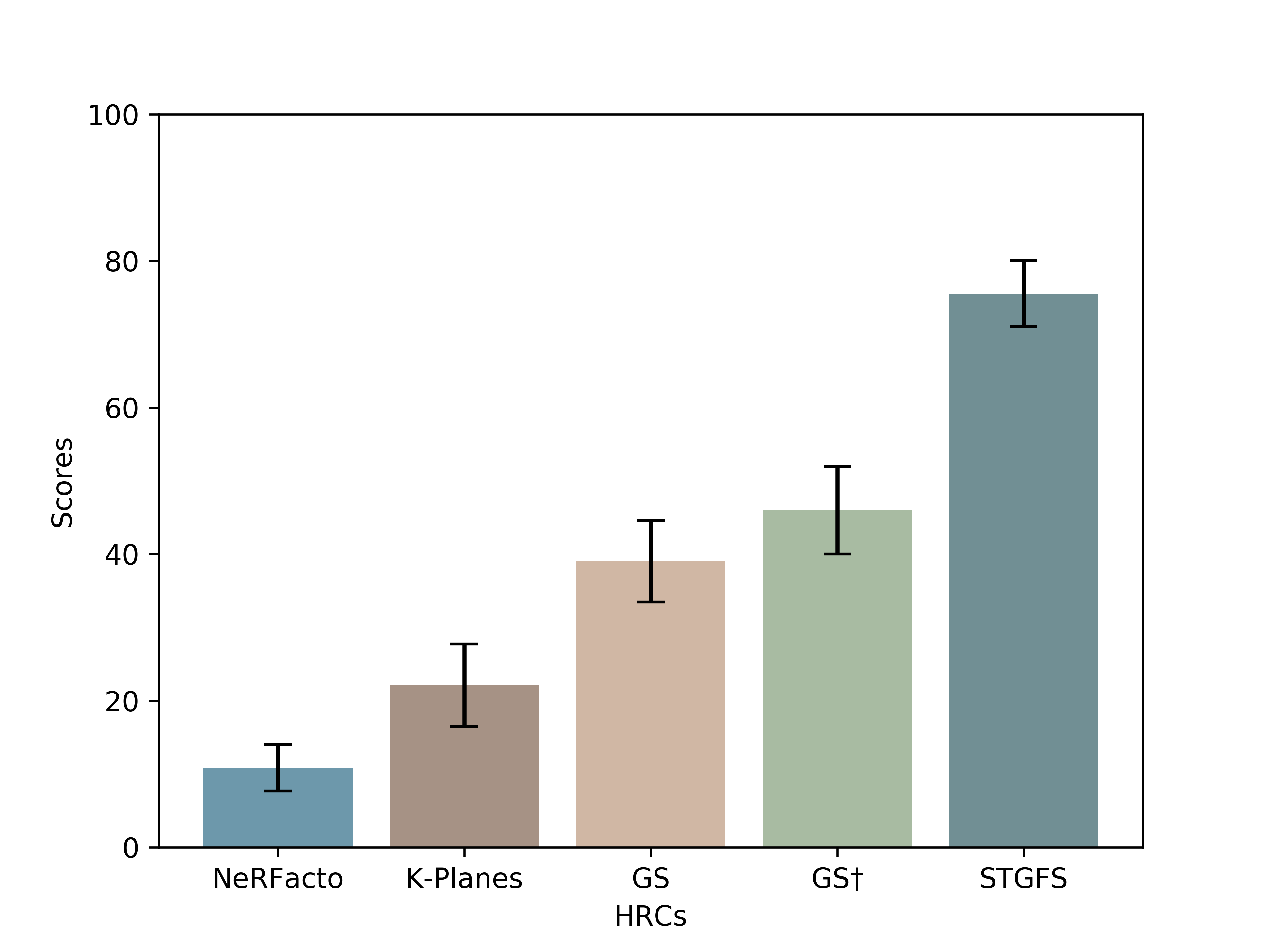}
		\centerline{(b). Blocks}
	\end{minipage}
	\begin{minipage}[b]{0.19\textwidth}
		\centering
		\includegraphics[width=\textwidth]{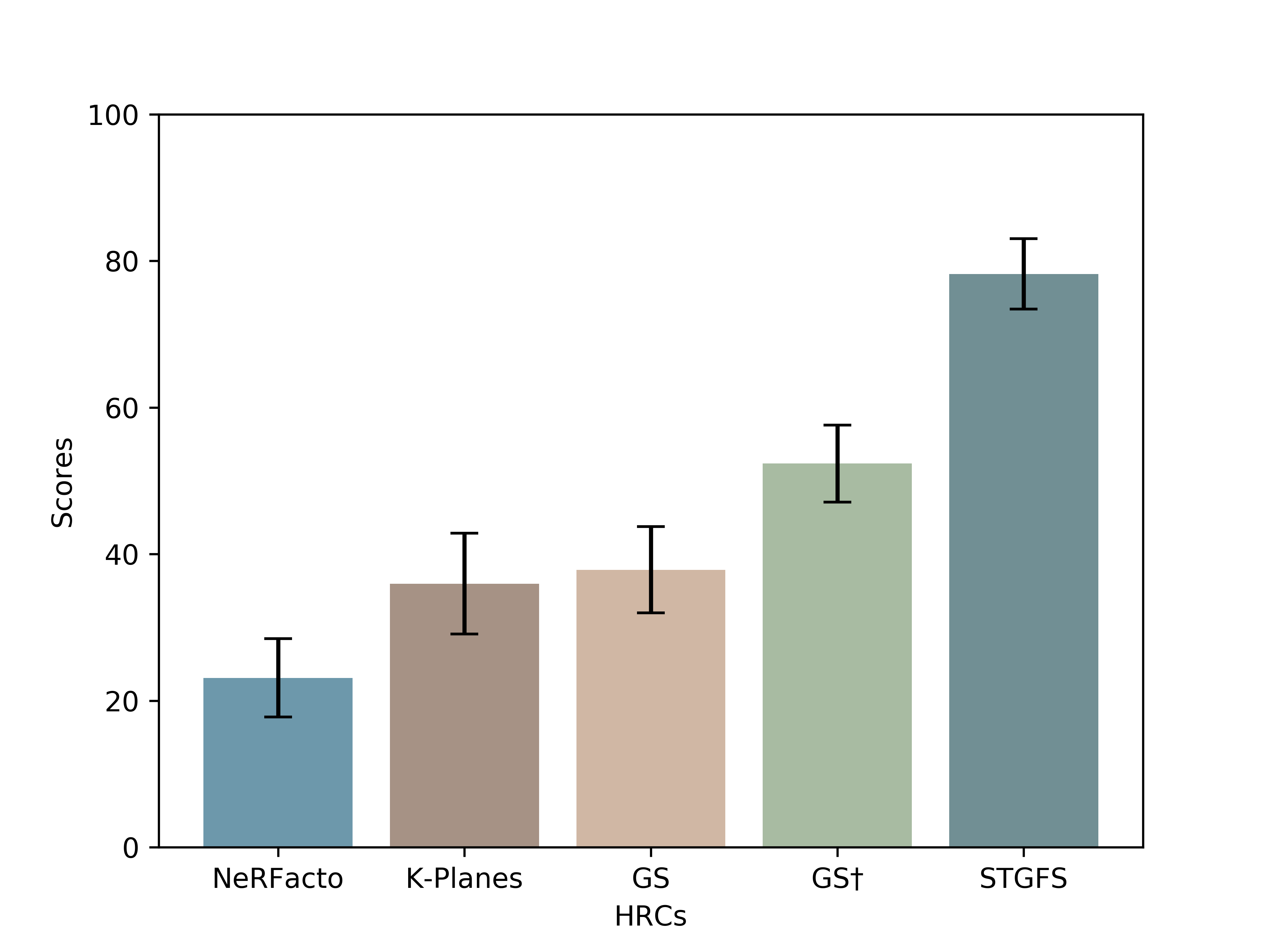}
		\centerline{(c). Breakfast}
	\end{minipage}
	\begin{minipage}[b]{0.19\textwidth}
		\centering
		\includegraphics[width=\textwidth]{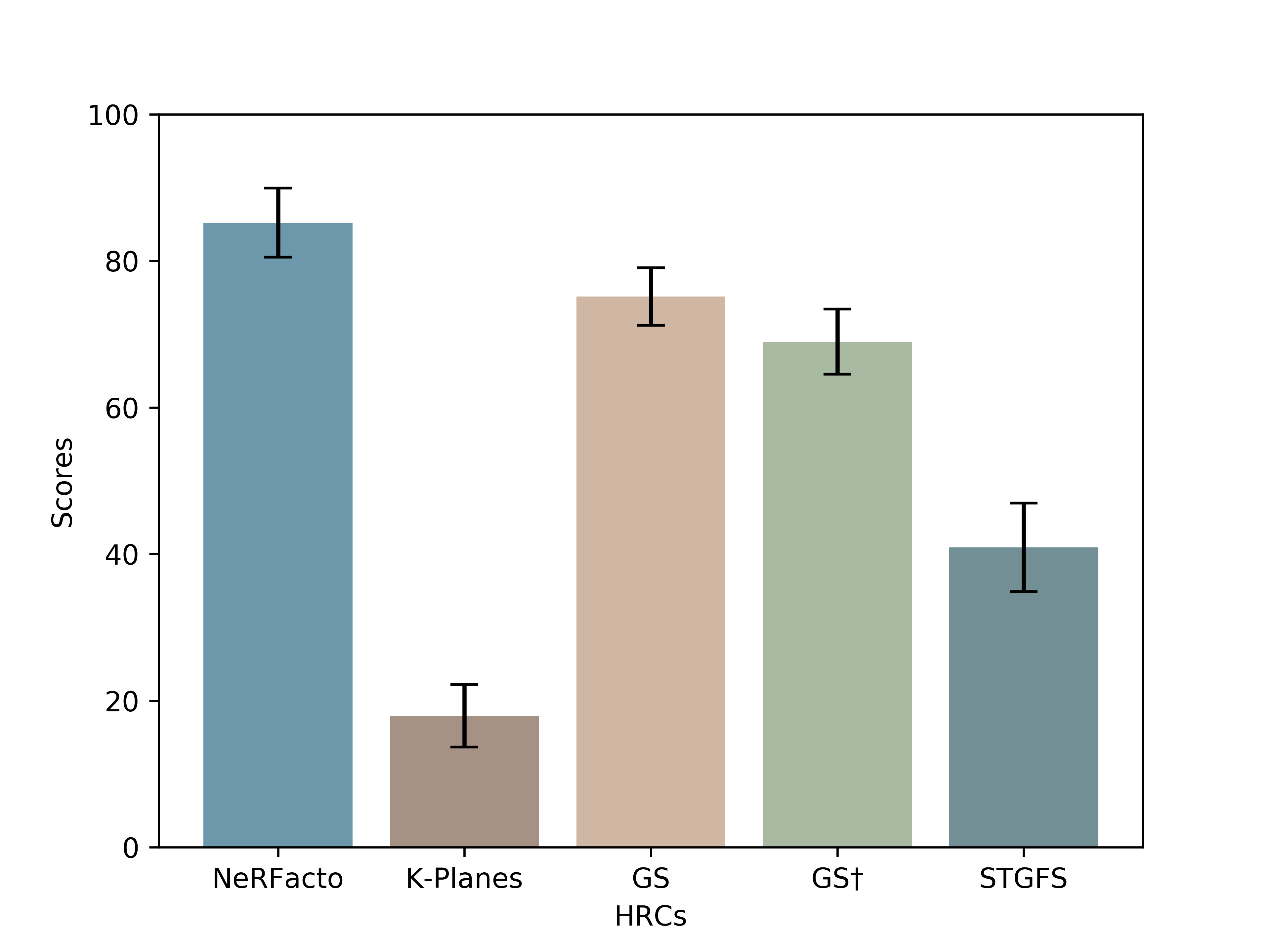}
		\centerline{(d). Carpark}
	\end{minipage}
	\begin{minipage}[b]{0.19\textwidth}
		\centering
		\includegraphics[width=\textwidth]{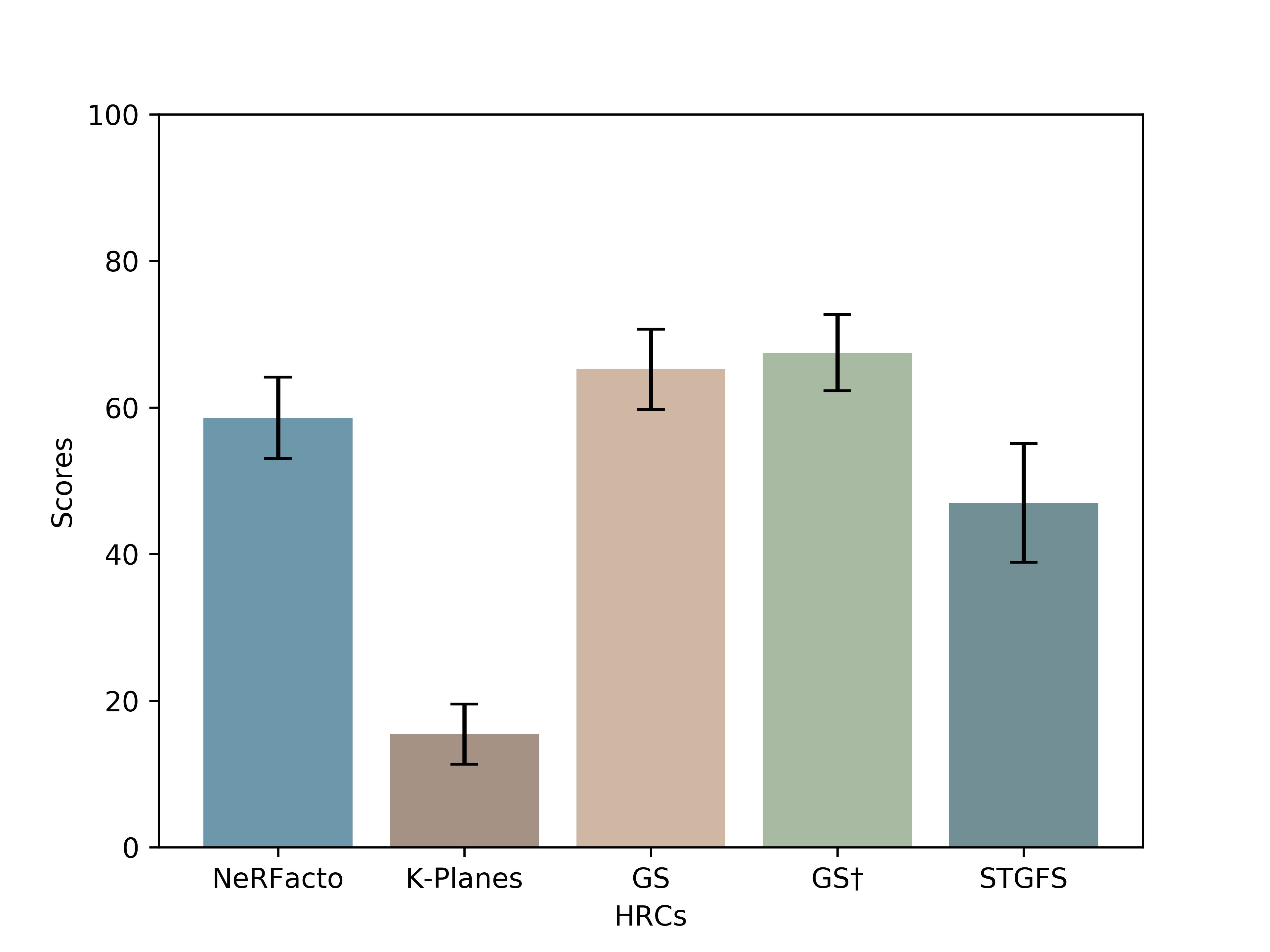}
		\centerline{(e). CBABasketball}
	\end{minipage}
	\begin{minipage}[b]{0.19\textwidth}
		\centering
		\includegraphics[width=\textwidth]{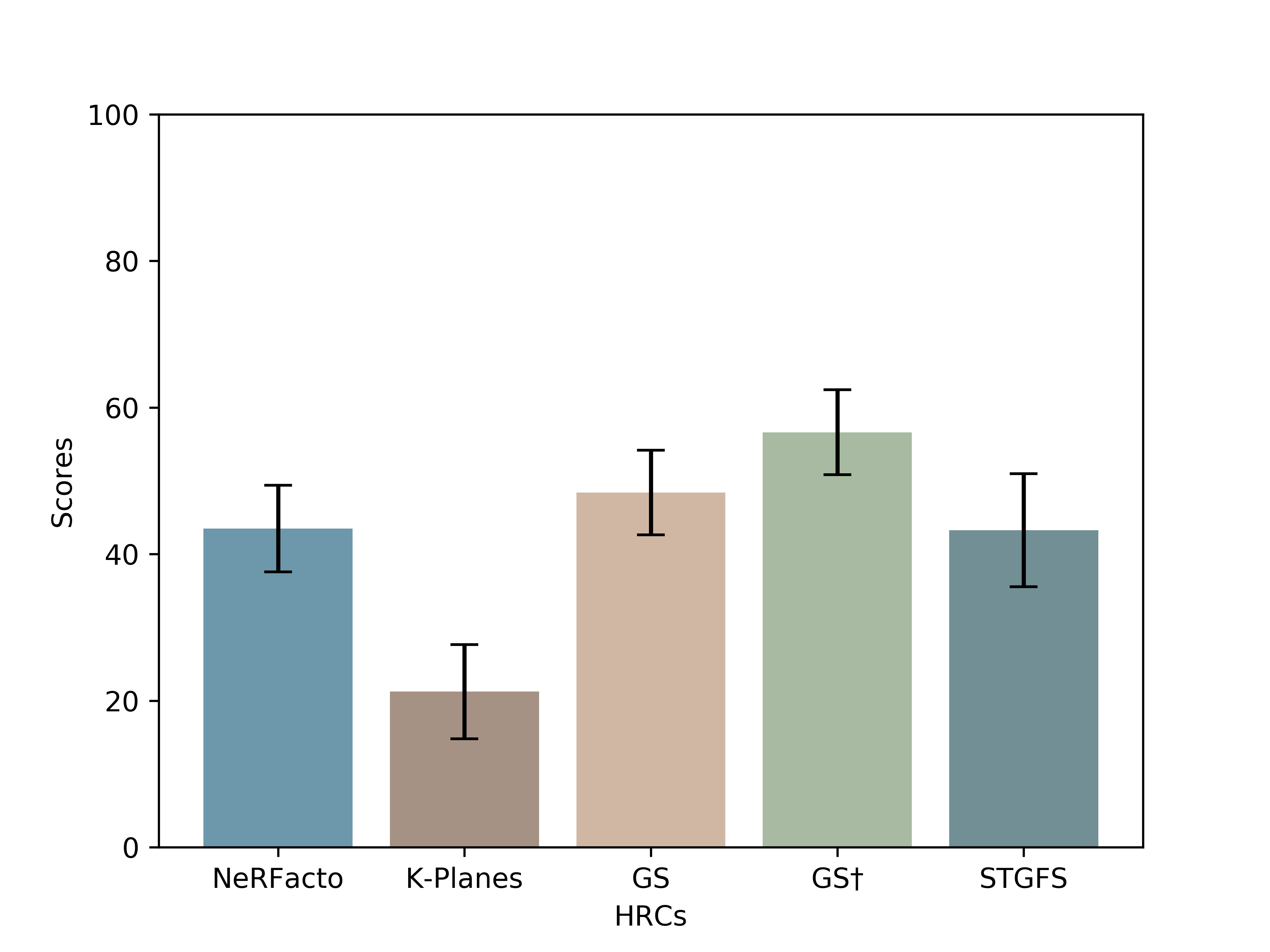}
		\centerline{(f). Fencing}
	\end{minipage}
	\begin{minipage}[b]{0.19\textwidth}
		\centering
		\includegraphics[width=\textwidth]{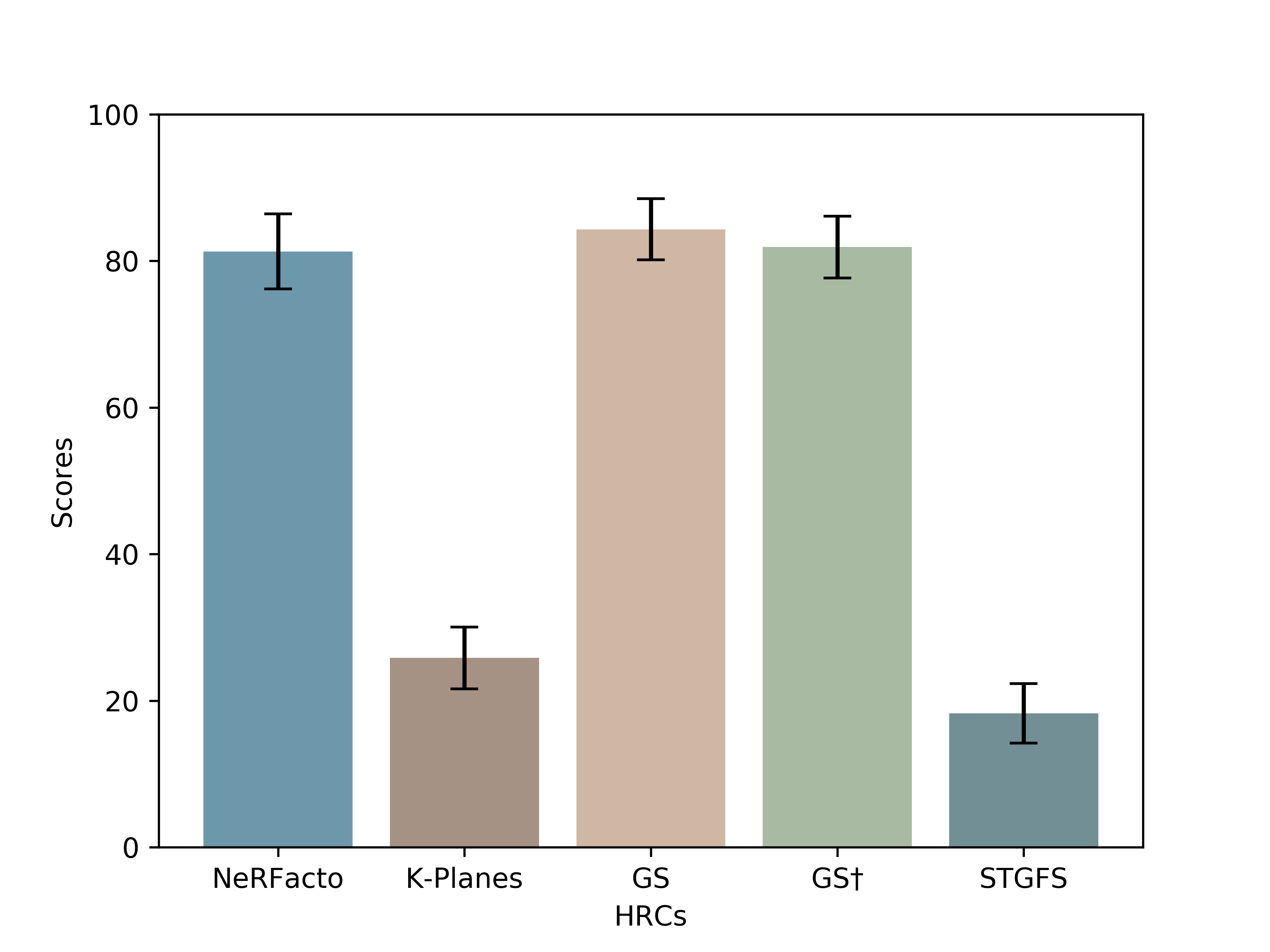}
		\centerline{(g). Frog}
	\end{minipage}
	\begin{minipage}[b]{0.19\textwidth}
		\centering
		\includegraphics[width=\textwidth]{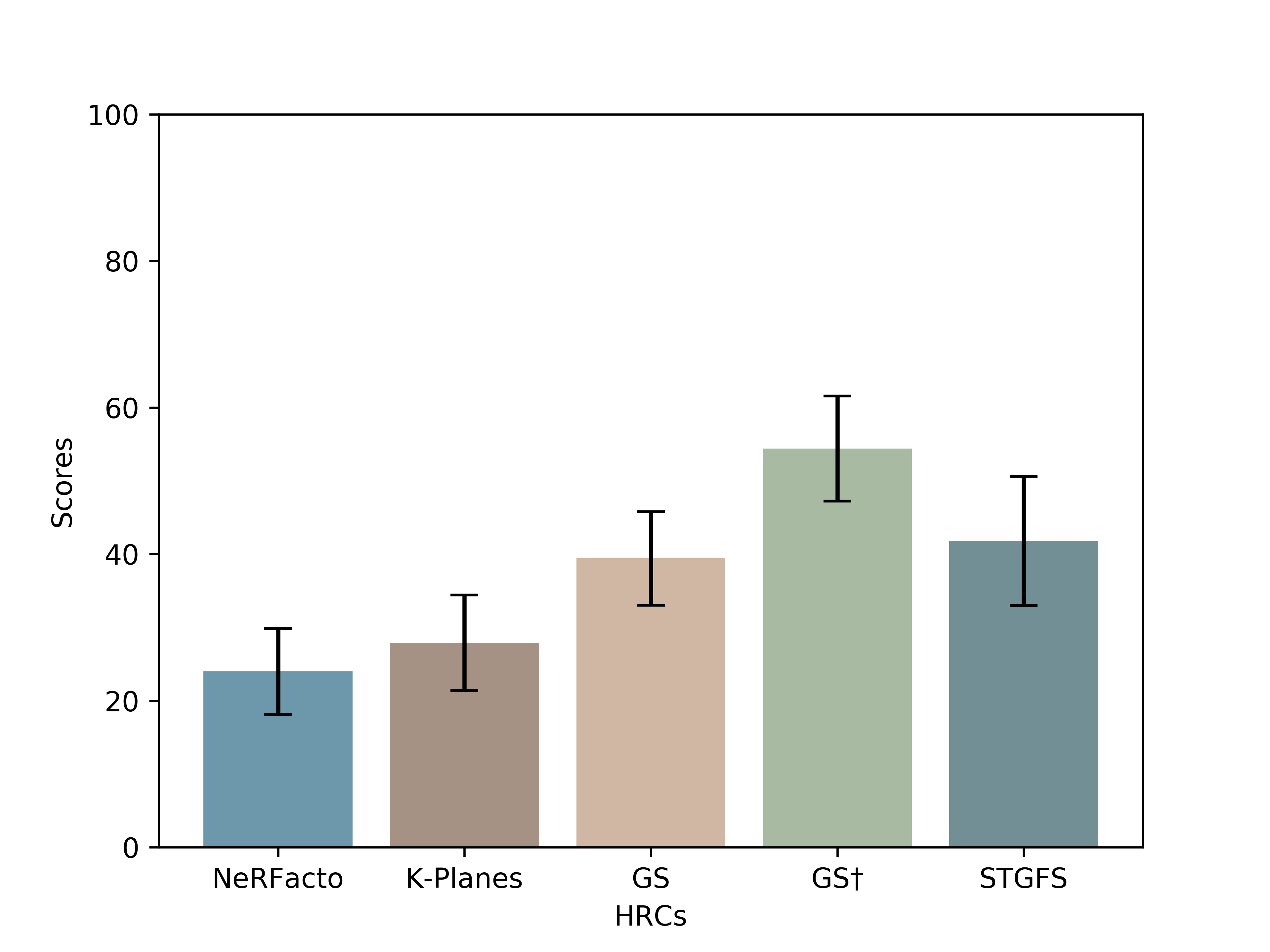}
		\centerline{(h). MartialArts}
	\end{minipage}
	\begin{minipage}[b]{0.19\textwidth}
		\centering
		\includegraphics[width=\textwidth]{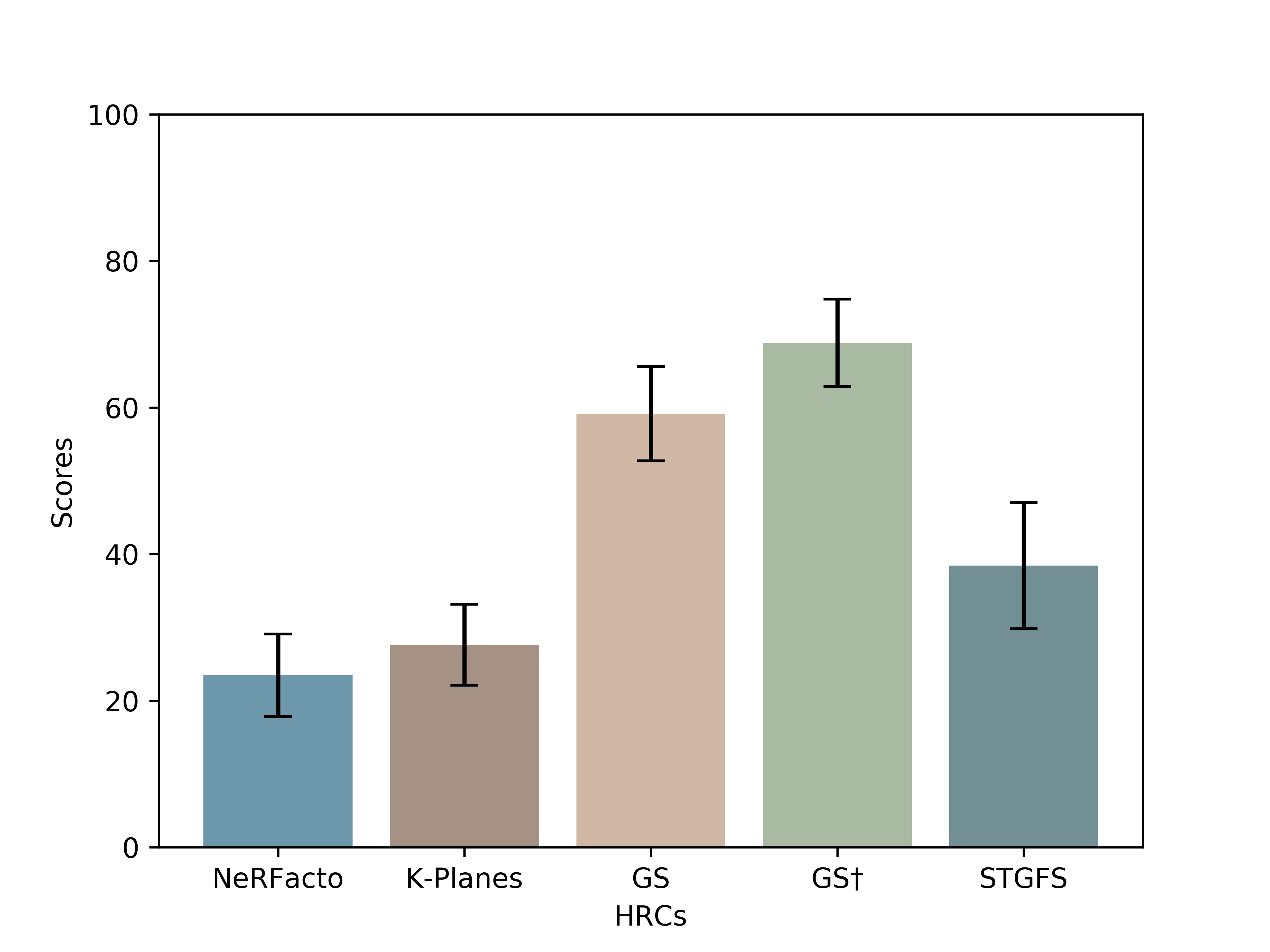}
		\centerline{(i). Painter}
	\end{minipage}
	\begin{minipage}[b]{0.19\textwidth}
		\centering
		\includegraphics[width=\textwidth]{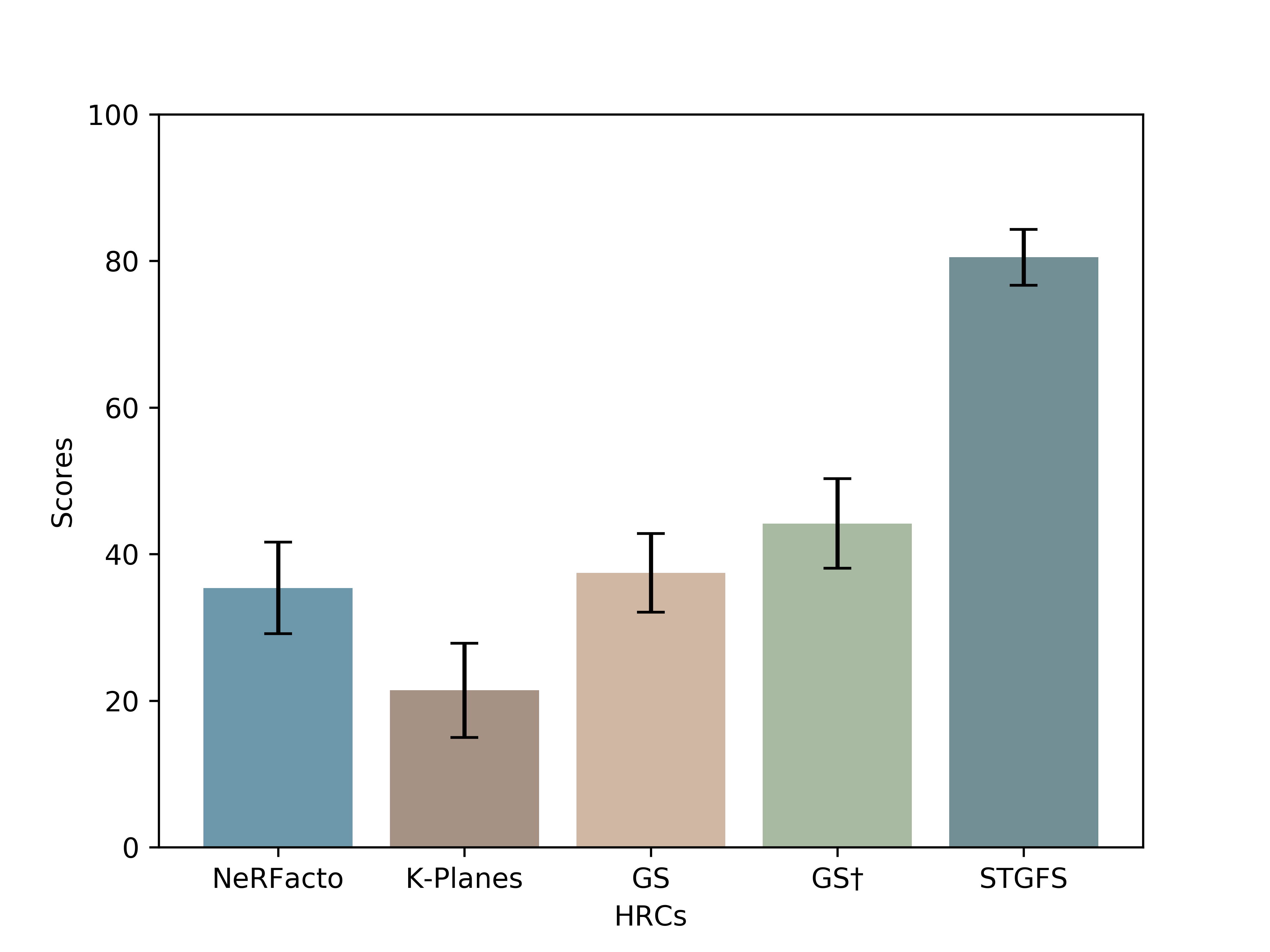}
		\centerline{(j). Pandemonium}
	\end{minipage}
	\begin{minipage}[b]{0.19\textwidth}
		\centering
		\includegraphics[width=\textwidth]{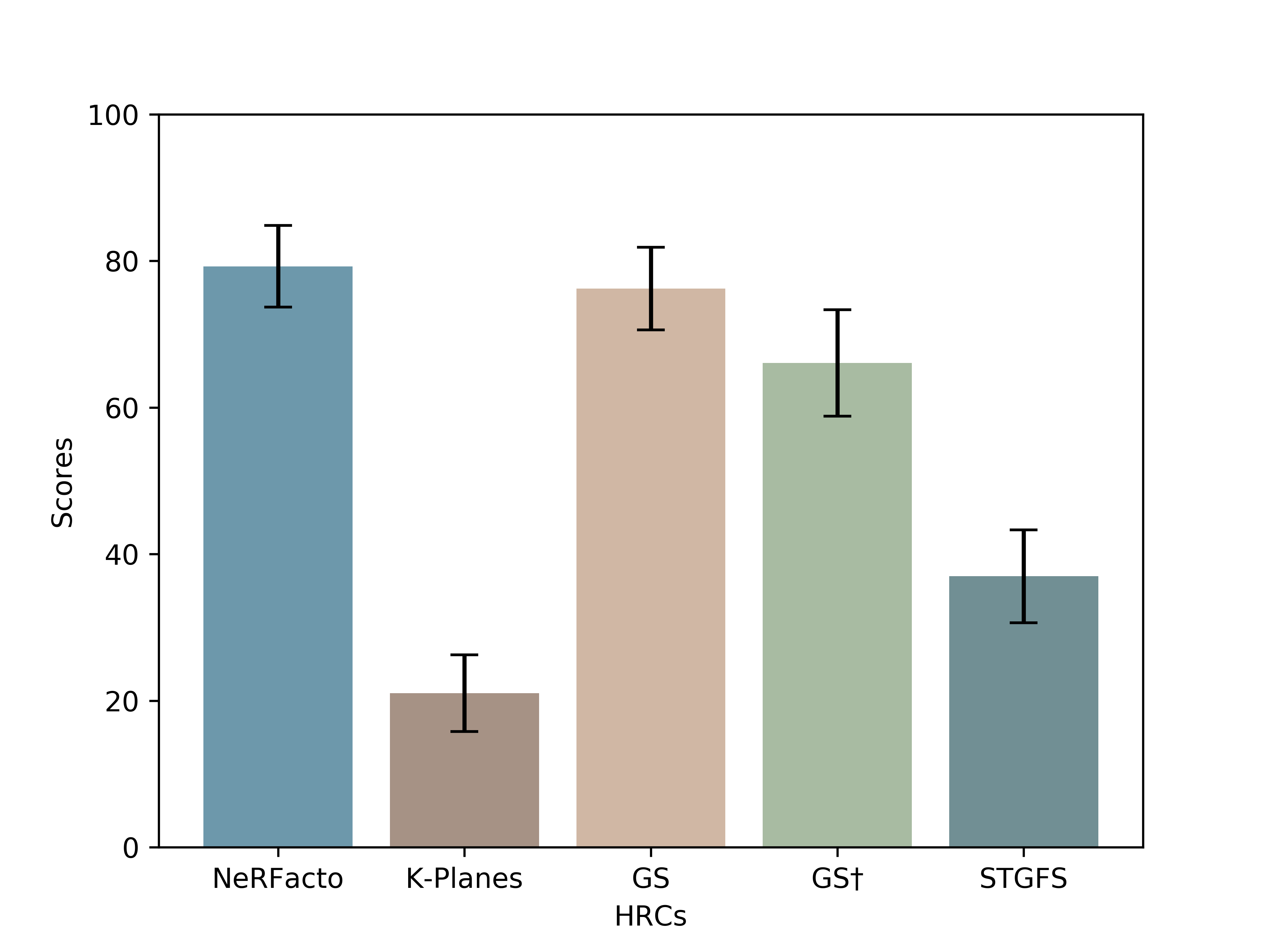}
		\centerline{(k). PoznanStreet}
	\end{minipage}
	\begin{minipage}[b]{0.19\textwidth}
		\centering
		\includegraphics[width=\textwidth]{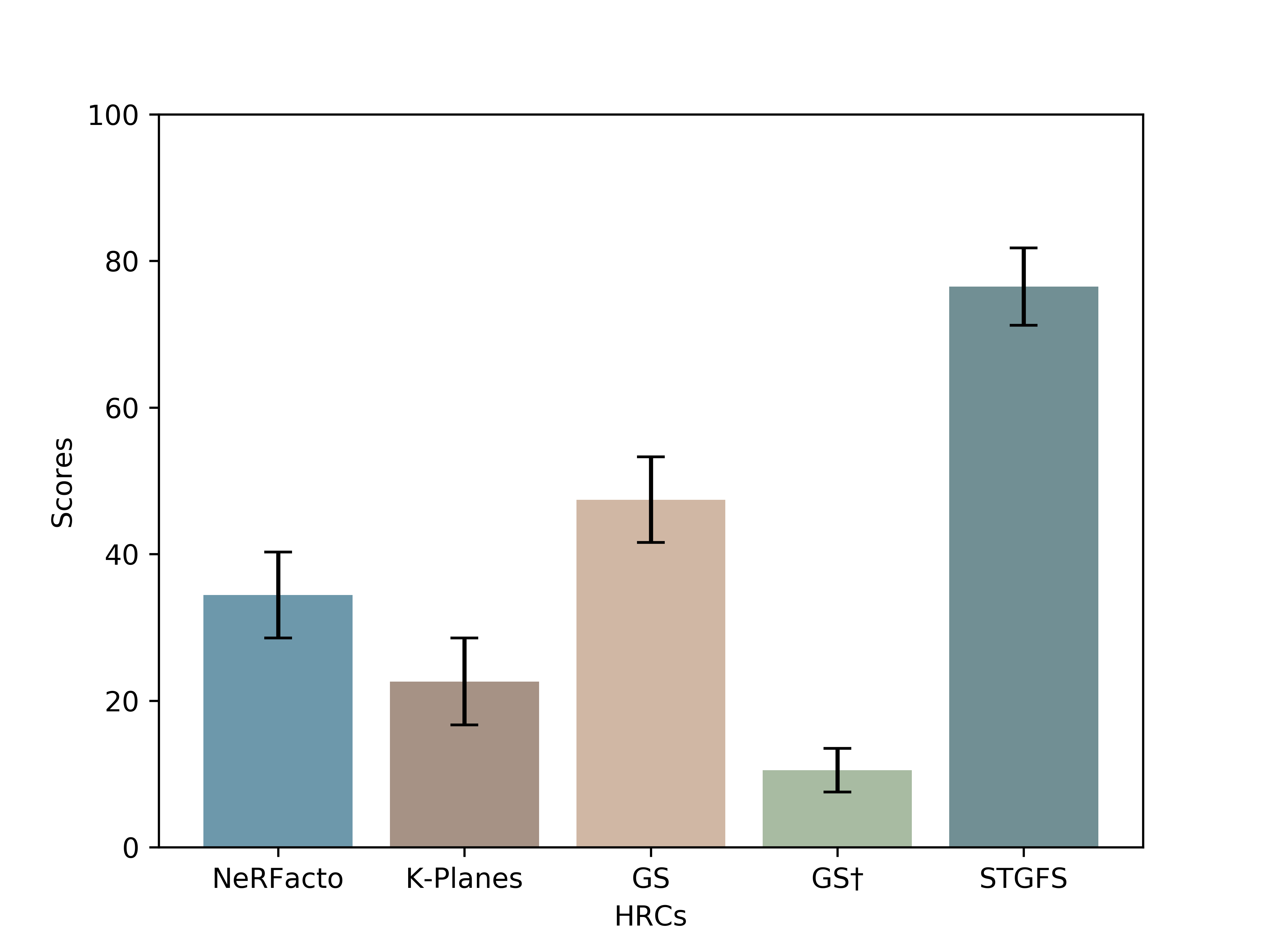}
		\centerline{(l). CoffeeMartini}
	\end{minipage}
	\begin{minipage}[b]{0.19\textwidth}
		\centering
		\includegraphics[width=\textwidth]{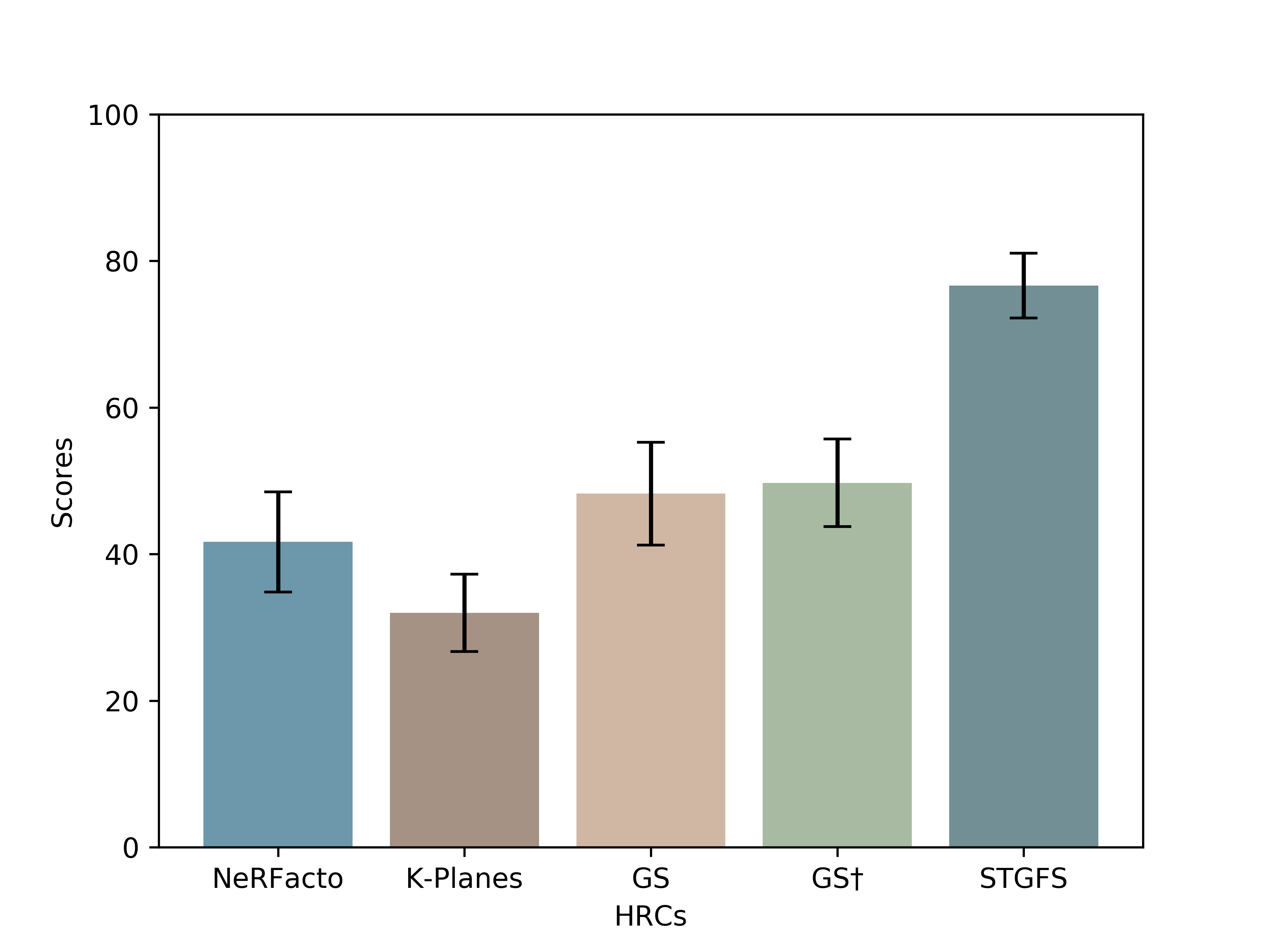}
		\centerline{(m). Flamesteak}
	\end{minipage}
	\caption{MOS of multi-view subjective test.}
	\label{fig:MOS-D}
\end{figure*}

\begin{figure*}[h!] 
	\centering
	\begin{minipage}[b]{0.19\textwidth}
		\centering
		\includegraphics[width=\textwidth]{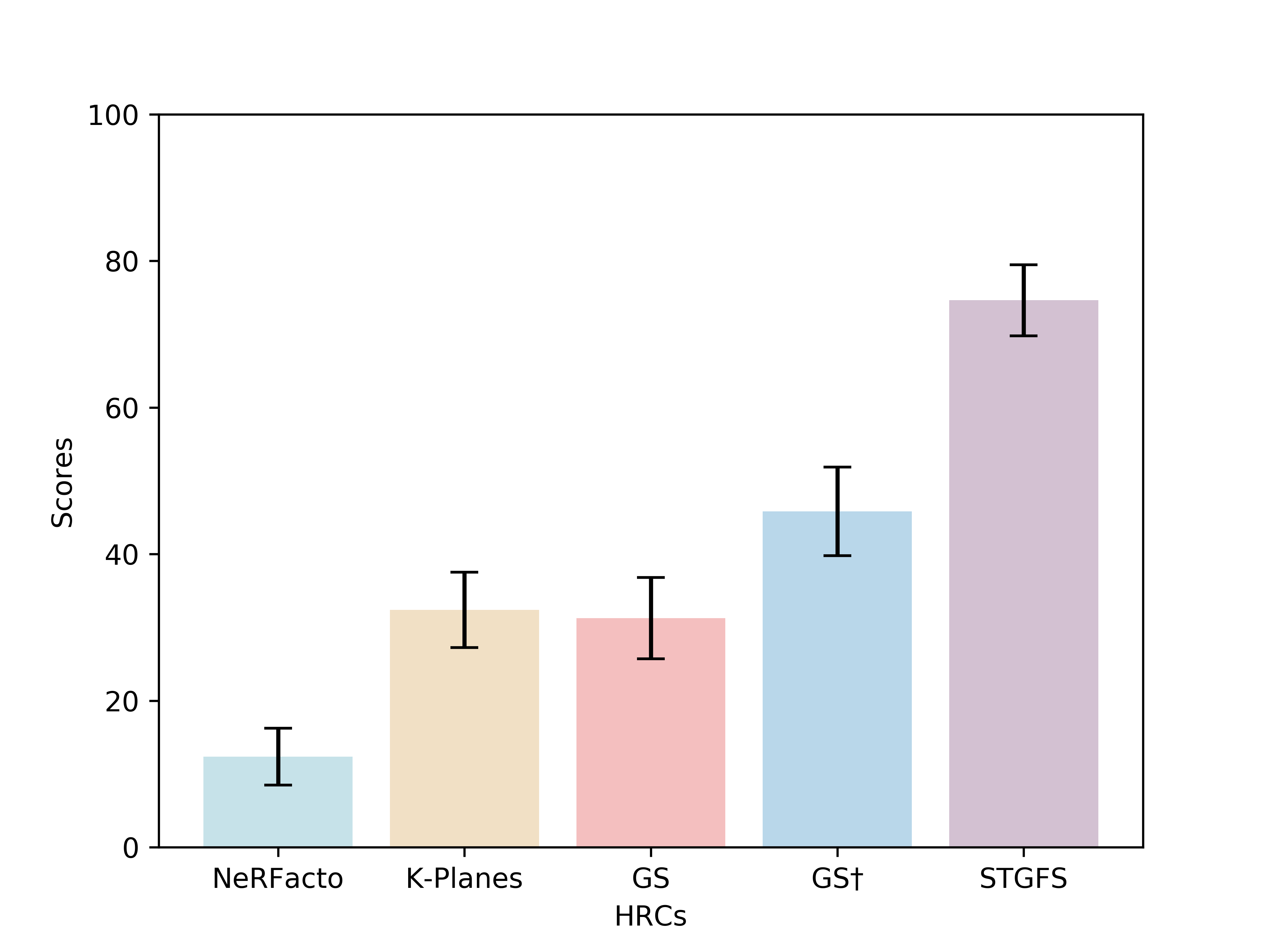}
		\centerline{(a). Barn}
	\end{minipage}
	\begin{minipage}[b]{0.19\textwidth}
		\centering
		\includegraphics[width=\textwidth]{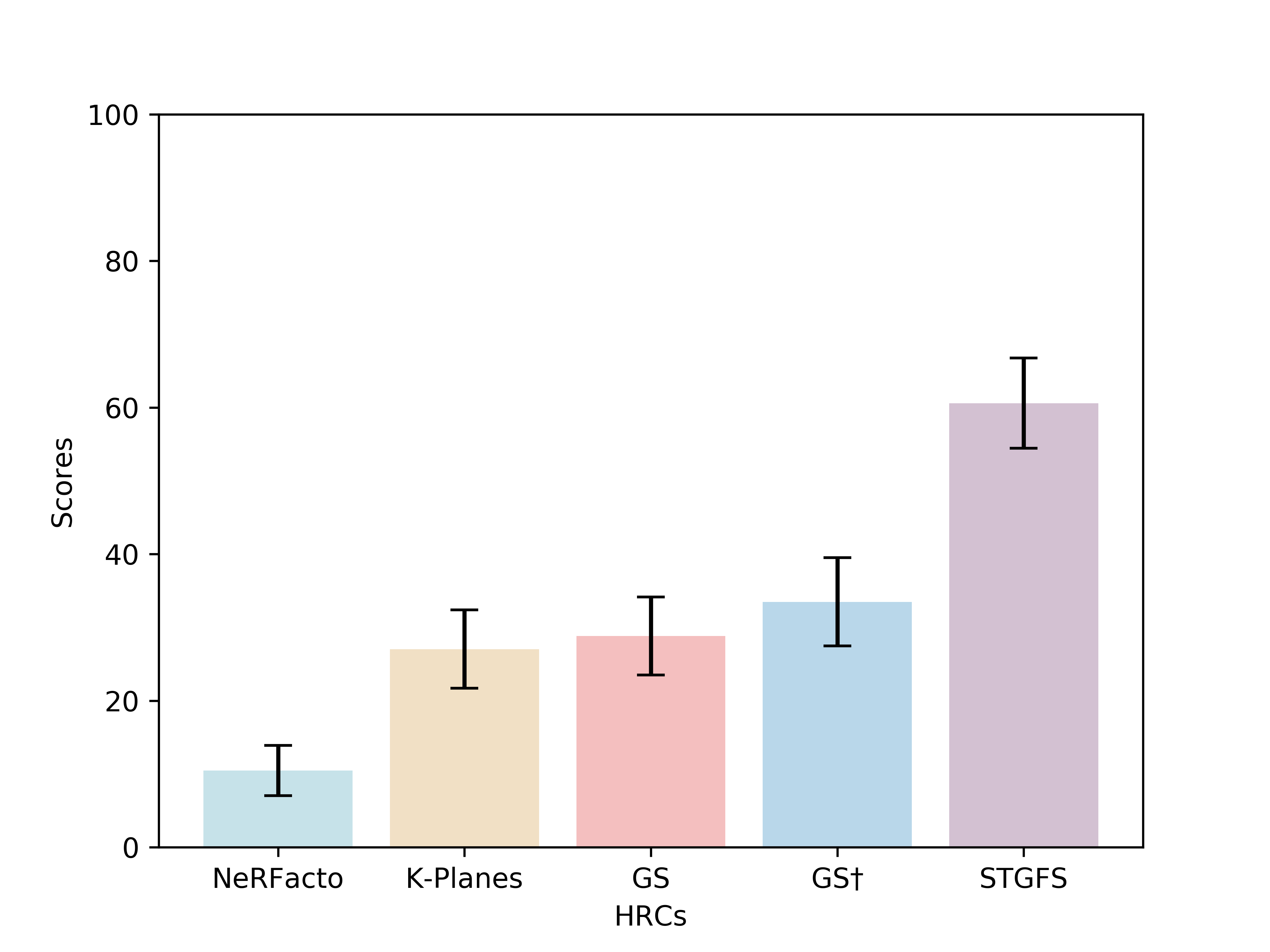}
		\centerline{(b). Blocks}
	\end{minipage}
	\begin{minipage}[b]{0.19\textwidth}
		\centering
		\includegraphics[width=\textwidth]{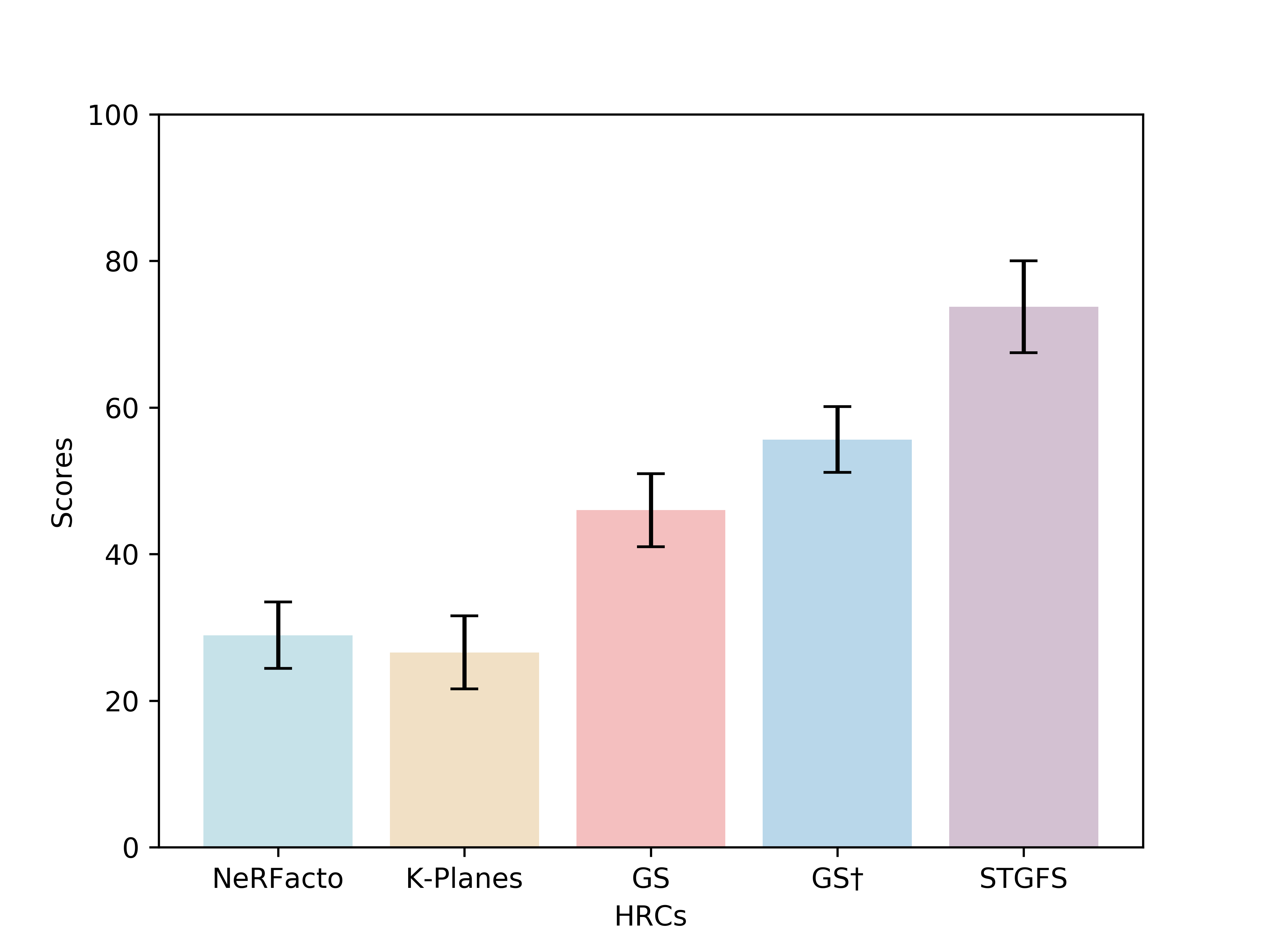}
		\centerline{(c). Breakfast}
	\end{minipage}
	\begin{minipage}[b]{0.19\textwidth}
		\centering
		\includegraphics[width=\textwidth]{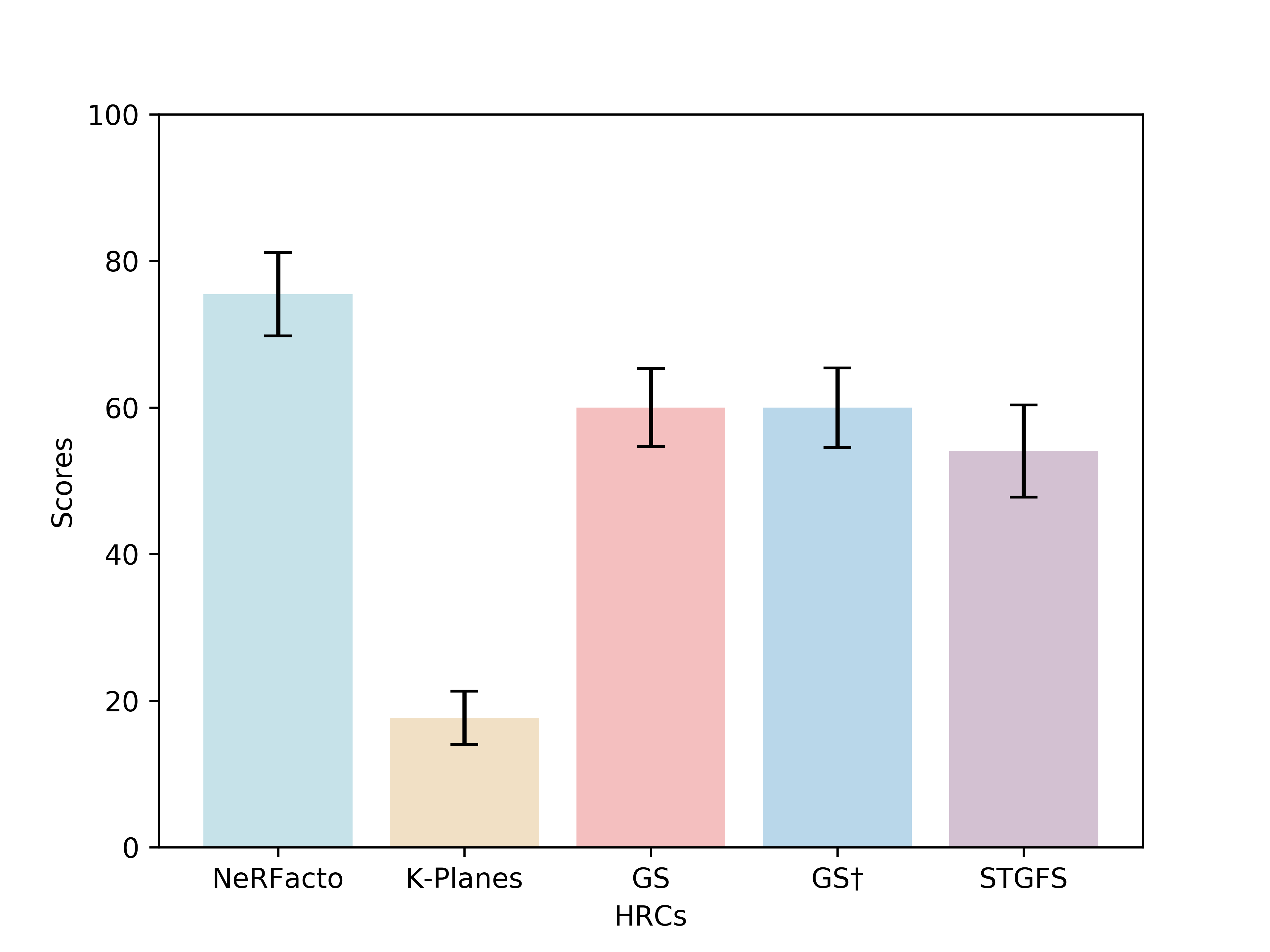}
		\centerline{(d). Carpark}
	\end{minipage}
	\begin{minipage}[b]{0.19\textwidth}
		\centering
		\includegraphics[width=\textwidth]{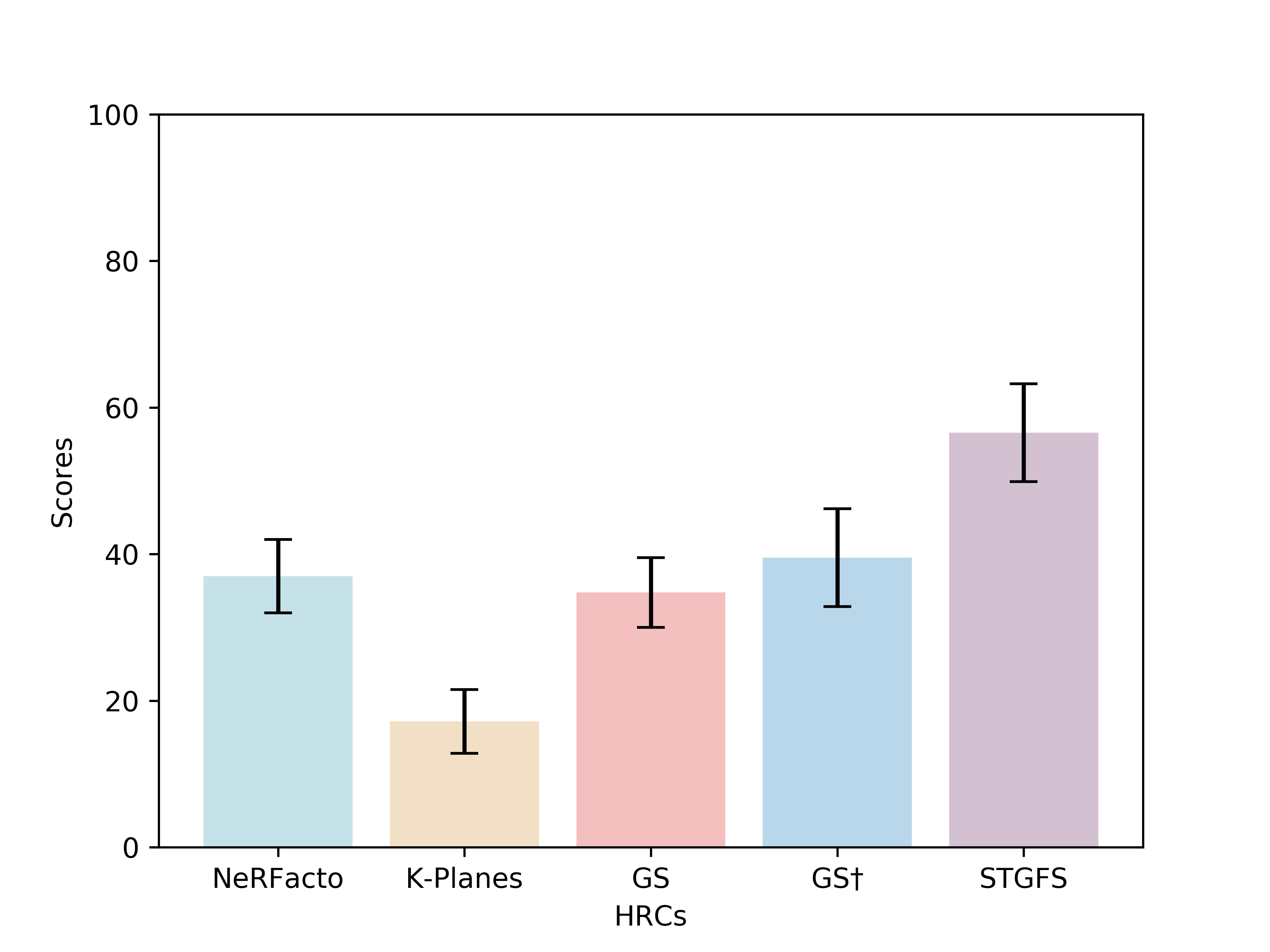}
		\centerline{(e). CBABasketball}
	\end{minipage}
	\begin{minipage}[b]{0.19\textwidth}
		\centering
		\includegraphics[width=\textwidth]{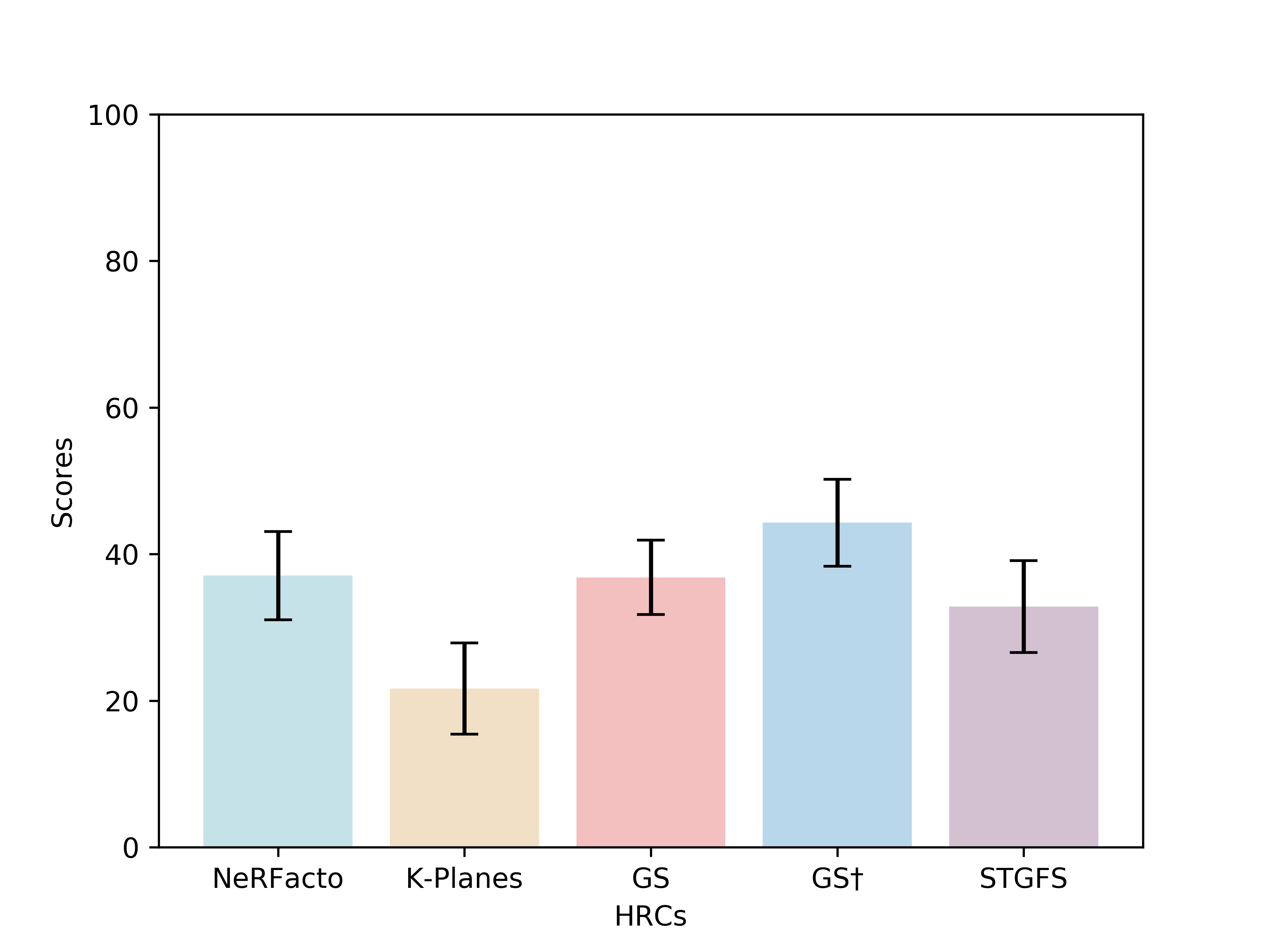}
		\centerline{(f). Fencing}
	\end{minipage}
	\begin{minipage}[b]{0.19\textwidth}
		\centering
		\includegraphics[width=\textwidth]{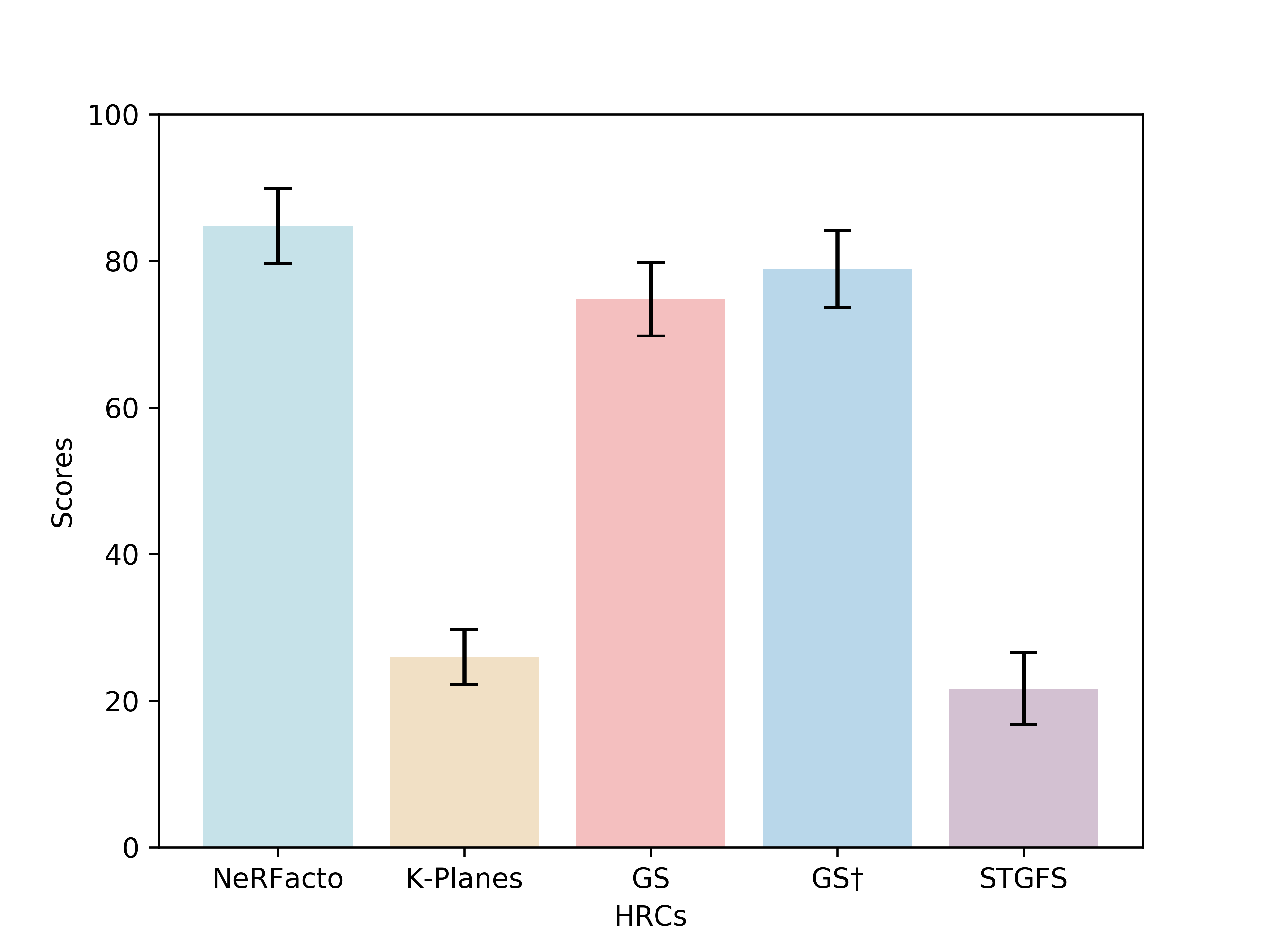}
		\centerline{(g). Frog}
	\end{minipage}
	\begin{minipage}[b]{0.19\textwidth}
		\centering
		\includegraphics[width=\textwidth]{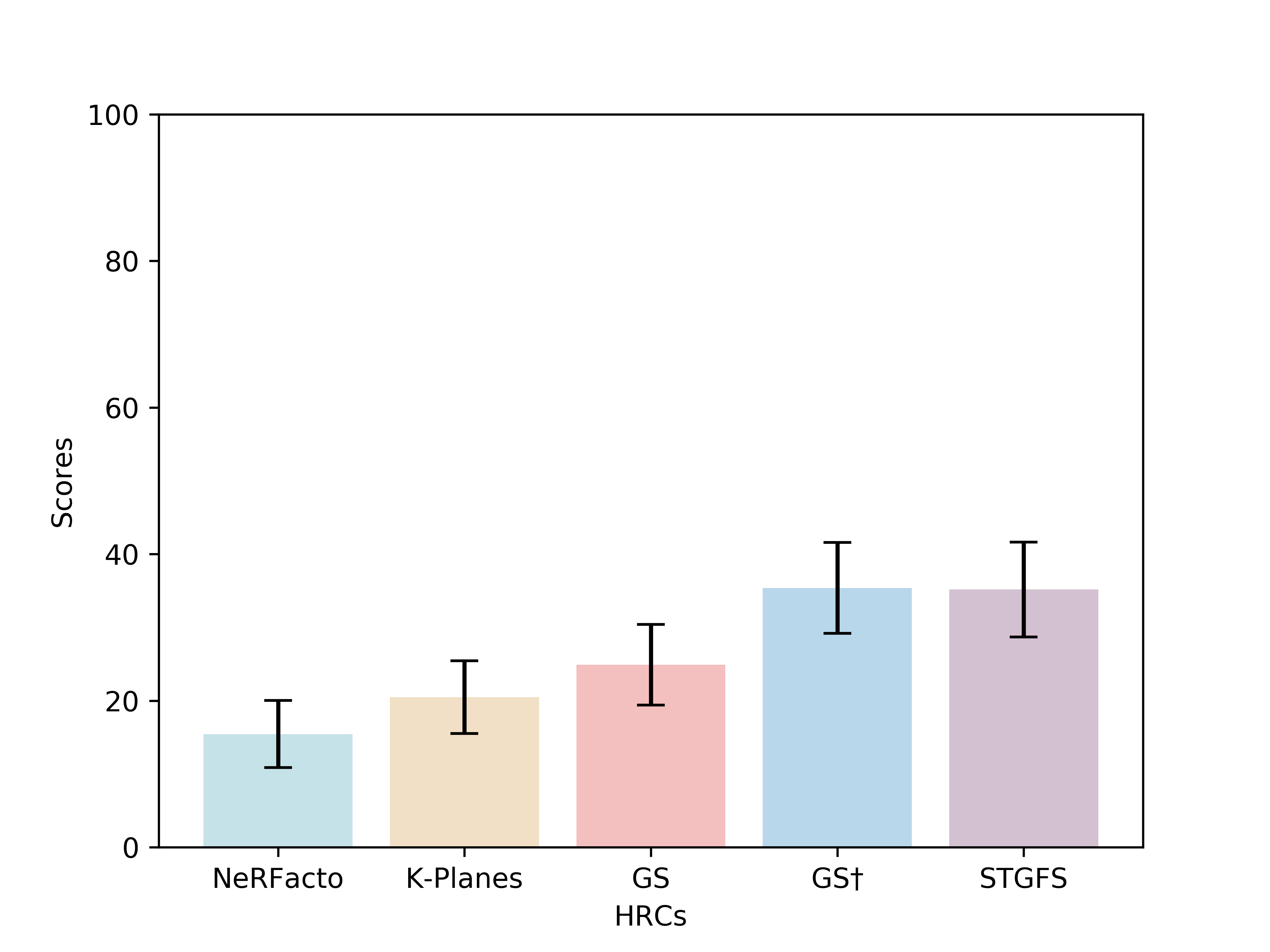}
		\centerline{(h). MartialArts}
	\end{minipage}
	\begin{minipage}[b]{0.19\textwidth}
		\centering
		\includegraphics[width=\textwidth]{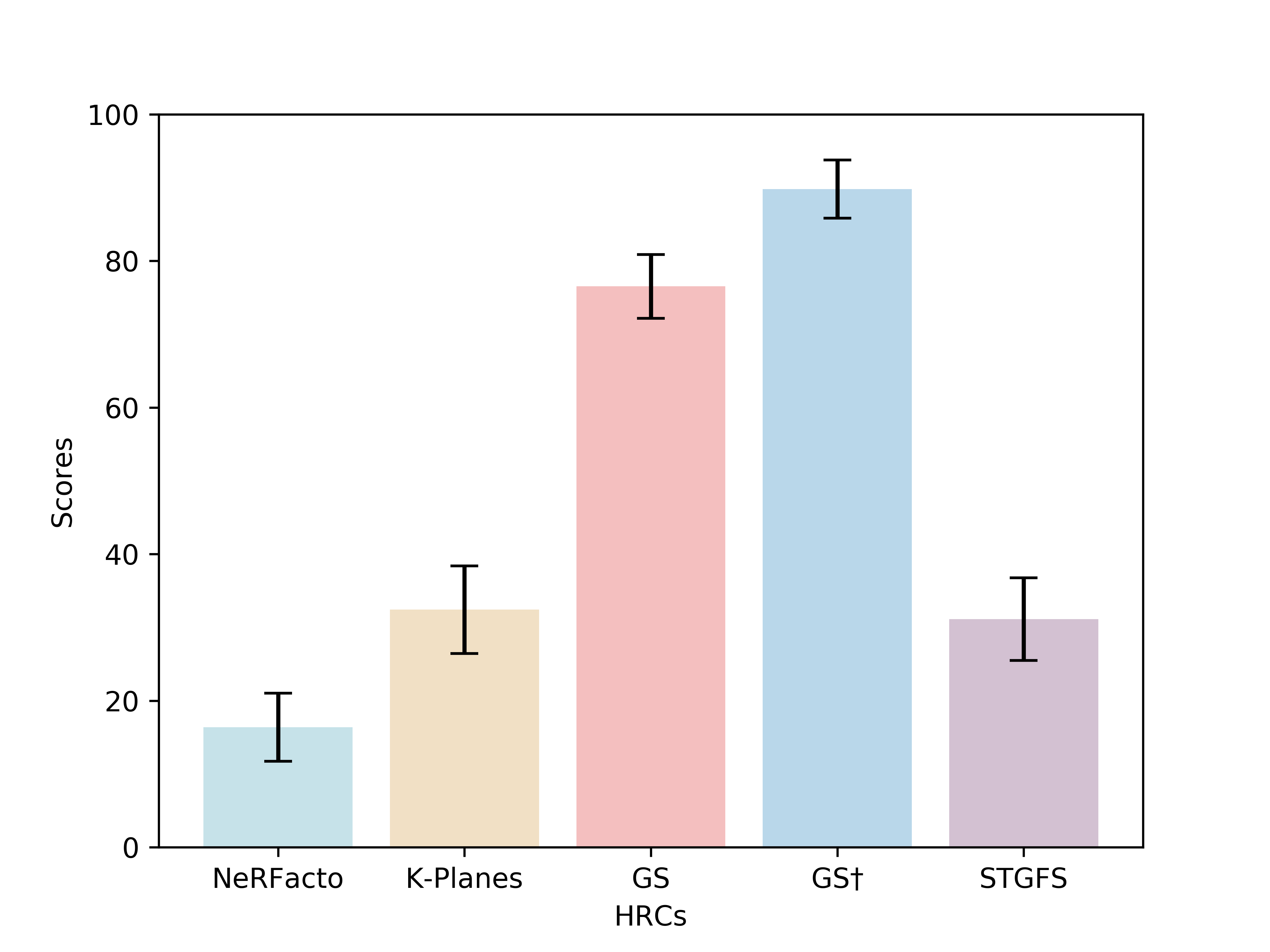}
		\centerline{(i). Painter}
	\end{minipage}
	\begin{minipage}[b]{0.19\textwidth}
		\centering
		\includegraphics[width=\textwidth]{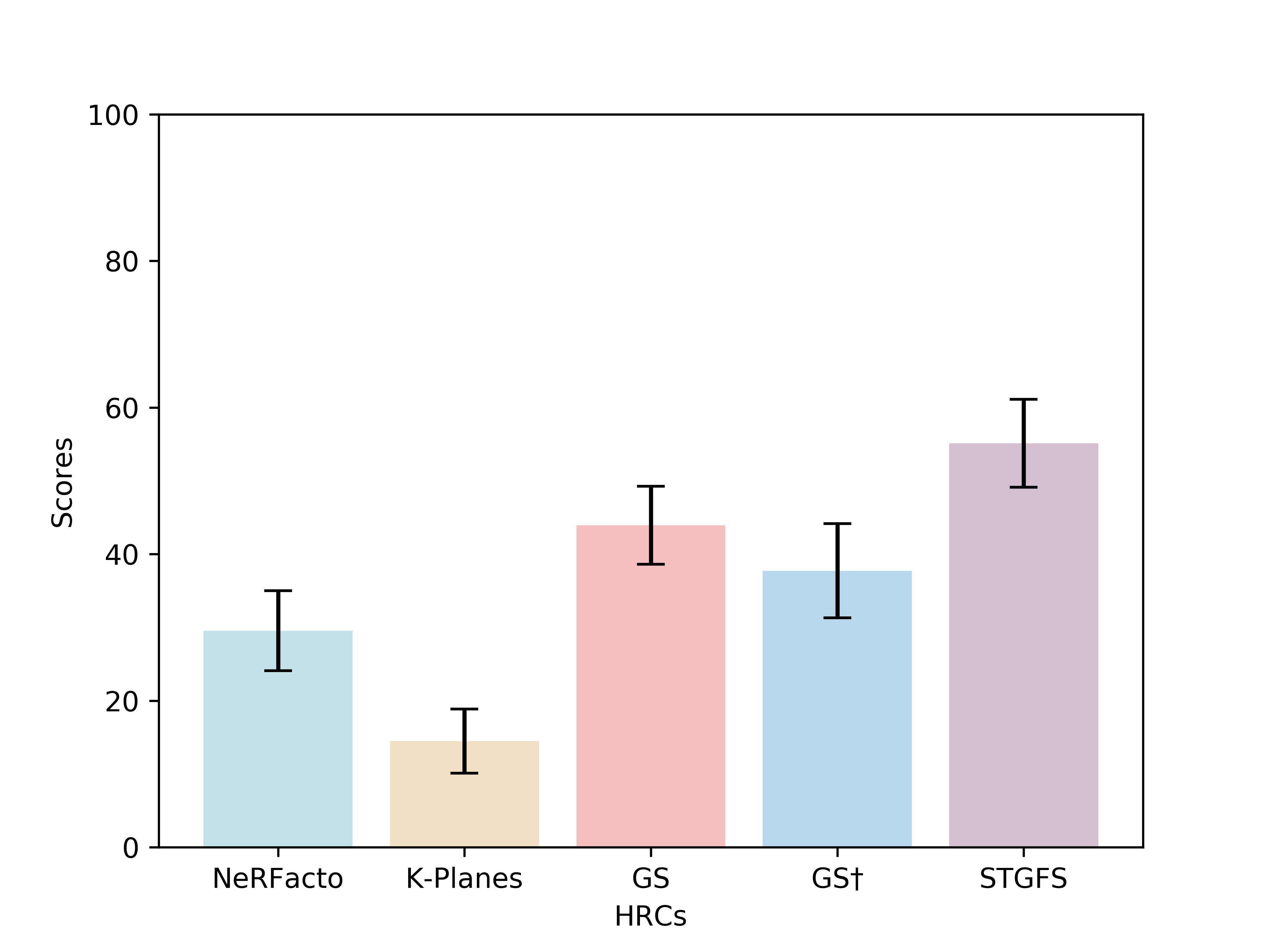}
		\centerline{(j). Pandemonium}
	\end{minipage}
	\begin{minipage}[b]{0.19\textwidth}
		\centering
		\includegraphics[width=\textwidth]{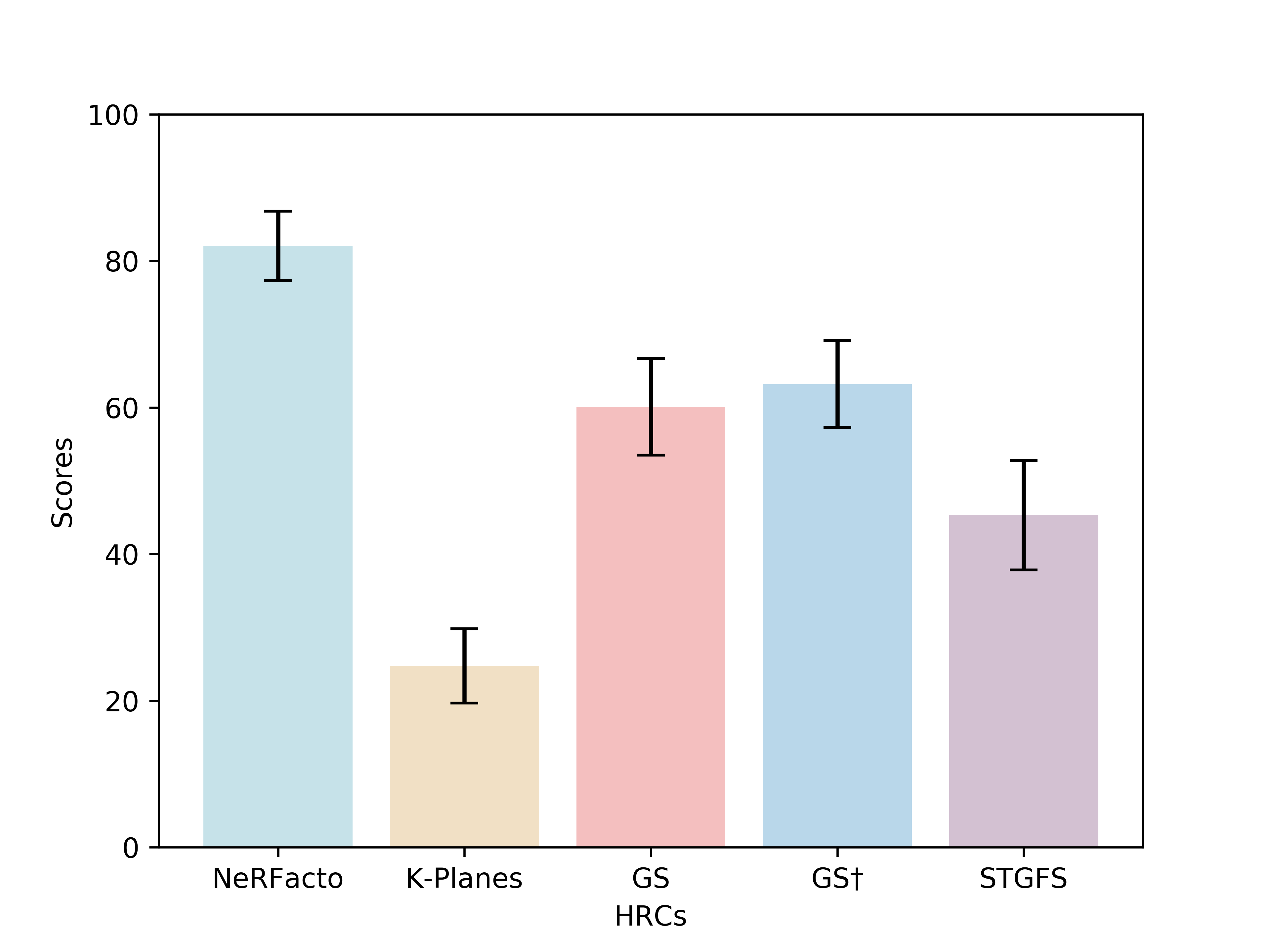}
		\centerline{(k). PoznanStreet}
	\end{minipage}
	\begin{minipage}[b]{0.19\textwidth}
		\centering
		\includegraphics[width=\textwidth]{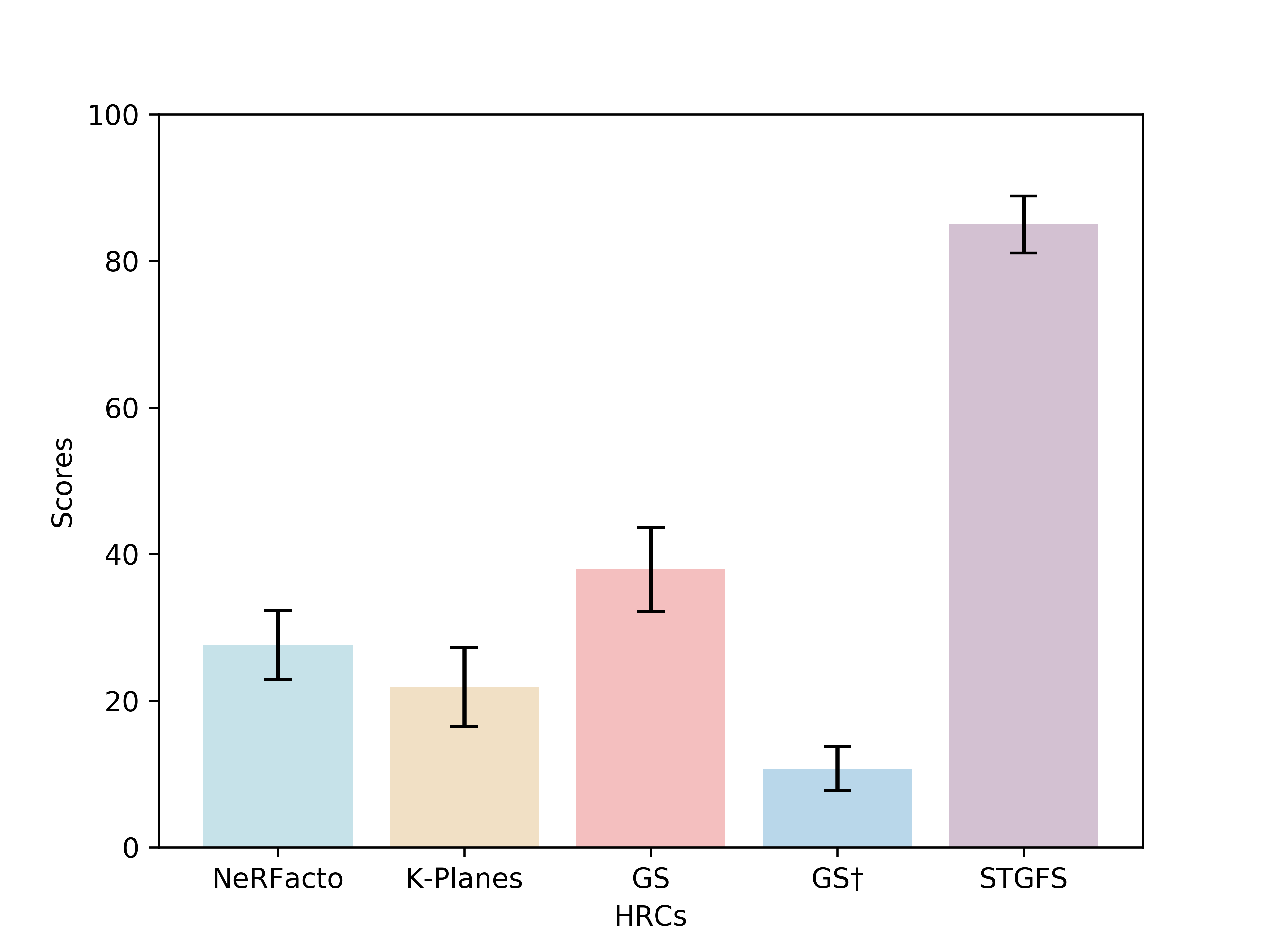}
		\centerline{(l). CoffeeMartini}
	\end{minipage}
	\begin{minipage}[b]{0.19\textwidth}
		\centering
		\includegraphics[width=\textwidth]{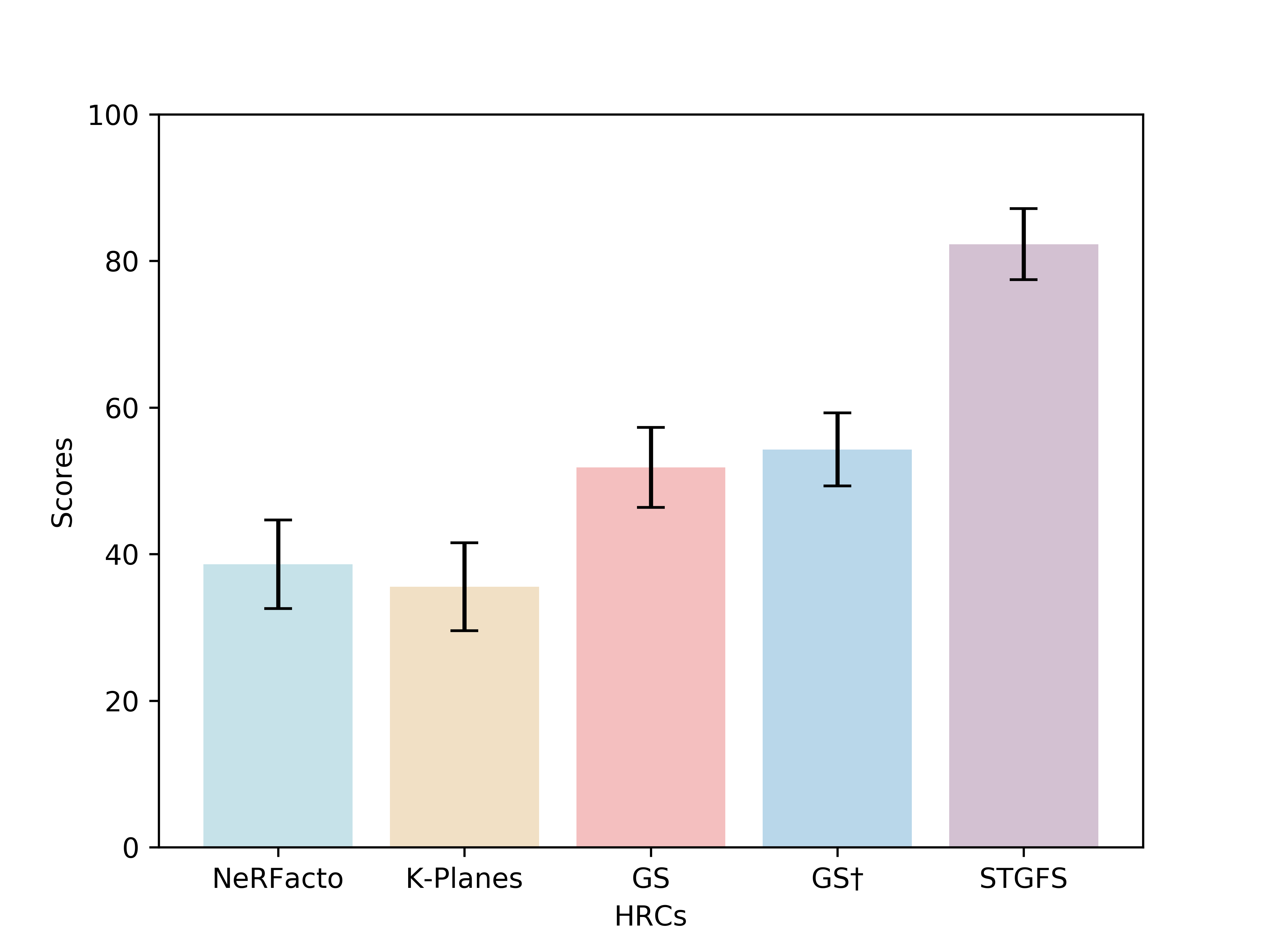}
		\centerline{(m). Flamesteak}
	\end{minipage}
	\caption{MOS of single-view subjective test.}
	\label{fig:MOS-S}
\end{figure*}

\subsection{Advantages of The Proposed Database}  
\label{sec:3e}
Although some previous NVSQA databases were proposed, there are some potential unexplored gaps in the NVSQA domain. The advantages of our database are as follows, corresponding to the limitations described in Section \ref{sec:1}.
\par \textbf{1)} We innovatively introduce both GS-based and NeRF-based methods in the subjective experiment to dig out their visual quality. The effects of GS-based methods on the human visual system were explored for the first time. Although GSC-QA \cite{yang2024benchmark} also assessed the quality of the GS method, it primarily investigates the GS compression quality, rather than the quality of the GS-based method itself. Thus, the proposed dataset essentially is not of the same type as the one they introduced.
\par \textbf{2)} More real-world scenes are presented in our database. It can be seen that the number of real-world scenes of the presented dataset exceeds the majority of methods. Although FFV \cite{liang2024perceptual} has more real-world scenes than ours, their PVS only contains the front-facing view. 
\par \textbf{3)} The types of PVS in our dataset are various, covering multi-view visual pathways (360°, front-facing) and the single-view visual pathway. To the best of our knowledge, it's the first time to explore the quality of the single-view visual pathway and the relationship between multi-view and single-view visual pathways.
\par \textbf{4)} The exploration of NVS for dynamic scenes in the NVSQA dataset marks the first instance of such a study. This also contrasts with GSC-QA \cite{yang2024benchmark} primarily assessing NVS compression quality although it includes dynamic scenes. Our dataset provides an in-depth analysis of the NVS methods themselves, rather than their compression quality.
\par These advantages significantly contribute to the development of the NVSQA community, giving a thorough understanding of the strengths and weaknesses of NVS methods, thereby advancing the whole NVS-related field.

\section{Result Analysis}
\label{sec:4}
Results are divided into score distribution, overall mean opinion score, the relationship between different visual paths, and the relationship between different HRCs.

\subsubsection{Score Distribution}
The score distributions of both multi-view and single-view tests are respectively calculated with 10-point intervals, as illustrated in Fig \ref{fig:Frequency}. It can be seen that the largest discrepancies only are 5.3\% and 8.5\% in the multi-view and single-view tests respectively. More notably, apart from the extreme score ranges (such as 0-9, 80-89, 90-99), the largest discrepancies in other intervals remain below 1.5\% and 7.2\% in multi-view and single-view tests, respectively. These observations demonstrated a broad and uniform distribution.
\subsubsection{Overall Mean Opinion Score}
To analyze and assess the quality of NVS methods, Mean Opinion Score (MOS), as a common and fundamental measure in quality assessment \cite{ITU2023}, is used in our proposed database, calculated by averaging the scores of all valid participants. Fig. \ref{fig:MOS-Overall} depicted the average MOS of both subjective experiments. Fig. \ref{fig:MOS-D} and Fig. \ref{fig:MOS-S} depicted the MOS of different scenes in the multi-view and single-view tests, respectively. These Figures present the 95\% confidence interval, where the narrow confidence interval reflects consistency in the experimental results, indicating that the subjective tests are more reliable. More importantly, several key observations can be summarized as follows.
\begin{itemize}
	\item \textbf{Discrepancies across scenarios}. Participant ratings have discrepancies across scenarios, confirming the ability of NVS to have various adaptability to different scenarios.
	\item \textbf{Quality comparison between front-facing and 360° PVSs}. The overall quality of front-facing PVSs was better than that of 360° PVSs. For instance, the number of front-facing PVSs surpassing 60-score was 14, compared to 8 for 360° PVSs. Similarly, for the PVSs above 70-score, the counts of the former and latter were 10 and 4, respectively. 
	\item \textbf{Quality comparison between multi-view and single-view tests}. The quality in the single-view condition generally showed better than that in the multi-view condition. For example, the number of ratings exceeding 60-score was 26 in the single-view test while there were 22 in the multi-view tests. For scores over 70, there are 18 and 14 for the former and latter, respectively.
	\item \textbf{NVS method comparison}. In the multi-view test, STGFS performed the best. Specifically, the number of scores larger than 60 in NeRFacto, K-Planes, GS, GS$\dagger$, and STGFS is 3, 0, 4, 5, and 6, while for scores exceeding 70, the numbers are 3, 0, 3, 1, and 6 respectively. In addition, STGFS still ranked as the best in the single-view test. Same as the above order, scores over 60 are 3, 0, 3, 3, and 5, and scores over 70 at 3, 0, 2, 2, and 4, respectively.
\end{itemize}
\subsubsection{The Effect of Different Visual Paths}
\begin{figure}[t]
	\centering
	\includegraphics[width=0.48\textwidth]{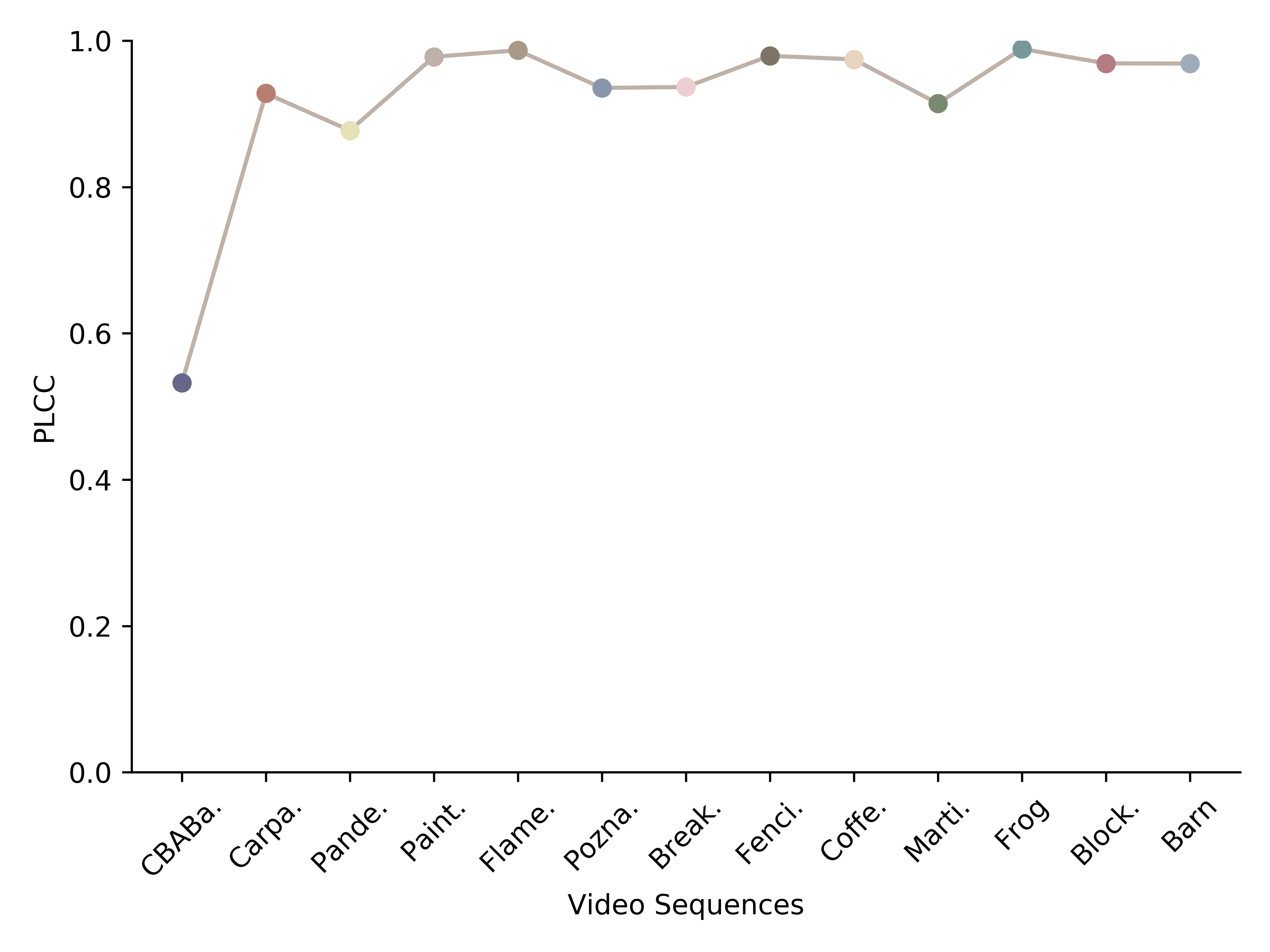}
	\caption{PLCC between multi-view and single-view pathways.}
	\label{fig:PLCCView}
\end{figure}

Fig. \ref{fig:PLCCView} depicted the relationship between the multi-view and single-view tests by using the Pearson Linear Correlation Coefficient (PLCC) metric. The results demonstrated that some sequences have high similarities between the two tests, especially in Painter, CoffeeMarini, Frog, Barn, and PoznanStreet. Note that most of them are front-facing PVSs that have a small moving path compared to 360° PVSs, thereby contributing to the rationalization of this phenomenon. It also can be evident that there are larger discrepancies between the two tests in the condition of 360° PVSs, such as Pandemonium, CBABasketball, and MartialArts. These results showed the observer preference varied when the visual path changed, which compiles well with the conclusion in \cite{tabassum2024quality}. However, these discrepancies between the two subjective tests are subtle, indicating that evaluating the performance of NVS methods using simpler single-view videos can also provide valuable insights.

\subsubsection{The Effect of Different HRCs}
\begin{figure}[t]
	\centering
	\includegraphics[width=0.48\textwidth]{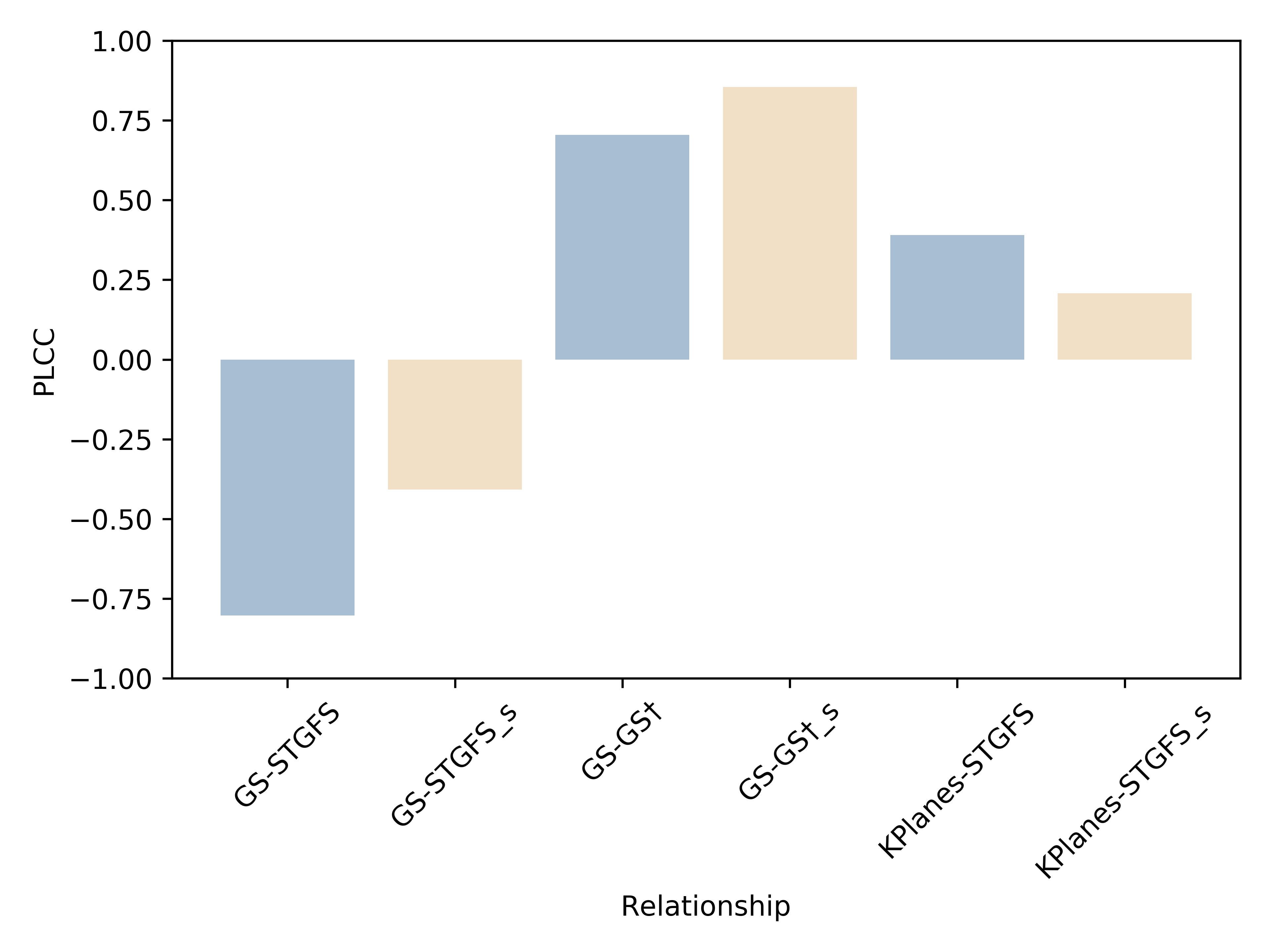}
	\caption{The relationship between different HRCs.}
	\label{fig:Relation}
\end{figure}

To further analyze the influences discussed in Section \ref{sec:3} B, the relationship among some HRCs using PLCC metric is drawn as Fig. \ref{fig:Relation}. The relationships between some methods exhibit low or even negative correlations, indicating their distinct characteristics. In contrast, GS and GS$\dagger$ show a stronger correlation in both multi-view and single-view tests, which is reasonable since the difference between them is only the training iteration. More observations also can be found in Fig. \ref{fig:MOS-D} and Fig. \ref{fig:MOS-S}, which relate to the discussions mentioned in Section \ref{sec:3} B. Contrary to the static NVS method, the dynamic NVS method did not achieve better quality. This is reflected in the worse quality of PVSs generated by K-Planes compared to NeRFacto. STGFS, as a dynamic-based method, performed better than the other methods both in the multi-view and single-view tests. Additionally, GS with fewer training iterations GS$\dagger$ performs even better than the formal GS, demonstrating rapid convergence of GS. In multi-view tests, GS$\dagger$ surpasses the quality of formal GS in 9 out of 13 scenes, while in single-view tests, GS$\dagger$ excels in 11 out of 13 scenes. As NVS methods of the same type, STGFS is obviously better than K-Planes, showing the strong NVS ability and accuracy in GS-based methods. Meanwhile, generally GS-Based methods is better than NeRF-based methods, as evidenced by the average MOS across scenes. The average MOS across scenes for the multi-view test in NeRFacto, K-Planes, GS, GS$\dagger$, and STGFS are 42.32, 25.43, 52.61, 54.15, and 57.29, respectively, while they respectively are 38.12, 24.46, 46.73, 49.89, and 54.46 for the single-view test. Clearly, the scores for GS-based methods consistently exceed those of NeRF-based methods.

\begin{table*}[t]
    \centering
    \caption{Performance Comparison of Objective Quality Assessment Metrics in the NVS dataset. \textbf{Bold} indicates the best performance, while \uline{underline} signifies the second-best performance.}
    \begin{tabular}{@{}llcc|cccc|cccc@{}}
        \toprule
        \multirow{2}{*}{Type} & \multirow{2}{*}{Year} & \multirow{2}{*}{Name} & \multirow{2}{*}{FR/NR} & \multicolumn{4}{c}{NVS Database - Static/Single view} & \multicolumn{4}{c}{NVS Database - Dynamic/Multi view} \\ \cmidrule(lr){5-8} \cmidrule(lr){9-12}
         &  &  &  & PLCC & SRCC & KRCC & RMSE & PLCC & SRCC & KRCC & RMSE \\ \midrule
         & - & PSNR & FR & 0.8131 & 0.6256 & 0.4906 & 10.4398 & - & - & - & - \\
         & - & PSNR-L & FR & 0.8298 & 0.6207 & 0.4883 & 10.1055 & - & - & - & - \\
         & 2004 & SSIM \cite{wang2004image} & FR & 0.8118 & 0.6063 & 0.4900 & 10.3934 & - & - & - & - \\
         & 2003 & MS-SSIM \cite{wang2003multiscale} & FR & \textbf{0.8941} & \uline{0.7242} & \uline{0.6047} & \textbf{8.2191} & - & - & - & - \\
         & 2006 & VIF \cite{sheikh2006image} & FR & 0.7915 & 0.5335 & 0.4067 & 10.9885 & - & - & - & - \\
         & 2011 & FSIM \cite{zhang2011fsim} & FR & \uline{0.8663} & 0.6775 & 0.5616 & 8.9557 & - & - & - & - \\
         & 2000 & NQM \cite{damera2000image} & FR & 0.8636 & 0.7129 & 0.5689 & 8.9238 & - & - & - & - \\
         & 2005 & IFC \cite{sheikh2005information} & FR & 0.7872 & 0.5139 & 0.3924 & 11.1049 & - & - & - & - \\
         & 2011 & HDR-VDR-2 \cite{mantiuk2011hdr} & FR & 0.7506 & 0.6740 & 0.5352 & 11.7070 & - & - & - & - \\
       IQA  & 2018 & HaarPSI \cite{reisenhofer2018haar} & FR & 0.8439 & 0.6412 & 0.5133 & 9.7189 & - & - & - & - \\
         & 2012 & BRISQUE \cite{mittal2012no} & NR & 0.7340 & 0.5379 & 0.4002 & 12.4378 & 0.6871 & \uline{0.5520} & \uline{0.4194} & 14.4020 \\
         & 2013 & NIQE \cite{mittal2012making} & NR & 0.7513 & 0.4876 & 0.3606 & 11.9987 & \uline{0.7041} & 0.4163 & 0.2895 & 14.0488 \\
         & 2015 & PIQE \cite{venkatanath2015blind} & NR & 0.5434 & 0.3388 & 0.2385 & 15.8610 & 0.5317 & 0.3433 & 0.2570 & 16.0666 \\
         & 2015 & IL-NIQE \cite{zhang2015feature} & NR & 0.6271 &	0.4075 &	0.3014 &	14.3277 &	0.6391 & 	0.4283 &	0.3225 &	14.2720 \\
         & 2019	& ENIQA\cite{chen2019no} &NR	&0.6982 	&0.4927 	&0.3738 	&12.6777 	&0.6387 	&0.4393 	&0.3316 	&15.0315 \\
         &2019	& NBIQA	\cite{ou2019novel} &NR	&0.7554 	&0.5902 	&0.4566 	&11.9597 	&\textbf{0.7294} 	& 
        \textbf{0.5854} 	& \textbf{0.4382} &	\textbf{13.6221} \\
         & 2018 & BMPRI \cite{min2018blind} & NR & 0.6868 & 0.4708 & 0.3465 & 13.0289 & 0.6794 & 0.5111 & 0.3886 & 13.8710 \\ \midrule
         & 2015 & HDR-VQM \cite{narwaria2015hdr} & FR & 0.8546 & \textbf{0.7821} & \textbf{0.6601} & \uline{8.4136} & - & - & - & - \\
         & 2017 & SpEED \cite{bampis2017speed} & RR & 0.6985 & 0.7250 & 0.5707 & 14.9227 & - & - & - & - \\
    VQA     & 2016 & VIIDEO \cite{mittal2015completely} & NR & 0.6746 & 0.3030 & 0.2200 & 13.0545 & 0.6463 & 0.3846 & 0.2923 & 15.0095 \\
         & 2019 & TLVQM \cite{korhonen2019two} & NR & 0.6801 & 0.4378 & 0.3212 & 12.6783 & 0.6988 & 0.4974 & 0.3766 & \uline{13.7236} \\
         & 2020 & NRVQA-NSTSS \cite{dendi2020no} & NR & 0.7730 & 0.5577 & 0.4281 & 11.6966 & 0.6308 & 0.4420 & 0.3202 & 14.9596 \\
         & 2021 & RAPIQUE \cite{tu2021rapique} & NR & 0.6619 & 0.4414 & 0.3307 & 13.2234 & 0.6357 & 0.3990 & 0.2983 & 14.9106 \\ 
        \bottomrule
    \end{tabular}
    \label{tab:OPC}
\end{table*}

\section{Objective Metric Experiment}
\label{sec:5}
\subsection{Objective Experimental Setting}

The aforementioned subjective experiments provided two new databases for objective metrics: a no-reference multi-view NVS dataset and a reference single-view NVS dataset. To comprehensively evaluate the performance of objective metrics on the proposed NVS datasets, we selected over 20 representative image and video quality assessment (IQA and VQA) metrics for performance testing. The IQA metrics include PSNR \cite{hore2010image} , PSNR-L , SSIM \cite{wang2004image}, MS-SSIM \cite{wang2003multiscale}, VIF \cite{sheikh2006image}, FSIM \cite{zhang2011fsim}, NQM \cite{damera2000image}, IFC \cite{sheikh2005information}, HDR-VDR-2 \cite{mantiuk2011hdr}, HaarPSI \cite{reisenhofer2018haar}, BRISQUE \cite{mittal2012no}, NIQE \cite{mittal2012making}, PIQE \cite{venkatanath2015blind},  IL-NIQE \cite{zhang2015feature}, ENIQA \cite{chen2019no}, NBIQA \cite{ou2019novel}, and BMPRI \cite{min2018blind}. The VQA metrics include HDR-VQM \cite{narwaria2015hdr}, SpEED \cite{bampis2017speed}, VIIDEO \cite{mittal2015completely}, TLVQM \cite{korhonen2019two}, NRVQA-NSTSS \cite{dendi2020no}, and RAPIQUE \cite{tu2021rapique}. Based on the extent to which methods use reference information, they can be classified into Full-Reference (FR), Reduced-Reference (RR), and No-Reference (NR). FR metrics assess quality by comparing the distorted image/video to a high-quality reference. RR metrics utilize indirect information from the reference, while NR metrics evaluate quality solely based on the distorted content itself. To prevent overlap between training and validation sets, we adopt the leave-two-fold-out cross-validation approach in our experiments. The data is divided into multiple splits until the validation sets cover all SRCs. These splits are then used for training and testing. Pearson Linear Correlation Coefficient (PLCC), Spearman Rank-order Correlation Coefficient (SRCC), Kendall Rank Correlation Coefficient (KRCC), and Root Mean Square Error (RMSE), respectively reflecting linear relationship, monotonicity, rank correlation, and predictive accuracy, are employed to assess the performance of objective metrics in fitting subjective scores.
\subsection{Performance analysis}

\par \textbf{Reference metrics (FR + RR) $vs$ NR.} Due to the absence of reference data on the multi-view NVS dataset, this section focuses solely on the performance analysis of the single-view NVS dataset. As shown in Table \ref{tab:OPC}, the best and second-best methods on the single-view NVS dataset are FR metrics. Additionally, the best FR metrics outperform the best NR metrics in PLCC, SRCC, KRCC, and RMSE by 0.1211, 0.1919,  0.2035, and 3.4775, respectively. Furthermore, the average performance of the FR and RR metrics in PLCC, SRCC, KRCC, and RMSE is 0.8171 0.6531, 0.5235, and 10.3244, which surpasses the NR metrics' performance of 0.1275, 0.1926, 0.1801, and 2.6705. These experimental results underscore the significant impact that high-quality reference data has on the performance of subjective metrics.
\par \textbf{IQA $vs$ VQA metrics.}  On the single-view NVS dataset, as shown in Table \ref{tab:OPC}, the MS-SSIM method achieves the best performance in 2 evaluation metrics and the second-best in the other 2 evaluation metrics. The VQA method HDR-VQM obtained two best performances among the four evaluation metrics. The average performance for the IQA methods in PLCC, SRCC, KRCC, and RMSE are 0.7675, 0.5680, 0.4429, and 11.3440, respectively, outperforming VQA methods by 0.0437, 0.0268, 0.0211, and 0.9875. On the multi-view NVS dataset, the IQA method NBIQA achieves the best performance in 4 evaluation metrics, with second-best performances in all evaluation metrics also produced by IQA methods. Furthermore, the average performance of IQA methods is superior to that of VQA methods across all evaluation metrics, presenting that IQA methods are more effective on both reference and no-reference scenarios.
\par \textbf{Performance comparison between single-view and multi-view datasets.} Due to the inability to test reference methods on the no-reference multi-view dataset, this subsection focuses on comparing the performance of NR methods across the two datasets. As shown in Table \ref{tab:OPC}, the best NR performance on the reference single-view dataset surpasses that of the no-reference multi-view dataset by 0.0436, 0.0048, 0.0184, and 1.9255 in PRCC, SRCC, KRCC and RMSE evaluation metrics, respectively. The average performance of NR methods on the reference single-view NVS dataset outperforms that on the no-reference multi-view NVS dataset across all evaluation metrics. Overall, the NR methods demonstrate better performance on the single-view NVS dataset, indicating that the scenarios in single-view NVS data are relatively simpler.
\section{Conclusion}
\label{sec:6}
In this paper, we proposed a comprehensive NVS quality assessment database, meticulously curated with specific considerations for NVS method selection, scene coverage, and assessment pathways. First, for NVS method selection, our database focuses on the distortions for both NeRF-based and GS-based methods, providing in-depth analysis and indicating their discrepancies. Second, our database contains 13 real-world scenes captured by both large and small distance camera arrays, almost exceeding that in previous databases. Third, we investigated the influences of multi-view and single-view PVSs, and their relationship on the human vision system. Fourth, it's the first time to explore the impact of dynamic scenes with moving objects within NVS methods. Finally, a comprehensive benchmark of various state-of-the-art objective metrics on the proposed database is established, revealing that existing methods still struggle to accurately capture subjective quality. In summary, our study reveals distinct impacts of NVS methods on human perception under dynamic, real-world scenarios and across multi-view and single-view visual pathways, thereby establishing a solid foundation for future research in both NVSQA and NVS methods.

\section{Acknowledgement}
Approval of all ethical and experimental procedures and protocols was granted by the Medical Ethical Committee Approval of Shenzhen University Health Science Center under Application No. PN-202400161.

%

\bibliographystyle{IEEEtran} 
\bibliography{egbib} 

\vspace{11pt}

\vfill

\end{document}